\documentclass{article}

\usepackage[eandd,final]{neurips_2026}

\usepackage[utf8]{inputenc}
\usepackage[T1]{fontenc}
\usepackage{amsmath, amsfonts}
\usepackage{graphicx}
\usepackage{booktabs}
\usepackage{hyperref}
\usepackage{xcolor}

\title{ObjView-Bench: Rethinking Difficulty and Deployment for Object-Centric View Planning}

\author{%
  Sicong Pan\thanks{Also affiliated with the Lamarr Institute for Machine Learning and Artificial Intelligence and the Center for Robotics, University of Bonn, Germany.} \\
  University of Bonn\\
  \texttt{pan@cs.uni-bonn.de} \\
  \And
  Hao Hu\thanks{Also affiliated with Moore Threads Technology Co., Ltd., China.} \\
  Independent Researcher\\
  \texttt{hozierhu@gmail.com} \\
  \And
  Xuying Huang$^*$ \\
  University of Bonn\\
  \texttt{huang@cs.uni-bonn.de} \\
  \And
  Benno Wingender$^*$ \\
  University of Bonn\\
  \texttt{wingende@cs.uni-bonn.de} \\
  \And
  Maren Bennewitz$^*$ \\
  University of Bonn\\
  \texttt{maren@cs.uni-bonn.de} \\
}

\begin{document}

\maketitle

\begin{abstract}
Object-centric view planning is a core component of active geometric 3D reconstruction in robotics, yet existing evaluations often conflate object complexity, planning difficulty, budget assumptions, and physical reachability constraints.
As a result, conclusions drawn from idealized view-planning evaluations may not reliably predict performance under realistic reconstruction settings.
We introduce ObjView-Bench, an evaluation framework for rethinking difficulty and deployment in object-centric view planning.
First, we disentangle three quantities underlying view-planning evaluation: omnidirectional self-occlusion as an object-side attribute, observation saturation difficulty, and protocol-dependent planning difficulty defined through a set-cover formulation.
This separation supports controlled dataset construction, analysis of slow-saturation objects, and a case study showing that planning difficulty-aware sampling can improve learned view planners.
Second, we design deployment-oriented evaluation protocols that reveal how budget regimes and reachable-view constraints alter method behavior.
Across classical, learned, and hybrid planners, ObjView-Bench shows that difficulty, budget, and reachability constraints substantially change method rankings and failure modes.
Project page: \url{https://objview-bench.github.io/}
\end{abstract}

\section{Introduction}
\label{sec:introduction}

\begin{figure*}[t]
  \centering
  \includegraphics[width=0.95\textwidth]{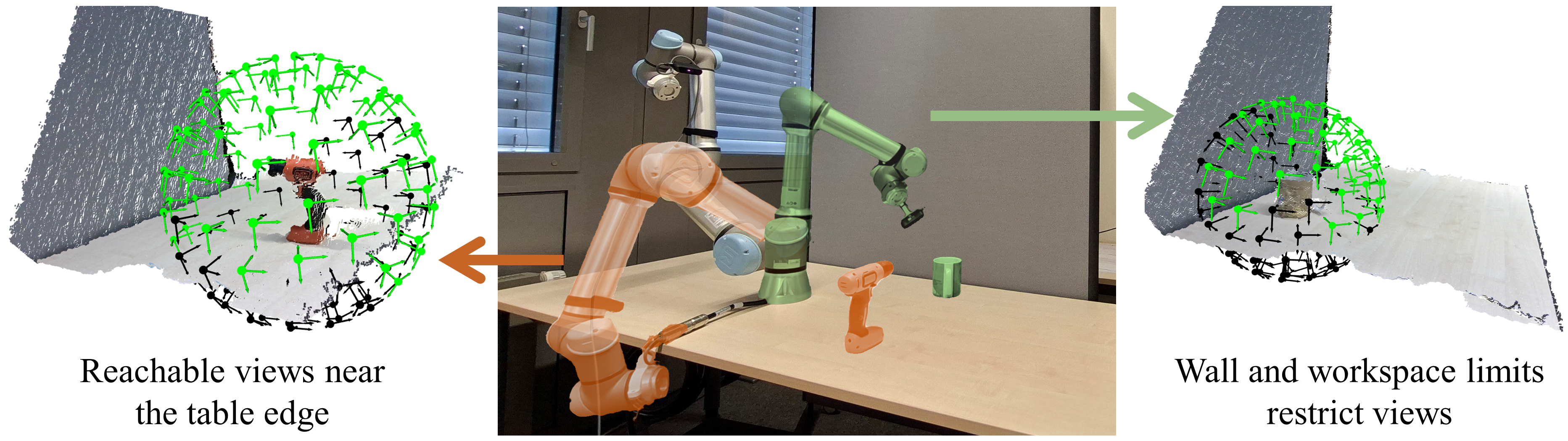}
  \caption{
  Object-centric view planning in a tabletop robot deployment.
  Starting from a home pose, the robot has to select feasible camera poses to reconstruct an object with unknown geometry. 
  Hereby, different object geometries and placements induce different reachable-view subsets: green views are executable, while black views are filtered out by inverse kinematics and collision checks.
  The transparent robot overlays illustrate representative feasible actions under different trials.
  }
  \label{fig:teaser}
  \vspace{-0.4cm}
\end{figure*}

View planning has long been studied in robot active vision across diverse applications~\citep{zeng2020view}.
Object-centric view planning for active geometric 3D reconstruction is a basic yet deployment-critical instance of this problem: given an unknown object, a robot has to decide where to look next in order to recover its geometry under limited sensing and motion resources.
As illustrated in Figure~\ref{fig:teaser}, this decision is shaped by both the object being reconstructed and the deployment workspace.
A simple object may be covered by a few generic views, whereas objects with cavities or inter-part occlusions may require more specific viewing directions; under a limited view budget, such differences affect what can be reconstructed.
At the same time, real robotic platforms rarely provide the full idealized view space: workspace, collision, and kinematic constraints restrict which camera poses are physically reachable.

However, current evaluation practice often treats these factors as incidental setup choices rather than as variables that can change the conclusion of a benchmark.
Despite extensive progress in view-planning methods for 3D reconstruction, with next-best-view (NBV) planning as a common formulation~\citep{alsadik2025structured}, studies commonly differ in object sources, action spaces, budgets, stopping criteria, and assumptions on the simulated or real robotic system.
This fragmentation of evaluation makes results difficult to compare across methods, and more importantly, can obscure why a method obtains a low surface coverage score.
It may indicate surfaces that are not visible from any feasible external viewpoint, surfaces hidden by deep concavities, or physically unreachable camera poses.
It may also indicate a planner-level limitation, such as repeatedly selecting redundant or low-informative views under a fixed budget.
As a result, rankings obtained under a single idealized view-space protocol may not reliably indicate which methods are suitable for deployment.

To bridge this gap, we introduce ObjView-Bench, a difficulty-aware and deployment-oriented benchmark for object-centric view planning.
Our contributions are threefold:
(i) We formalize difficulty semantics for object-centric view planning by disentangling omnidirectional self-occlusion, observation saturation difficulty, and protocol-dependent planning difficulty.
(ii) We construct a difficulty-characterized benchmark with an online hidden-geometry evaluation system, normalized surface coverage, budget protocols, and reachable-view constraints.
(iii) We evaluate representative classical, learned, and hybrid planners, showing that difficulty, budget, and reachability assumptions substantially change method behavior and failure modes.
To facilitate reproducible evaluation, we make our code, dataset, and evaluation platform available through our project website.

\section{Related Work}
\label{sec:literature}

\textbf{Object-centric view planning methods.}
Existing methods can be broadly grouped into classical planners, learned view predictors, and hybrid completion-planning approaches. 
Classical methods use information-gain objectives, geometric heuristics, or optimization-based search~\citep{daudelin2017adaptable,delmerico2018comparison, pan2023global}; learned methods predict views through classification, score regression, or reinforcement learning~\citep{mendoza2020supervised,pan2024integrating,zeng2020pc,li2024boundary,peralta2020next}; and hybrid methods first infer intermediate geometry through shape completion before performing explicit NBV or coverage-based planning~\citep{dhami2023pred,liu2024nbv}. 
Rather than proposing another planner, ObjView-Bench evaluates those three method families under shared difficulty and deployment semantics.

\textbf{Evaluation protocols and deployment assumptions.}
Many view-planning methods evaluate reconstruction quality under a fixed observation budget, typically after a predefined number of views~\citep{kriegel2015efficient,jia2025pb}.
Yet stopping criteria are themselves part of the evaluation semantics: some works use map-stabilization criteria~\citep{yervilla2022bayesian}, while others use set-cover-inspired rules tied to coverage completion and path cost~\citep{pan2022scvp}. 
Evaluation also varies with the physical view-space assumptions of the underlying platform, such as feasible mobile-manipulator configurations~\citep{vasquez2017view}, tabletop robotic-arm views~\citep{pan2021global}, or aerial free-space camera poses~\citep{chen2024gennbv}. 
These choices change the reachable view space, action parameterization, and collision constraints, and are often embedded directly into each method's planning and evaluation setup. 
This makes method comparisons difficult without explicit budget, stopping criteria, and reachability protocols.

\textbf{Object 3D reconstruction resources.}
Real-world 3D scanning datasets, such as the Stanford 3D Scanning Repository~\citep{krishnamurthy1996fitting} and Linemod~\citep{hinterstoisser2012model}, contain valuable captured objects, but their scale is often too small to support systematic difficulty stratification.
Large-scale 3D asset collections such as ShapeNet~\citep{chang2015shapenet} and Objaverse~\citep{deitke2023objaverse} offer broader coverage of object types, but they are not designed around the task-specific difficulty.
Consequently, existing active reconstruction studies often construct their own object pools from these datasets and treat objects as uniformly valid test instances.

\section{Disentangling Object Complexity and Planning Difficulty}
\label{sec:difficulty}

\subsection{Motivation for Disentanglement}

Objects are usually represented as 3D meshes, making it natural to define surface coverage: a method is rewarded for recovering a larger fraction of the ground-truth surface.
However, raw surface coverage implicitly treats the entire surface as an equally recoverable target.
This assumption can be misleading under strong self-occlusion or internal structures, where \textit{some ground-truth surface regions may never be observable from any external viewpoint}.
In such cases, low raw coverage does not necessarily indicate poor planning; it may instead reflect an observable-surface ceiling, which we capture through the omnidirectional self-occlusion attribute.

Even among surface regions that are externally observable, objects can differ substantially in how their observable surfaces are revealed.
Some geometries expose most of their observable surface from a small number of generic views, whereas \textit{others contain narrow cavities or concavities that are visible only from specific viewing angles}.
These objects may eventually achieve high coverage, but only after the view space is sampled densely enough to reveal such low-probability regions; we refer to
this as observation saturation difficulty.

Finally, the central question of active 3D reconstruction is how efficiently a robot can cover the target surface with a limited number of selected views.
Since \textit{different views often observe overlapping surface regions, the difficulty of this problem depends on the combinatorial structure of their coverage sets}: some views may be highly redundant, while others provide complementary observations.
Therefore, planning difficulty should be defined under a fixed candidate-view protocol, where the task is to select a compact set of views that cover the target surface as completely as possible.

\subsection{Preliminaries}

A view $v \in SE(3)$ denotes a camera pose from which the object is observed. 
Let $\mathcal{V}$ be a candidate view space. 
In the canonical object-centric setting used for difficulty characterization, $\mathcal{V}$ is instantiated as a spherical view set, where camera centers are sampled on an external sphere and oriented toward the object center, following a common setup in the view-planning literature for uniformly covering viewing directions around an object.
We consider a normalized object mesh $M$ in an object-centric coordinate frame. 
For difficulty characterization, we discretize the mesh surface into a finite set of surface voxels at resolution $r$, where $r$ is measured in the same normalized coordinate units, i.e., relative to the unit object scale.
We denote this discretized ground-truth surface by $S_{\mathrm{gt}}^{(r)}$.
Given a view $v$, we define $\mathcal{O}(v; r) \subseteq S_{\mathrm{gt}}^{(r)}$ as the subset of ground-truth surface voxels observable from $v$ under the visibility model.
In practice, this visibility operator is instantiated by ray-based first-hit visibility, so that only surface elements visible from the camera contribute to $\mathcal{O}(v; r)$.

\subsection{Object Complexity: From Omnidirectional Self-Occlusion to Observation Saturation}

For each view-set size $N$, let $V_N$ denote a set of $N$ approximately uniform object-centric views, obtained by solving the Tammes problem on the viewing sphere~\citep{lai2023iterated}.
The jointly observable surface voxels under $V_N$ and the corresponding observable-surface cardinality are
\begin{equation}
    S_N^{(r)} = \bigcup_{v \in V_N} \mathcal{O}(v; r), \qquad Y^{(r)}(N) = |S_N^{(r)}|.
\end{equation}
This curve measures how much of the ground-truth surface can be revealed as the external view space is sampled more densely.
Since Tammes sets of different cardinalities are not necessarily nested, we later use a monotone saturating approximation of $Y^{(r)}(N)$ when defining the observable-surface ceiling and observation saturation difficulty.
Let $N_\star^{(r)}$ denote the first evaluated view-set size at which the monotone approximation $\widehat{Y}^{(r)}(N)$ enters a stable low-gain regime.
We define the saturated observable-surface cardinality and the corresponding observable-surface ratio as:
\begin{equation}
Y_{\mathrm{sat}}^{(r)} = \widehat{Y}^{(r)}(N_\star^{(r)}),
\qquad
r_{\mathrm{vis}}^{(r)}
=
{Y_{\mathrm{sat}}^{(r)}}/{|S_{\mathrm{gt}}^{(r)}|}.
\end{equation}
The quantity $r_{\mathrm{vis}}^{(r)}$ is the \textbf{omnidirectional self-occlusion attribute} that estimates the externally observable fraction of the discretized surface.
We define the \textbf{observation saturation difficulty} as the saturation view count $d_{\mathrm{sat}}^{(r)} = N_\star^{(r)}$.

\subsection{Planning Difficulty under a Fixed Candidate-View Protocol}

We use a fixed Tammes candidate-view set $\mathcal{V}_{\mathrm{plan}}$ here and define
 the protocol-coverable target universe for an object as $U^{(r)} = \bigcup_{v_i \in \mathcal{V}_{\mathrm{plan}}} \mathcal{O}(v_i; r)$.
We define \textbf{planning difficulty} as the minimum number $d_{\mathrm{plan}}^{(r)}$ of views required to cover this target universe (Set Cover problem):
\begin{equation}
d_{\mathrm{plan}}^{(r)}
=
\min_{x_i \in \{0,1\}}
\sum_{i=1}^{|V_{\mathrm{plan}}|} x_i,
\quad
\mathrm{s.t.}
\quad
\forall u \in U^{(r)},\;
\sum_{i:\,u \in \mathcal{O}(v_i;r)} x_i \ge 1 .
\end{equation}

\begin{figure*}[t]
  \centering
  \begin{tabular}{c}
    \includegraphics[width=0.70\textwidth]{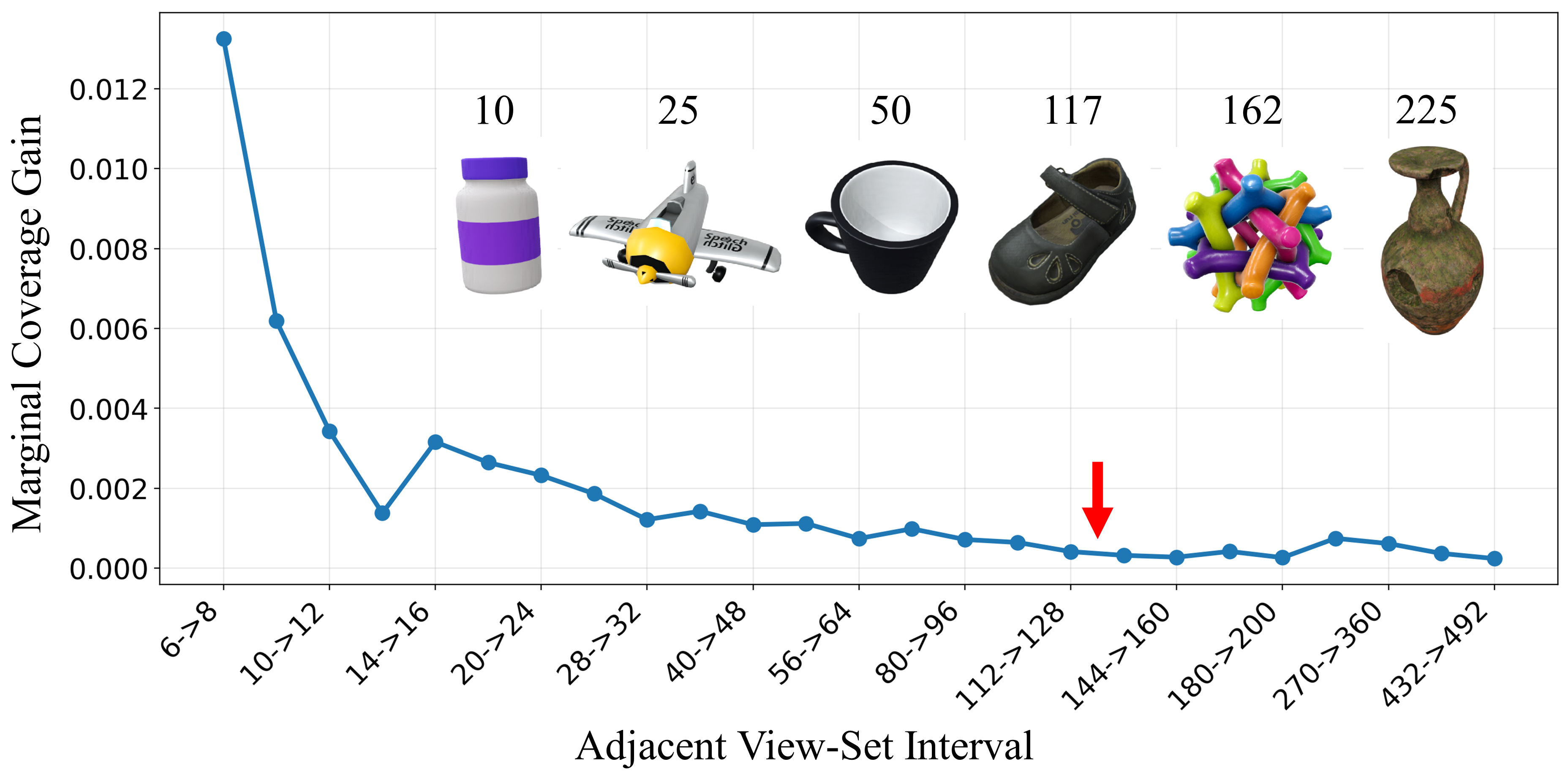} \\
    \small (a) Saturation behavior and 128-view protocol
  \end{tabular}

  \begin{tabular}{cc}
    \includegraphics[width=0.50\textwidth]{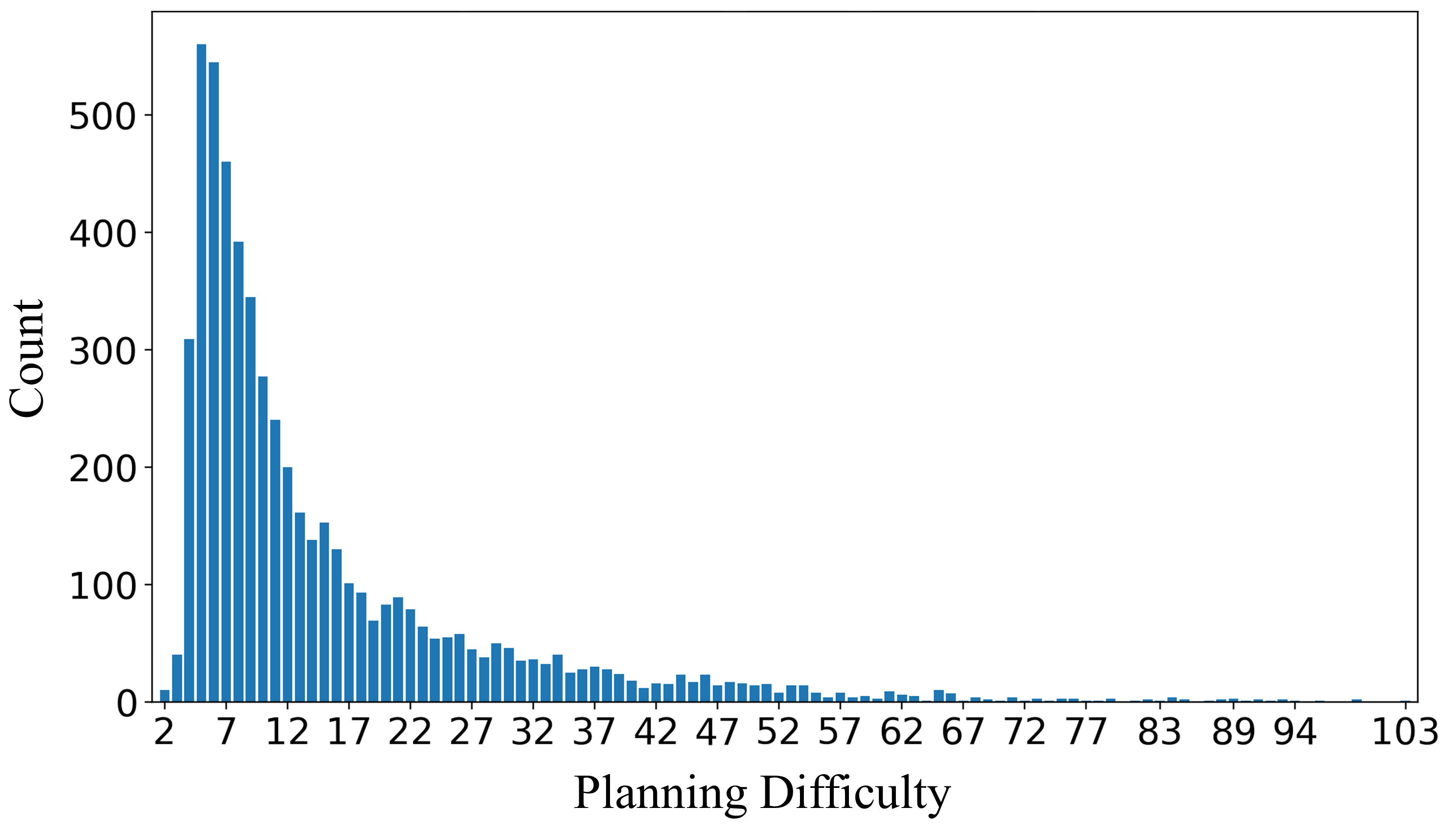} &
    \includegraphics[width=0.40\textwidth]{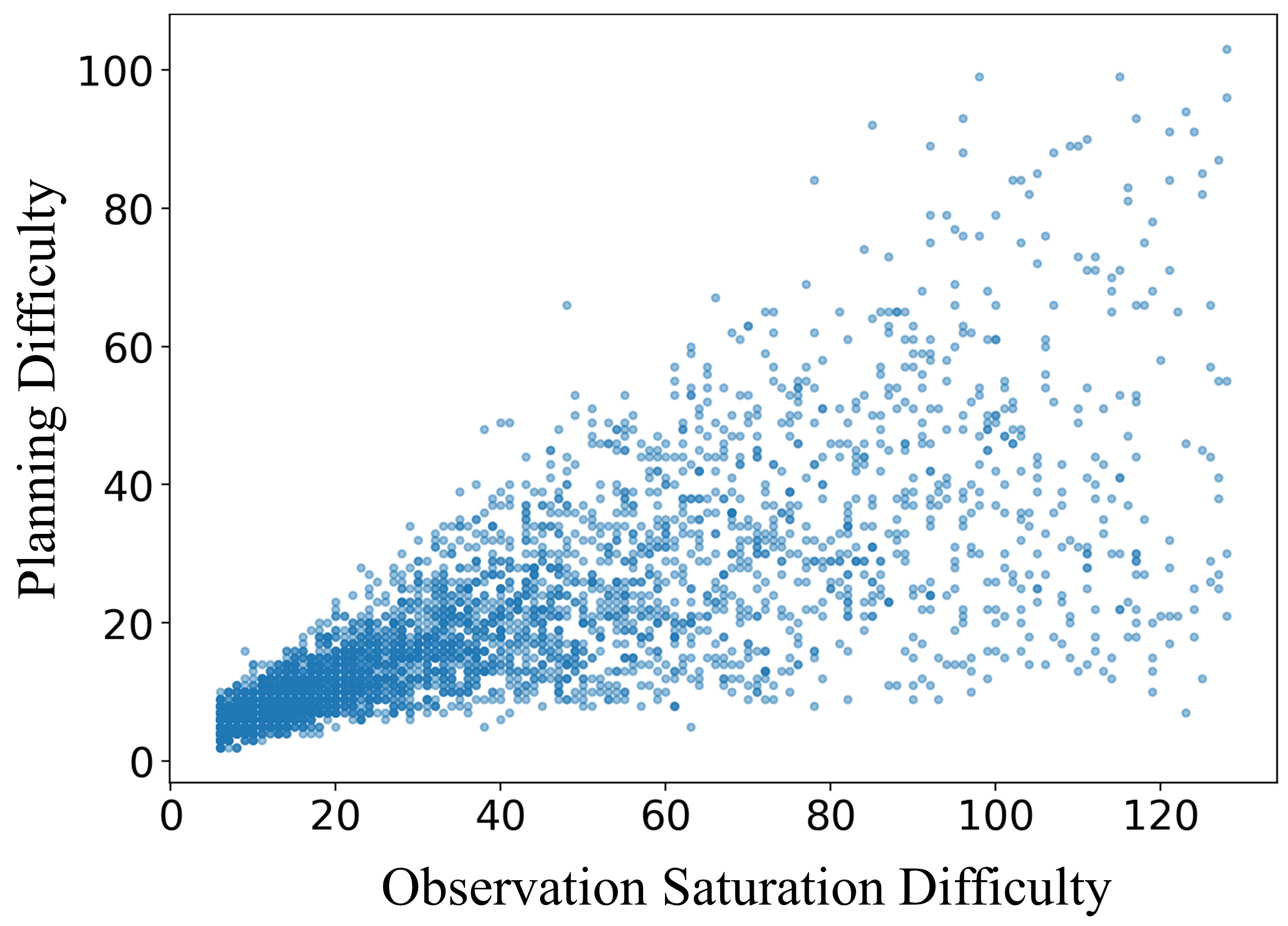} \\
    \small (b) Planning difficulty distribution &
    \small (c) Saturation vs. planning
  \end{tabular}

  \caption{
  Difficulty characterization of ObjView-Bench.
  (a) Representative objects are ordered by observation saturation difficulty $d_{\mathrm{sat}}$ and overlaid on the marginal gain curve; the arrow marks the 128-view protocol.
  (b) Planning difficulty $d_{\mathrm{plan}}$ is long-tailed in the main pool.
  (c) Observation saturation difficulty $d_{\mathrm{sat}}$ and planning difficulty $d_{\mathrm{plan}}$ are related but not equivalent.
  }
  \label{fig:difficulty_statistics}
  \vspace{-0.2cm}
\end{figure*}

\subsection{Implications for Benchmark Design}

In summary, the three introduced quantities play distinct roles in ObjView-Bench.
The omnidirectional self-occlusion attribute $r_{\mathrm{vis}}^{(r)}$ provides a reference ceiling for interpreting raw surface coverage and for reporting normalized surface coverage.
The observation saturation difficulty $d_{\mathrm{sat}}^{(r)}$ characterizes slow-saturation reveal behavior. 
We therefore use $d_{\mathrm{sat}}^{(r)}$ as a ranking signal for identifying slow-saturation or open-challenge objects.
In our benchmark instantiation, we also choose the cardinality of the fixed planning view set $\mathcal{V}_{\mathrm{plan}}$ based on saturation statistics of the main object pool, using the maximum $d_{\mathrm{sat}}^{(r)}$ over non-slow-saturation objects as a conservative coverage criterion.
The planning difficulty $d_{\mathrm{plan}}^{(r)}$  measures how many complementary views are needed in a fixed candidate-view set.
We use it as the difficulty notion for benchmark stratification and analysis.
Further details of the visibility operators, saturation fitting, and set-cover annotation are provided in Appendix~\ref{app:difficulty}.

\section{Dataset Construction and Difficulty Statistics}
\label{sec:dataset}

ObjView-Bench is constructed from large-scale 3D assets with the goal of obtaining a reliable and difficulty-characterized object pool for object-centric view planning. 
We use a staged construction pipeline and first apply quality filtering using Objaverse++~\citep{lin2025objaverse++}.
We then perform conservative geometry filtering and stratify the remaining candidates using lightweight shape and fill statistics.
Next, we run the difficulty annotation pipeline defined in Section~\ref{sec:difficulty}.
Finally, we conduct a manual reliability review after converting assets to a standardized rendering format.

After manual review, ObjView-Bench contains 5,691 objects. 
Figure~\ref{fig:difficulty_statistics} summarizes the difficulty characterization under the canonical view protocol with $r=0.02$.
Observation saturation varies substantially across objects, and the marginal gain curve motivates the fixed 128-view candidate protocol.
Qualitatively, slow-saturation objects are dominated by recurring restricted-visibility structures, such as deep cavities, aperture-mediated internal surfaces, and inter-part occlusions.
The $d_{\mathrm{sat}}$--$d_{\mathrm{plan}}$ scatter further shows that observation saturation difficulty is related to but distinct from planning difficulty.
We therefore separate 5,523 objects into the main pool and identify 168 slow-saturation objects whose $d_{\mathrm{sat}}$ exceeds the 128-view threshold used to instantiate the planning protocol.

From the main pool, we construct Main-Train-Public and Main-Hidden-Test.
Because planning difficulty is long-tailed, we sample the hidden test set with a two-stage difficulty-aware rule: first covering all non-empty $d_{\mathrm{plan}}$ buckets when possible, then allocating the remaining quota approximately proportional to bucket frequencies.
This avoids a test set dominated by low-difficulty objects while preserving the natural difficulty distribution.
Main-Hidden-Test contains 528 objects; the remaining 4,995 main-pool objects form Main-Train-Public.
The 168 slow-saturation objects are kept as a separate diagnostic subset.
Additional construction details, dataset statistics, and protocol sensitivity analyses are provided in Appendix~\ref{app:dataset}.

\section{Deployment-Oriented Evaluation Protocols and Benchmarking System}
\label{sec:protocol}

\subsection{Online Object-Centric Reconstruction Benchmark and Metrics}

ObjView-Bench evaluates methods in an online object-centric reconstruction setting.
To mimic realistic active perception, methods do not receive the object geometry, but only sequential RGB-D observations rendered from the camera poses they select.
For each episode, the benchmark server holds the hidden object geometry and renders RGB-D observations.
At each step, the algorithm returns a camera view pose. 
The benchmark checks feasibility, renders the corresponding RGB-D observation, updates the accumulated reconstruction state, and records per-step metrics.

At each step, the evaluator converts the newly rendered RGB-D frame into an observed point cloud and fuses it into an accumulated point cloud. 
Evaluation is performed on raw observation point clouds, thereby decoupling reconstruction evaluation from any particular mapping representation.
We report Normalized Surface Coverage (NSC), Path Cost (PC), the number of acquired views (\#Views), and Planning Runtime (PR) as the main metrics.
SC@$\tau$ measures the fraction of the full ground-truth surface point cloud covered by the accumulated observations, using $\tau$ as the threshold for counting a point as covered.
NSC@$\tau$ normalizes SC@$\tau$ by a view-space-specific observable-surface reference computed from a dense 360-view Tammes set after applying the corresponding reachability constraint.
Path cost measures accumulated camera travel distance in normalized object coordinates, using a collision-avoiding shell distance when straight transitions intersect the unit object.
Planning runtime records algorithm-reported computation time. 

\subsection{Budget Protocols and Stopping Semantics}

Objects with different geometric and planning difficulty may require substantially different numbers of views to reach comparable reconstruction quality, making budget a central deployment variable.
We evaluate budget behavior using three representative deployment settings: a small fixed budget~($K=5$), a large fixed budget ($K=30$), and automatic stopping.
Here, $K$ denotes the maximum number of algorithm-selected views after the standardized initial observation.
Fixed-budget settings thus ask what reconstruction quality a method can deliver with up to $K$ feasible view actions.

The automatic setting represents free-budget execution with natural stopping.
Based on the preceding difficulty characterization, 128 selected views provide a conservative upper bound that allows the observable surface of benchmark
objects to saturate in practice, while keeping evaluation finite.
This cap is reached only if no stopping criterion triggers earlier.
Methods with an explicit finite plan terminate according to their native stopping criterion. 
For iterative next-view methods without a natural stopping criterion, we use a benchmark-side Map-Stabilization criterion: $\mathrm{MS}@\delta$ voxelizes the accumulated observed point cloud at resolution $\delta$ and stops when the relative count gain is below $1\%$ for three consecutive steps.

\subsection{Reachability-Constrained View Spaces}

Beyond budget, deployment also constrains which views can be executed.
We therefore evaluate methods under two representative view-space settings: \textit{whole} and \textit{quarter}.
The \textit{whole} setting approximates the ideal object-centric view sphere, while \textit{quarter} restricts execution to one quarter of that sphere as a strongly constrained reachable-view region.
All methods are evaluated with the same reachability-aware execution interface. 
When the view space is constrained, methods may query feasibility from the benchmark and filter out unreachable views, analogous to checking inverse kinematics and collision constraints before executing a camera pose on a real robot.
Implementation details of the benchmark APIs and evaluation protocols are provided in Appendix~\ref{app:benchmark}.

\section{Experiments}
\label{sec:experiment}

\subsection{Baselines and Setup}

\begin{table}[t]
\centering
\caption{
Baseline families and execution interfaces under constrained views.
}
\label{tab:baselines}
\small
\begin{tabular}{llll}
\toprule
Method & Family & Execution Mode & Feasibility Handling \& Ordering\\
\midrule
Random+TSP & Classical & Budgeted plan & Solve over feasible views, TSP \\
RSE~\citep{delmerico2018comparison} & Classical & Iterative & Select best feasible view \\
RSE-Mov~\citep{delmerico2018comparison} & Classical & Iterative & Select best feasible view \\
NBV-Net~\citep{mendoza2020supervised} & Learned & Iterative & Stop if best view infeasible \\
BENBV-Net~\citep{li2024boundary} & Learned & Iterative & Select best feasible view \\
MA-SCVP~\citep{pan2024integrating} & Learned & Set cover plan & Filter infeasible views, TSP \\
PoinTr-C+NBV~\citep{dhami2023pred} & Hybrid & Iterative & Select best feasible view \\
PoinTr-C+SCP~\citep{pan2025efficient} & Hybrid & Set cover plan & Solve over feasible views, TSP  \\
PoinTr-C+MCP~\citep{pan2023global} & Hybrid & Budgeted plan & Solve over feasible views, TSP \\
\bottomrule
\end{tabular}
\vspace{-0.2cm}
\end{table}

ObjView-Bench is designed to study how different planning paradigms behave under difficulty, budget, and reachability shifts, rather than to produce a fine-grained leaderboard within a single method subclass.
We therefore select representative methods with distinct decision mechanisms from each family, so that changes in ranking can be interpreted at the paradigm level.
The selected baselines cover classical planning, learned next-view or view-set prediction, and hybrid completion-guided planning.
Table~\ref{tab:baselines} summarizes families, execution modes, and feasibility handling.
Baseline implementations and training details are provided in Appendix~\ref{app:baselines}.

\textbf{Execution mode.}
The baselines also differ in execution mode, which determines which deployment settings they can naturally enter.
Random+TSP and PoinTr-C+MCP are budgeted finite-plan methods and are evaluated under fixed-budget settings.
MA-SCVP and PoinTr-C+SCP produce finite view sets with native set-cover plan termination and are evaluated under the automatic setting.
RSE, RSE-Mov, NBV-Net, BENBV-Net, and PoinTr-C+NBV operate as iterative replanning methods and can thus be evaluated under both fixed-budget and automatic-stopping (with MS@0.02) protocols.

\textbf{Feasibility handling.}
We attach the same minimal reachability-aware adaptation to all methods.
Score-based methods select the highest-scoring feasible candidate; set-prediction methods remove infeasible views before execution; and classical or hybrid planners optimize directly over the feasible candidate subset.
NBV-Net is the exception: it outputs a single discrete view index rather than ranked candidates, so an infeasible prediction terminates the episode.

\textbf{Training setup.}
All learned methods are trained only on Main-Train-Public and evaluated on Main-Hidden-Test.
These learned baselines are trained under the unconstrained \textit{whole} view-space setting, so that the \textit{quarter} evaluation explicitly measures reachability shift at test time.

\textbf{Candidate view spaces.}
All candidate-set-compatible methods, including classical and hybrid planners, use the canonical Tammes-128 view set motivated by the difficulty statistics.
Discrete learned methods retain their trained Tammes-128 action spaces.
BENBV-Net retains its native dynamic boundary-based interface: Tammes-128 is used for rollout initialization, while subsequent candidate views are generated dynamically and evaluated with feasibility checks.
The effect of candidate-view discretization on planner performance is examined in Appendix~\ref{app:candidate_sensitivity}.

\subsection{Budget Regimes Change Method Ranking}

\begin{table}[t]
\centering
\caption{
Best-quality method per family and budget regime in the \textit{whole} view space, selected by NSC@0.02.
Rank is the overall quality rank before family selection, using unrounded values.
Views include the initial observation.
}
\label{tab:budget_quality_family}
\small
\begin{tabular}{lllrrrrr}
\toprule
Setting & Family & Best Method & Rank & \#Views & NSC@0.02 & Path Cost & Time bin (s) \\
\midrule
$K=5$ & Classical & RSE & 1 & 6.0 & 0.979 & 25.70 & $0.5$--$1$ \\
$K=5$ & Learned & NBV-Net & 2 & 6.0 & 0.978 & 24.77 & $0.5$--$1$ \\
$K=5$ & Hybrid & PoinTr-C+MCP & 3 & 6.0 & 0.973 & 19.72 & $0.5$--$1$ \\
\midrule
$K=30$ & Classical & Random+TSP & 2 & 31.0 & 0.997 & 42.21 & $<0.1$ \\
$K=30$ & Learned & BENBV-Net & 6 & 28.0 & 0.995 & 66.98 & $3$--$5$ \\
$K=30$ & Hybrid & PoinTr-C+NBV & 1 & 31.0 & 0.998 & 100.65 & $>5$ \\
\midrule
Automatic & Classical & RSE & 1 & 9.8 & 0.991 & 41.71 & $1$--$3$ \\
Automatic & Learned & MA-SCVP & 4 & 49.5 & 0.990 & 57.19 & $0.1$--$0.5$ \\
Automatic & Hybrid & PoinTr-C+SCP & 2 & 15.9 & 0.991 & 30.75 & $0.5$--$1$ \\
\bottomrule
\end{tabular}
\vspace{-0.2cm}
\end{table}

\begin{table}[t]
\centering
\caption{
Shortest-path method per family and budget regime in the \textit{whole} view space, requiring mean NSC@0.02 $\geq 0.95$.
Rank is among all qualifying methods.
Views include the initial observation and may reflect early termination.
}
\label{tab:budget_path_family}
\small
\begin{tabular}{lllrrrrr}
\toprule
Setting & Family & Best Method & Rank & \#Views & NSC@0.02 & Path Cost & Time bin (s) \\
\midrule
$K=5$ & Classical & RSE & 4 & 6.0 & 0.979 & 25.70 & $0.5$--$1$ \\
$K=5$ & Learned & BENBV-Net & 1 & 6.0 & 0.956 & 15.64 & $0.5$--$1$ \\
$K=5$ & Hybrid & PoinTr-C+MCP & 2 & 6.0 & 0.973 & 19.72 & $0.5$--$1$ \\
\midrule
$K=30$ & Classical & Random+TSP & 2 & 31.0 & 0.997 & 42.21 & $<0.1$ \\
$K=30$ & Learned & NBV-Net & 1 & 9.9 & 0.988 & 41.73 & $1$--$3$ \\
$K=30$ & Hybrid & PoinTr-C+MCP & 3 & 31.0 & 0.997 & 52.68 & $0.5$--$1$ \\
\midrule
Automatic & Classical & RSE-Mov & 3 & 12.1 & 0.991 & 33.89 & $1$--$3$ \\
Automatic & Learned & BENBV-Net & 1 & 10.4 & 0.986 & 27.49 & $1$--$3$ \\
Automatic & Hybrid & PoinTr-C+SCP & 2 & 15.9 & 0.991 & 30.75 & $0.5$--$1$ \\
\bottomrule
\end{tabular}
\vspace{-0.2cm}
\end{table}

We first evaluate budget behavior in the unconstrained \textit{whole} view space, isolating budget from reachability constraints. 
Iterative methods may terminate early when their native action interface cannot propose a valid new view, such as NBV-Net predicting an already visited index or BENBV-Net exhausting available boundary points. 
For budgeted coverage planners such as PoinTr-C+MCP, we fill the remaining budget with farthest candidates when the optimization target is exhausted, reflecting that fixed-budget deployment encourages using the available views to improve coverage.

Tables~\ref{tab:budget_quality_family} and~\ref{tab:budget_path_family} report the best quality and qualifying path-efficiency representative from each family, together with its overall rank.
The representatives and their relative rankings change across budget regimes.
At $K=5$, RSE and NBV-Net achieve the highest quality, while PoinTr-C+MCP remains competitive with a shorter path.
At $K=30$, PoinTr-C+NBV achieves the highest NSC@0.02, while Random+TSP attains nearly the same coverage with a much shorter path.
Thus, a simple random-acquisition pipeline with TSP-style ordering becomes competitive when sufficient views are available.
NBV-Net has the shortest qualifying path at $K=30$, but this result includes early termination at only 9.9 views on average.

Automatic stopping exposes another behavior.
RSE achieves the highest NSC@0.02, while PoinTr-C+SCP attains nearly identical quality with a shorter path.
MA-SCVP remains the highest-quality learned representative, but executes 49.5 views compared with 15.9 for PoinTr-C+SCP, indicating that learned set prediction does not necessarily produce a compact plan.
The quality and path-efficiency representatives also differ: BENBV-Net yields the shortest qualifying automatic path, while PoinTr-C+SCP provides higher quality at a modestly greater path cost.
These results show that fixed-budget and automatic-stopping evaluations are not interchangeable, and that quality, executed views, and motion cost must be interpreted jointly.

Total planning time is reported in coarse bins because exact wall-clock time depends on system load. 
Learned methods are generally fast due to amortized inference, while GPU-accelerated RSE remains practical but becomes more expensive as the number of replanning steps grows. 
Optimization-based planners (TSP, SCP, MCP) are also efficient at our candidate-set scale. 
Thus, runtime differences mainly reflect the execution mode: iterative planning, one-shot inference, or finite-plan optimization.

\subsection{Reachability Constraints Change Method Behavior}

\begin{table}[t]
\centering
\caption{
Reachability-induced degradation under each method's maximum-view deployment setting.
For each method, we compare \textit{whole} (W) and \textit{quarter} (Q) view spaces under the evaluated budget protocol that executes the most views.
$\Delta$NSC and $\Delta$SC are computed as W minus Q.
}
\label{tab:reachability_drop}
\small
\begin{tabular}{llrrrrr}
\toprule
Method & Family & \#Views W/Q & NSC@0.02 W/Q & $\Delta$NSC & SC@0.02 W/Q & $\Delta$SC \\
\midrule
Random+TSP & Classical & 31.0/31.0 & 0.997/0.998 & -0.0006 & 0.943/0.808 & 0.136 \\
RSE & Classical & 31.0/31.0 & 0.997/0.998 & -0.0014 & 0.943/0.809 & 0.135 \\
RSE-Mov & Classical & 31.0/31.0 & 0.996/0.998 & -0.0023 & 0.942/0.809 & 0.134 \\
\midrule
NBV-Net & Learned & 9.9/1.2 & 0.988/0.472 & 0.5155 & 0.932/0.387 & 0.545 \\
BENBV-Net & Learned & 28.0/23.9 & 0.995/0.979 & 0.0157 & 0.940/0.784 & 0.157 \\
MA-SCVP & Learned & 49.5/12.5 & 0.990/0.967 & 0.0235 & 0.937/0.772 & 0.165 \\
\midrule
PoinTr-C+NBV & Hybrid & 31.0/31.0 & 0.998/0.998 & -0.0005 & 0.944/0.809 & 0.136 \\
PoinTr-C+SCP & Hybrid & 15.9/14.8 & 0.991/0.994 & -0.0025 & 0.937/0.799 & 0.138 \\
PoinTr-C+MCP & Hybrid & 31.0/31.0 & 0.997/0.998 & -0.0007 & 0.944/0.808 & 0.136 \\
\bottomrule
\end{tabular}
\vspace{-0.2cm}
\end{table}

\begin{table}[!t]
\centering
\caption{
Map-stabilization behavior on main and slow-saturation objects.
Both MS and NSC are reported at resolution 0.02, and $\Delta$NSC denotes Tammes-128 NSC minus MS NSC.
}
\label{tab:slow_saturation_ms}
\small
\begin{tabular}{lrrrrrrr}
\toprule
Pool & N & $d_{\mathrm{sat}}$ & $d_{\mathrm{plan}}$ & MS Views & MS NSC & Tammes-128 NSC & $\Delta$NSC \\
\midrule
Main-Hidden & 528 & 32.2 & 18.5 & 9.0 & 0.997 & 1.000 & 0.0026 \\
Slow-Saturation & 168 & 161.7 & 45.9 & 12.7 & 0.988 & 0.996 & 0.0081 \\
\bottomrule
\end{tabular}
\vspace{-0.2cm}
\end{table}

We next evaluate how methods behave when the executable view space is constrained.
To focus on reachability-induced degradation rather than small-budget effects, Table~\ref{tab:reachability_drop} compares \textit{whole} and \textit{quarter} view spaces under each method's maximum-view deployment setting.
Two effects emerge. 
First, absolute surface coverage drops substantially for almost all methods, with $\Delta$SC around 0.13--0.17 for most robust methods, confirming that reachability constraints remove access to a large portion of the full object surface. 
Second, NSC remains high for most classical and hybrid planners because it is normalized by the view-space-specific observable-surface reference; these methods still cover most of what is reachable, even though their absolute SC decreases.

Learned planners reveal a more sensitive failure mode. 
NBV-Net is the most extreme case: its NSC drops from 0.988 to 0.472 and its executed views decrease from 9.9 to 1.2, reflecting its single-index action interface under infeasible predictions. 
BENBV-Net and MA-SCVP can filter infeasible candidates or predicted view sets, but their SC drops remain larger than most classical and hybrid planners. 
This suggests that feasibility handling alone does not fully preserve their reconstruction quality under constrained view spaces.
Overall, reachability constraints interact not only with the feasible views, but also with each method's action representation and execution behavior.
Complete results across budget and reachability settings are provided in Appendix~\ref{app:full_results}.

\subsection{Slow-Saturation Diagnostic Analysis}

The slow-saturation subset is used as a diagnostic split rather than a primary leaderboard component. 
We therefore analyze it with a ground-truth oracle rollout instead of comparing planner rankings. 
At each step, the oracle selects the candidate view with the largest additional ground-truth voxel coverage, and the Tammes-128 reference is obtained after exhausting the 128-view candidate protocol.

Table~\ref{tab:slow_saturation_ms} shows that MS@0.02 triggers after a similar number of views in both pools, but its meaning differs across difficulty regimes. 
On Main-Hidden-Test, MS@0.02 reaches NSC@0.02 of 0.997 and leaves only a 0.0026 gap to the Tammes-128 reference. 
On the slow-saturation subset, the mean $d_{\mathrm{sat}}$ and $d_{\mathrm{plan}}$ are much larger, and the residual NSC gap after MS is roughly three times larger. 
For slow-saturation objects, additional views can still reveal small but real surface regions after the observed map appears stable. 
Thus, map stabilization is a practical stopping signal, but it measures observed-map stabilization rather than long-horizon observable-surface saturation.
Additional slow-saturation diagnostics are provided in Appendix~\ref{app:slow_saturation_case_study}.

\subsection{Planning Difficulty-Aware Sampling}

\begin{figure*}[!t]
  \centering
  \begin{tabular}{cc}
    \includegraphics[width=0.62\textwidth]{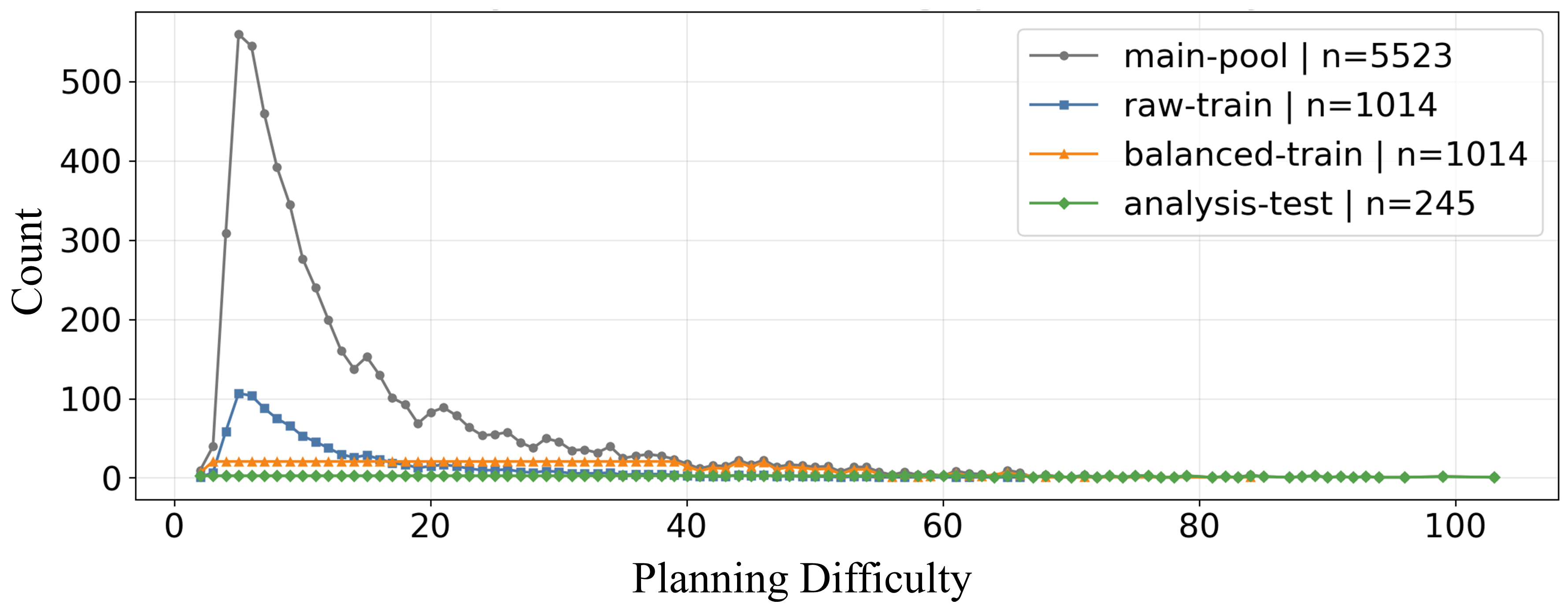} &
    \includegraphics[width=0.32\textwidth]{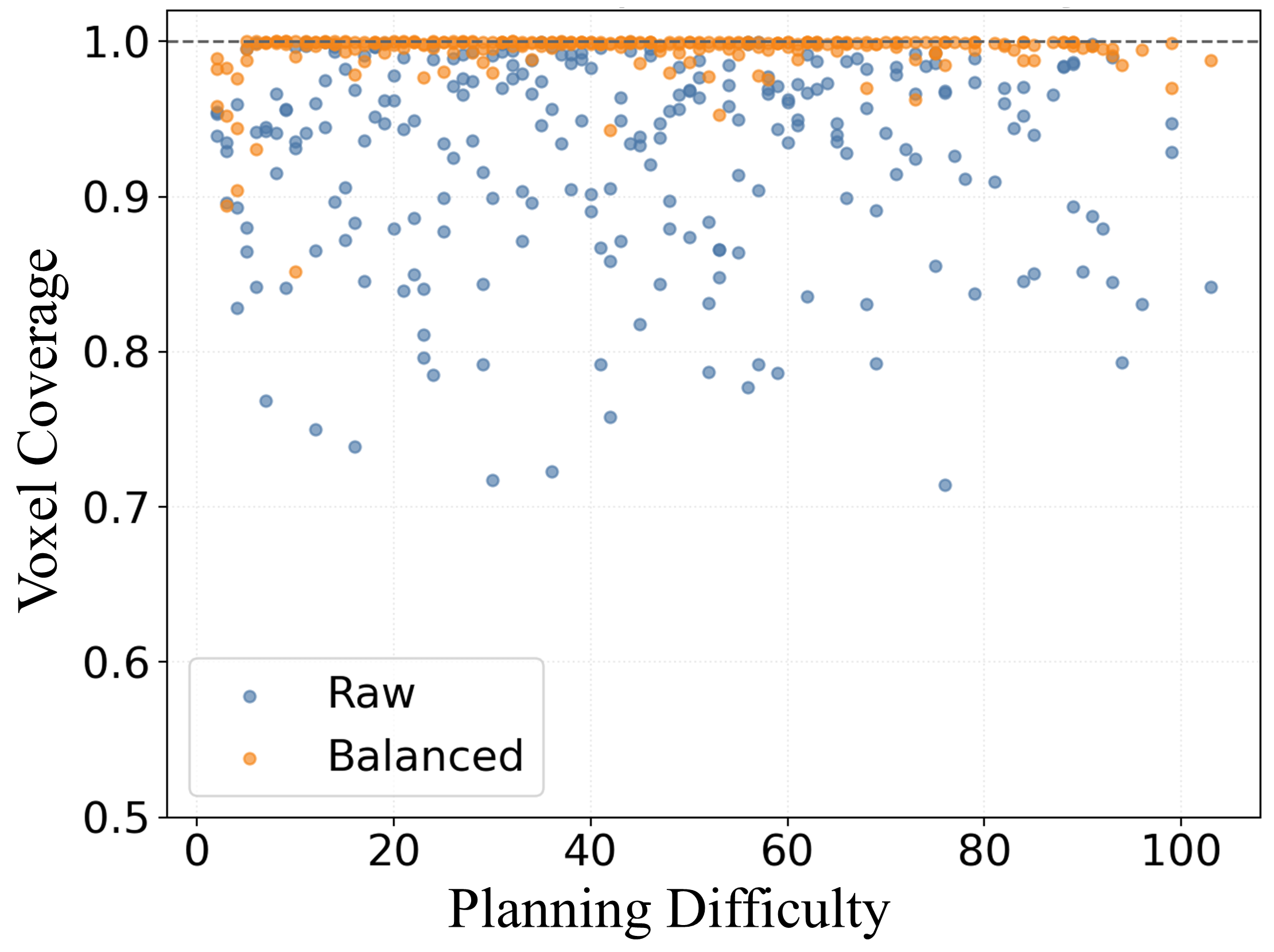} \\
    \small (a) Planning-difficulty distribution &
    \small (b) Difficulty-aware training effect
  \end{tabular}
    \caption{
    Planning difficulty-aware sampling.
    (a) Balanced sampling increases exposure to medium- and high-$d_{\mathrm{plan}}$ objects compared with raw sampling.
    (b) This improves MA-SCVP achieved voxel coverage across the planning-difficulty axis.
    }
  \label{fig:difficulty_aware_sampling}
  \vspace{-0.2cm}
\end{figure*}

We also test whether $d_{\mathrm{plan}}$ is only a descriptive statistic or also a useful axis for learning analysis. 
We construct two equal-size training subsets from the main pool: Main-Raw-Train follows the natural long-tailed $d_{\mathrm{plan}}$ distribution, while Main-Balanced-Train increases exposure to medium- and high-difficulty buckets. 
Both are evaluated on the held-out Main-Analysis-Test split.
Figure~\ref{fig:difficulty_aware_sampling} shows that $d_{\mathrm{plan}}$ can serve as an operational training-distribution axis.
Using \mbox{MA-SCVP} as a representative learned set-prediction planner, we evaluate achieved voxel coverage at \mbox{$r=0.02$}, matching the difficulty-annotation resolution.
Difficulty-aware sampling improves coverage on medium- and high-$d_{\mathrm{plan}}$ objects without changing the model architecture, suggesting that planning difficulty is a meaningful learning factor rather than only a descriptive benchmark statistic.

\subsection{Physical Reachability and Real-World Replay}

We further validate the deployment-oriented evaluation using a UR5e robot with an RGB-D camera~(Figure~\ref{fig:teaser}).
Across five objects and five placements, only 45.4/128 candidate views are physically reachable on average (range: 26--74), motivating the \textit{quarter}
constraint as a standardized reachability stress test rather than an exact robot workspace model.
Further details and visualizations are provided in Appendix~\ref{app:robot:reachability} and the supplementary video.

\begin{figure}[t]
  \centering
  \includegraphics[width=0.95\linewidth]{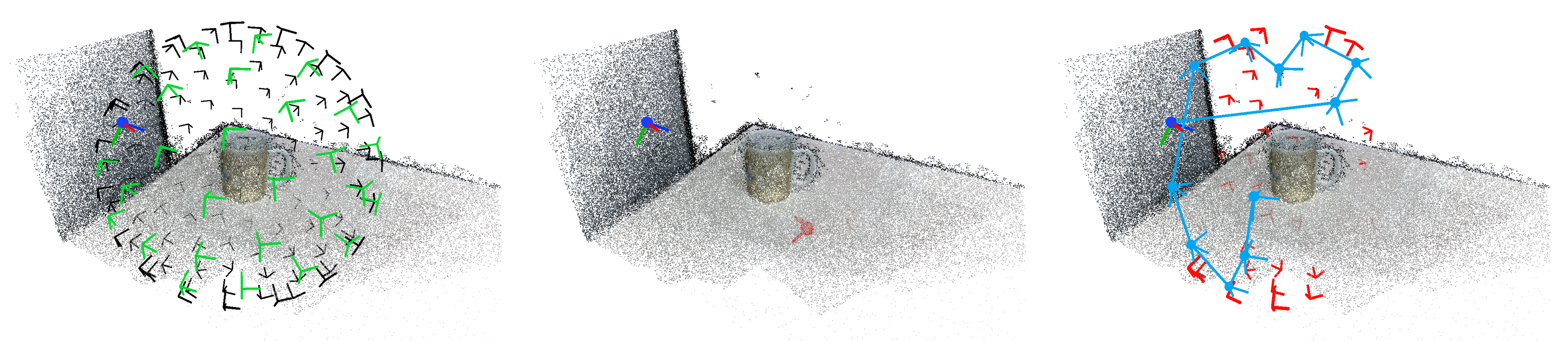}
  \caption{
    Representative learned-planner failures under physical constraints.
    Left: reachable (green) and infeasible (black) views.
    Middle: NBV-Net terminates after predicting an infeasible view.
    Right: MA-SCVP retains feasible views (blue) after filtering infeasible predictions (red).
  }
  \label{fig:robot_learned_reachability_showcase}
  \vspace{-0.2cm}
\end{figure}

Figure~\ref{fig:robot_learned_reachability_showcase} illustrates how physical reachability interacts with different learned action interfaces.
The aggregate executability analysis is provided in Appendix~\ref{app:robot:executability}.

We further evaluate eight compatible baselines through sequential replay of shared real RGB-D captures from physically reachable views.
This complements the executability diagnostic with real-observation-based planning evaluation.
The protocol and full results are provided in Appendix~\ref{app:robot:replay}.

\section{Lessons for Future View-Planning Methods}
\label{sec:lesson}

ObjView-Bench is intended not only as a leaderboard, but also as a diagnostic tool for identifying open challenges in object-centric view planning, especially for learning-based methods.
Our experiments highlight several research directions:
\textbf{(1) Scaling learning-based view planning.}
The canonical 128-view protocol and higher planning difficulty expose a harder regime for learning-based view planners than many prior object-centric settings.
Future learned planners should better handle large candidate view sets, diverse object geometries, and long planning horizons, where training stability and effective view selection become increasingly challenging.
\textbf{(2) Reachability-aware learning.}
Constrained view spaces interact strongly with learned action representations.
Test-time feasibility filtering is only a minimal adaptation; future methods should incorporate reachability information directly into the input, scoring function, or training objective.
\textbf{(3) Hybrid geometry learning and explicit planning.}
Completion-guided planning and explicit coverage optimization are currently robust under budget and reachability shifts, but introduce additional computation.
A promising direction is to accelerate the explicit planning stage.
\textbf{(4) Slow-saturation and budget allocation.}
Slow-saturation objects reveal a gap between map stabilization and observable-surface saturation.
This raises an open problem: how to allocate budget when remaining surfaces are hard to reveal, and marginal gains are small but persistent?
\textbf{(5) Planning difficulty as a learning signal.}
The difficulty-aware sampling experiment shows that planning difficulty is not merely a post-hoc statistic.
It may guide dataset construction, curriculum learning, and active learning by identifying objects that are more informative for training.

\section{Conclusion and Future Work}
\label{sec:conclusion}

We presented ObjView-Bench, a difficulty-aware and deployment-oriented benchmark, showing that self-occlusion, saturation, planning difficulty, budget, and reachability jointly shape view-planning behavior.
ObjView-Bench intentionally focuses on single-object, object-centric active geometric 3D reconstruction as a scalable and reproducible evaluation setting.
We further examine the benchmark's boundaries through physical reachability and real-RGB-D replay (Appendix~\ref{app:robot}), sensing-noise and sequential-fusion analyses (Appendix~\ref{app:sensing_noise}), and multi-object diagnostics (Appendix~\ref{sec:app_multi_object}).

Future work can extend this setting along several directions.
At the sensing level, extending our preliminary noise and fusion diagnostics toward systematic evaluation under depth noise and registration error remains an important direction.
At the task level, multi-object scenes, scene-level exploration, and multi-robot coordination introduce additional challenges beyond the current object-centric scope.
At the deployment level, integrating robot-specific motion planning and control would extend the current reachability-aware evaluation toward full closed-loop execution.
At the representation level, incorporating appearance-centric objectives and representations for novel-view synthesis, such as NeRF and 3D Gaussian Splatting, would require new difficulty definitions that account for texture and photometric fidelity.
Existing studies have explored NeRF-specific view requirements and geometry- and texture-aware view planning~\citep{pan2024many,pan2025dm}, but benchmark-level difficulty definitions remain an open problem.
At the objective level, while surface coverage is meaningful for geometric reconstruction and inspection, task-conditioned active perception may require metrics that capture task-relevant information rather than complete surface reconstruction, opening connections to active viewpoint selection in vision-language-action models~\citep{liu2026activevla}.
Finally, we plan to maintain a public leaderboard through our project website.

\begin{ack}
This work has partially been funded by the Deutsche Forschungsgemeinschaft (DFG, German Research Foundation) under grant 459376902 – AID4Crops, under Germany’s Excellence Strategy, EXC-2070 – 390732324 – PhenoRob, and by the German Federal Ministry of Research, Technology and Space (BMFTR) under the Robotics Institute Germany (RIG), grant No. 16ME0999.
This work did not involve an institutional collaboration between the University of Bonn and Moore Threads Technology Co., Ltd. The authors declare no other competing interests.
\end{ack}

\bibliographystyle{plainnat}
\bibliography{refs}

\newpage
\appendix

\section{Difficulty Annotation Details}
\label{app:difficulty}

This appendix provides implementation details for the difficulty annotation pipeline used to compute the observable-surface ratio $r_{\mathrm{vis}}^{(r)}$, the observation saturation difficulty $d_{\mathrm{sat}}^{(r)}$, and the planning difficulty $d_{\mathrm{plan}}^{(r)}$ used in ObjView-Bench.

\subsection{Surface Discretization and Camera Protocol}

For each normalized object, we start from the normalized surface point cloud and insert the points into an OctoMap representation at resolution $r=0.02$. Each occupied leaf node is treated as one discretized ground-truth surface element. The set of occupied leaf keys defines the discretized ground-truth surface:
\begin{equation}
S_{\mathrm{gt}}^{(r)}
=
\left\{
u
\;\middle|\;
u \text{ is an occupied OctoMap leaf key at resolution } r
\right\}.
\end{equation}

All difficulty annotations are computed in an object-centric coordinate frame. Unless otherwise specified, camera centers are sampled on a radius-$3$ viewing sphere, and each camera is oriented toward the object origin. We use image resolution $512 \times 512$, field of view $45^\circ$, maximum ray range $6.0$, and ignore unknown cells during ray casting. Candidate camera centers are generated from approximately uniform Tammes sphere point sets.

For observation saturation analysis, we evaluate the following view-set cardinalities:
\begin{equation}
\begin{aligned}
\mathcal{N} =
\{&
6,8,10,12,14,16,20,24,28,32,40,48,56,64,\\
&80,96,112,128,144,160,180,200,270,360,432,492
\}.
\end{aligned}
\end{equation}

For planning-difficulty annotation, we use the fixed planning candidate-view protocol $V_{\mathrm{plan}}$ described in the main paper.

\subsection{Visibility Operators}

A key implementation choice is how to define the visibility operator between a candidate view and a discretized surface voxel. In continuous rendering, visibility is naturally defined by casting one ray per image pixel and recording the first surface hit. However, difficulty annotation requires a discrete coverage matrix between candidate views and surface voxels. Directly quantizing render-style hit points back to voxels may introduce attribution ambiguity near voxel boundaries, thin structures, and nearby occupied cells.

We distinguish three visibility notions.

\paragraph{Render-style visibility.}
Render-style visibility casts rays from camera pixels and records the first occupied hit along each ray. It is the natural model for simulating what a depth camera observes, and is therefore suitable for RGB-D rendering or traditional voxel information-gain computation. However, when the continuous hit point is quantized back to an occupied voxel, the assigned voxel can be affected by discretization boundaries or neighboring occupied cells. This makes render-style visibility less suitable as a direct voxel-level coverage matrix for set-cover annotation.

\paragraph{Strict voxel-membership visibility.}
Strict voxel-membership visibility starts from a candidate voxel, projects its center into the image, back-projects the corresponding pixel ray, and counts the candidate voxel as visible only if the first hit belongs to the same voxel. This gives a conservative and unambiguous voxel-attribution model. It is appropriate when the goal is strict voxel-level assignment, but it can be overly restrictive for observation-oriented benchmark annotation because some surface regions that would be naturally observed may be discarded by the same-voxel requirement.

\paragraph{Inverse visibility.}
ObjView-Bench uses inverse visibility for all difficulty annotations. This choice is a middle ground between render-style observation and strict voxel membership. For each occupied candidate voxel, we first project the voxel center into the camera image. If the projected pixel is valid, we back-project the corresponding camera ray and perform a first-hit ray-cast query. If the ray hits an occupied cell, the hit cell is inserted into the visible set for that view. This candidate-guided procedure fixes the query direction before ray casting, reducing boundary-induced attribution drift while retaining an observation-oriented semantics.

For a view $v$, the resulting visible set is:
\begin{equation}
O(v;r)
=
\left\{
u \in S_{\mathrm{gt}}^{(r)}
\;\middle|\;
\begin{array}{l}
u \text{ is returned as an occupied first hit} \\
\text{by an inverse ray query from } v
\end{array}
\right\}.
\end{equation}

All results reported in the paper use the CUDA implementation of this operator, denoted \texttt{inverse\_cuda}. The CPU implementation is used only for debugging and consistency checks.

\subsection{CUDA-Accelerated Ray Casting}

The inverse visibility operator is implemented using a CUDA-accelerated OctoMap ray caster. Occupied OctoMap leaf keys are first inserted into a GPU hash table. Batched ray queries are then executed on the GPU. For each query ray, the kernel traverses voxels using a 3D grid traversal procedure until one of the following conditions is met: an occupied cell is found, the ray reaches the maximum range, or the query terminates according to the unknown-cell handling rule.

Each CUDA ray query returns a hit flag and an endpoint. The endpoint is converted back to an OctoMap key and checked against the occupied tree before being inserted into the per-view visible set. This batched implementation is important because difficulty annotation requires evaluating thousands of objects over multiple view-set cardinalities and candidate-view protocols. We use the same \texttt{inverse\_cuda} visibility operator for both saturation analysis and planning-difficulty annotation, ensuring that all difficulty quantities are defined under a consistent coverage semantics.

\subsection{Observable-Surface Saturation}

For each evaluated view-set size $N \in \mathcal{N}$, we load the corresponding Tammes view set $V_N$ and compute the union of inverse-visible surface voxels:
\begin{equation}
S_N^{(r)}
=
\bigcup_{v \in V_N} O(v;r).
\end{equation}

The observable-surface cardinality at view-set size $N$ is:
\begin{equation}
Y^{(r)}(N)
=
\left|S_N^{(r)}\right|.
\end{equation}

Because Tammes sets of different cardinalities are generated independently and are not guaranteed to be nested, the raw sequence $Y^{(r)}(N)$ may contain small non-monotone fluctuations. We therefore first construct a monotone approximation by taking the prefix maximum:
\begin{equation}
\bar{Y}^{(r)}(N_i)
=
\max_{j \leq i} Y^{(r)}(N_j),
\end{equation}
where $N_i$ denotes the $i$-th evaluated view-set size in ascending order.

We then fit a lognormal cumulative distribution function to the monotone curve:
\begin{equation}
\hat{Y}^{(r)}(N)
=
Y_{\max}
\,
\Phi
\left(
\frac{\log N - \mu}{\sigma}
\right),
\end{equation}
where $\Phi(\cdot)$ is the standard normal cumulative distribution function. The fitted curve provides a smooth approximation of the saturation behavior and avoids defining the saturation point from local fluctuations in non-nested view sets.

The observable-surface ratio is computed from the fitted asymptotic observable-surface cardinality:
\begin{equation}
r_{\mathrm{vis}}^{(r)}
=
A_{\mathrm{fit}}^{(r)}
=
\frac{
Y_{\max}
}{
\left|S_{\mathrm{gt}}^{(r)}\right|
}.
\end{equation}

The observation saturation view count is defined from the first continuous view-set size at which the fitted curve enters a sustained low-gain regime. In implementation, we use a forward difference with $\Delta N=10$:
\begin{equation}
G^{(r)}(N)
=
\hat{Y}^{(r)}(N+\Delta N)
-
\hat{Y}^{(r)}(N).
\end{equation}

We require this gain to remain below a fixed fraction of the total discretized surface size over a window of $20$:
\begin{equation}
G^{(r)}(N')
<
0.001
\left|S_{\mathrm{gt}}^{(r)}\right|,
\qquad
\forall N' \in [N, N+20].
\end{equation}
In practice, this condition is checked on sampled points of the fitted curve within the forward window.

The first continuous $N$ satisfying this condition is recorded as $B_{\mathrm{cont}}^{(r)}$. The observation saturation difficulty used for dataset statistics and split construction is the ceiling of this continuous value:
\begin{equation}
d_{\mathrm{sat}}^{(r)}
=
\left\lceil
B_{\mathrm{cont}}^{(r)}
\right\rceil .
\end{equation}

For completeness, the implementation also stores a discrete reference saturation count $B_{\mathrm{disc}}^{(r)}$ and a discrete observable-surface ratio $A_{\mathrm{disc}}^{(r)}$ computed from the evaluated Tammes cardinalities. These discrete values are retained for debugging and consistency checks. The benchmark statistics reported in the paper use the fitted observable-surface ratio $A_{\mathrm{fit}}^{(r)}$ and the continuous-ceiling saturation count $\lceil B_{\mathrm{cont}}^{(r)} \rceil$.

\subsection{Planning Difficulty as Set Cover}

Planning difficulty is computed under a fixed candidate-view protocol $V_{\mathrm{plan}}$. For each candidate view $v_i \in V_{\mathrm{plan}}$, we compute the inverse-visible voxel set $O(v_i;r)$. The protocol-coverable target universe is the union of all voxels visible from at least one planning view:
\begin{equation}
U^{(r)}
=
\bigcup_{v_i \in V_{\mathrm{plan}}} O(v_i;r).
\end{equation}

The planning difficulty is then defined as the minimum number of candidate views required to cover this universe:
\begin{equation}
\begin{aligned}
d_{\mathrm{plan}}^{(r)}
=
\min_{\mathbf{x}}
\quad
&
\sum_{i=1}^{|V_{\mathrm{plan}}|}
x_i
\\
\mathrm{s.t.}
\quad
&
\sum_{i:u \in O(v_i;r)}
x_i
\geq 1,
\qquad
\forall u \in U^{(r)},
\\
&
x_i \in \{0,1\},
\qquad
i=1,\ldots,|V_{\mathrm{plan}}|.
\end{aligned}
\end{equation}

We solve this integer program using Gurobi. The number of selected views in the optimal solution is recorded as $d_{\mathrm{plan}}^{(r)}$. For debugging and analysis, we also save the universe size, selected view indices, per-view visible voxel counts, and per-view unique voxel counts.

\subsection{Annotation Outputs}

\begin{table}[t]
\centering
\caption{Difficulty annotation parameters.}
\label{tab:annotation_params}
\begin{tabular}{ll}
\toprule
Parameter & Value \\
\midrule
Surface discretization & OctoMap occupied leaf voxels \\
Voxel resolution $r$ & $0.02$ \\
Camera image size & $512 \times 512$ \\
Field of view & $45^\circ$ \\
Viewing radius & $3.0$ \\
Look-at point & Object origin $(0,0,0)$ \\
Maximum ray range & $6.0$ \\
Unknown-cell handling & Ignore unknown cells \\
Visibility mode & \texttt{inverse\_cuda} \\
Evaluated saturation view sizes & $\mathcal{N}$ in Eq.~(5) \\
Reported observable-surface ratio & $A_{\mathrm{fit}}^{(r)}$ \\
Reported saturation difficulty & $\lceil B_{\mathrm{cont}}^{(r)} \rceil$ \\
Saturation $\Delta N$ & $10$ \\
Saturation window & $20$ \\
Saturation epsilon ratio & $0.001$ \\
Saturation curve model & Lognormal CDF \\
Planning candidate protocol & $|V_{\mathrm{plan}}| = 128$ \\
Planning solver & Gurobi mixed-integer programming \\
Reported planning difficulty & Selected view count of optimal set cover \\
\bottomrule
\end{tabular}
\end{table}

For each object, the saturation annotation stores the evaluated view-set sizes, raw observable-surface counts, monotone observable-surface counts, fitted saturation-curve parameters, and threshold parameters used to determine the low-gain regime. The reported observable-surface ratio $r_{\mathrm{vis}}^{(r)}$ is the fitted asymptotic ratio $A_{\mathrm{fit}}^{(r)}$, and the reported observation saturation difficulty $d_{\mathrm{sat}}^{(r)}$ is the ceiling of the continuous saturation point:
\begin{equation}
r_{\mathrm{vis}}^{(r)}
=
A_{\mathrm{fit}}^{(r)},
\qquad
d_{\mathrm{sat}}^{(r)}
=
\left\lceil
B_{\mathrm{cont}}^{(r)}
\right\rceil .
\end{equation}
For debugging and consistency checks, the annotation also stores the discrete reference values $A_{\mathrm{disc}}^{(r)}$ and $B_{\mathrm{disc}}^{(r)}$ computed from the evaluated Tammes cardinalities.

The planning-difficulty annotation stores the protocol-coverable universe size, the selected set-cover views, and the resulting planning difficulty $d_{\mathrm{plan}}^{(r)}$. The reported $d_{\mathrm{plan}}^{(r)}$ is the number of selected views in the optimal set-cover solution:
\begin{equation}
d_{\mathrm{plan}}^{(r)}
=
\left|
\left\{
i
\;\middle|\;
x_i^\star = 1
\right\}
\right|,
\end{equation}
where $\mathbf{x}^\star$ denotes the optimal binary solution of the set-cover problem.

These quantities play different roles in ObjView-Bench. The observable-surface ratio $r_{\mathrm{vis}}^{(r)}$ estimates the externally observable fraction of the object under the canonical view protocol. The observation saturation difficulty $d_{\mathrm{sat}}^{(r)}$ identifies objects whose observable surface is revealed only under dense view-space sampling. The planning difficulty $d_{\mathrm{plan}}^{(r)}$ characterizes the number of complementary views required under the fixed planning protocol and is used for benchmark stratification and difficulty-aware analysis.

\subsection{Summary of Annotation Parameters}

Table~\ref{tab:annotation_params} summarizes the main parameters used for difficulty annotation.

\section{Dataset Construction and Difficulty Analysis}
\label{app:dataset}

\subsection{Source Pool and Quality Prefiltering}

ObjView-Bench is constructed from large-scale 3D assets, but the goal is not to maximize the raw number of objects. Instead, we aim to obtain a reliable object pool for object-centric view planning, where each retained asset is semantically meaningful as an individual object, geometrically usable for reconstruction, and compatible with the benchmark rendering and difficulty-annotation pipeline.

We start from Objaverse++ annotations and apply a quality prefiltering stage before any task-specific difficulty annotation. Objaverse++ provides quality scores and binary asset traits for Objaverse objects, including whether an asset is transparent, scene-level, single-color, not a single object, or figure-like. We use these annotations to remove assets that are unlikely to be suitable for object-centric reconstruction. In our implementation, we retain only assets with the highest quality score and require the binary flags for multi-object, scene, figure, transparency, and single-color content to be false. This quality prefiltering stage reduces the initial Objaverse++-annotated pool from approximately $800$K assets to approximately $185$K candidate object assets.

This stage reduces the original large-scale asset pool to a smaller pool of quality-filtered object assets for geometry processing. Its purpose is not to define view-planning difficulty, but to remove clearly unsuitable assets before geometry filtering, rendering standardization, and difficulty annotation. The task-specific difficulty quantities are computed only in later stages using the visibility and set-cover pipeline described in Appendix~\ref{app:difficulty}.

\subsection{Geometry Filtering and Candidate Selection}

After quality prefiltering, we perform lightweight geometry filtering and diversity-preserving candidate selection. The goal of this stage is not to certify that every mesh is geometrically perfect, but to remove clearly invalid assets and retain a diverse candidate pool for downstream difficulty annotation. We therefore use conservative filtering rules and avoid overly specific geometry heuristics that would require subjective decisions about borderline object categories.

We first compute mesh-level geometry statistics for each quality-filtered asset, including vertex and face counts, finite-vertex checks, face-index validity, winding consistency, degenerate-face ratio, bounding-box statistics, convex-hull statistics, surface area, volume, and Euler number. We then apply a hard filtering stage to remove assets with loading failures, empty meshes, non-finite vertices, invalid face indices, inconsistent winding, too few or too many vertices or faces, or an excessively high degenerate-face ratio. In our implementation, we require at least $200$ vertices and $200$ faces, allow at most $200{,}000$ vertices and $200{,}000$ faces, and remove meshes whose degenerate-face ratio exceeds $0.02$.

After hard filtering, we apply a lightweight statistical filtering stage to remove extreme geometric outliers. We use robust $z$-score filtering on three statistics:
\begin{equation}
\log r_{\mathrm{bbox\_area}},
\qquad
\log r_{\mathrm{hull\_area}},
\qquad
e_{\mathrm{norm}}
=
\frac{
e_{\mathrm{Euler}}
}{
\log(1 + n_{\mathrm{face}})
}.
\end{equation}
Here $r_{\mathrm{bbox\_area}}$ and $r_{\mathrm{hull\_area}}$ are area-ratio statistics derived from the mesh and its bounding box or convex hull, $e_{\mathrm{Euler}}$ is the Euler number, and $n_{\mathrm{face}}$ is the number of faces. The normalized Euler statistic is used only as an extreme-outlier signal rather than as a strict topological validity criterion. We use robust filtering thresholds $\tau_{\mathrm{area}}=2.0$, $\tau_{\mathrm{hull}}=2.0$, and $\tau_{\mathrm{Euler}}=5.0$.

To preserve shape diversity, we assign each remaining object to a lightweight geometry stratum. We use two stratification variables: a four-way shape type and a three-way fill bucket. The shape type is determined from elongation and planarity statistics:
\begin{equation}
\mathrm{shape\_type}
\in
\{
\mathrm{compact},
\mathrm{elongated},
\mathrm{flat\_or\_sheet\_like},
\mathrm{bladed}
\}.
\end{equation}
Objects with high elongation and low planarity are labeled as elongated; objects with low elongation and high planarity are labeled as flat or sheet-like; objects with both high elongation and high planarity are labeled as bladed; and the remaining objects are labeled as compact. We use thresholds $1.8$ for both elongation and planarity. The fill bucket is obtained by dividing the fill-ratio distribution into three buckets:
\begin{equation}
\mathrm{fill\_bucket}
\in
\{
\mathrm{low},
\mathrm{mid},
\mathrm{high}
\}.
\end{equation}

The Cartesian product of these two variables gives $4 \times 3 = 12$ strata. We sample from these strata to obtain an approximately $12$K-object candidate pool for difficulty annotation. This stratified selection is intended to preserve broad variation in object shape and fill structure, while keeping the candidate pool computationally manageable for the expensive visibility and set-cover annotation stages. Some candidates are later removed because GLB downloading, point-cloud sampling, OBJ conversion, rendering standardization, or manual reliability review may fail.

\subsection{Geometry Normalization and Rendering Standardization}

After candidate selection, we standardize the geometry representation used by the benchmark. This stage serves two purposes. First, it provides a common object-centric coordinate frame for difficulty annotation and evaluation. Second, it ensures that the assets retained after filtering are compatible with the rendering pipeline used by the online benchmark.

For each candidate object, we sample a normalized surface point cloud from the original GLB asset. We use Poisson-disk surface sampling and compute the normalization transform from the sampled points. Let $\mathcal{P}_{\mathrm{raw}}$ denote the sampled raw point cloud. We compute the translation center as the point-cloud mean:
\begin{equation}
c
=
\frac{1}{|\mathcal{P}_{\mathrm{raw}}|}
\sum_{p \in \mathcal{P}_{\mathrm{raw}}} p,
\end{equation}
and the scale as the maximum distance from this center:
\begin{equation}
s
=
\max_{p \in \mathcal{P}_{\mathrm{raw}}}
\left\|
p - c
\right\|_2 .
\end{equation}
The normalized point cloud is then obtained as:
\begin{equation}
p_{\mathrm{norm}}
=
\frac{
p_{\mathrm{raw}} - c
}{
s
}.
\end{equation}
The same normalization parameters are later reused for the mesh representation, ensuring that point clouds, OBJ meshes, and benchmark camera poses are defined in the same object-centric frame.

In addition to normalized point clouds, we convert each retained GLB asset into a standardized OBJ representation for rendering and release. During this conversion, mesh components are merged into a single object-level mesh, transforms are applied, faces are triangulated, a unified UV atlas is generated, and the material appearance is baked into a single texture. The resulting OBJ, MTL, and texture files provide a stable input representation for PyTorch3D rendering and for manual reliability review. In our implementation, this conversion is performed with Blender 4.2.7.

Before manual review, we render fixed-view previews for each standardized object using the same PyTorch3D-based rendering pipeline as the benchmark evaluator. We generate four-view previews using image resolution $512 \times 512$, camera distance $3.0$, and field of view $45^\circ$. These previews expose rendering incompatibilities and visual abnormalities under the actual benchmark renderer, rather than under an unrelated visualization tool.

This standardization stage is intentionally placed after the initial geometry filtering and candidate selection. The earlier stages depend mainly on geometry statistics and visibility annotation, while OBJ conversion and texture baking are more expensive and can fail for a small fraction of assets. Therefore, failures in GLB downloading, point-cloud sampling, OBJ conversion, normalization, or preview rendering are handled conservatively by removing the affected candidates from later benchmark construction.

\subsection{Difficulty Annotation and Pool Separation}

After geometry standardization, we run the difficulty annotation pipeline described in Appendix~\ref{app:difficulty} on the candidate objects. For each object, this produces three task-specific quantities: the omnidirectional self-occlusion attribute $r_{\mathrm{vis}}^{(r)}$, the observation saturation difficulty $d_{\mathrm{sat}}^{(r)}$, and the planning difficulty $d_{\mathrm{plan}}^{(r)}$. In the implementation, the released metadata records these quantities as the fitted self-occlusion attribute, the continuous-ceiling saturation view count, and the selected-view count of the optimal set-cover solution, respectively.

We use these quantities to separate object-side saturation behavior from protocol-dependent planning difficulty. The observation saturation difficulty $d_{\mathrm{sat}}^{(r)}$ measures how densely the external view space must be sampled before the observable surface enters a low-gain regime. In contrast, $d_{\mathrm{plan}}^{(r)}$ measures how many complementary views are needed to cover the protocol-coverable target universe under the fixed planning candidate set. These two quantities are related but not equivalent, so we do not collapse them into a single hard-object score.

The boundary between the main pool and the slow-saturation diagnostic pool is determined from the saturation statistics rather than assumed a priori. As shown by the marginal-gain behavior in the main paper, the observable-surface gain becomes small around the 128-view regime for the majority of candidate objects. We therefore use $d_{\mathrm{sat}}^{(r)} \leq 128$ as the criterion for the main pool: these objects reach the low-gain saturation regime within the view density used to instantiate the fixed planning protocol. Retained objects whose saturation difficulty exceeds this threshold are kept separate as slow-saturation diagnostic samples:
\begin{equation}
\begin{cases}
\mathrm{main}, & d_{\mathrm{sat}}^{(r)} \leq 128, \\
\mathrm{slow\mbox{-}saturation}, & d_{\mathrm{sat}}^{(r)} > 128.
\end{cases}
\end{equation}

Before finalizing this split, we manually inspect the high-$d_{\mathrm{sat}}^{(r)}$ tail. This inspection revealed two qualitatively different cases. Many objects exhibit meaningful slow-saturation behavior, such as deep cavities, aperture-mediated internal surfaces, or inter-part occlusions that require specific viewing directions to reveal. These objects are retained as valid slow-saturation diagnostic samples. However, some objects with extremely large saturation estimates are dominated by asset or discretization artifacts rather than meaningful view-planning difficulty. Typical failure modes include very thin surfaces, problematic mesh faces, open sheet-like structures, or local geometry that allows ray-casting queries to penetrate through the intended surface. Such artifacts can create artificially persistent low-gain observations and inflate $d_{\mathrm{sat}}^{(r)}$ for reasons unrelated to object-centric view planning.

We therefore remove these unreliable extreme cases during the reliability review before constructing the final pools. In practice, the retained slow-saturation pool consists of high-saturation objects with stable visibility annotations, while artifact-dominated high-saturation cases are excluded. This exclusion is a conservative reliability filter rather than a removal of hard object categories: thin, elongated, or geometrically challenging objects are still retained in the main or slow-saturation pools when their annotations are stable. The main pool is organized primarily by $d_{\mathrm{plan}}^{(r)}$ for benchmark stratification, while the slow-saturation pool is kept separate as a diagnostic split for long-tail reveal behavior.

\subsection{Manual Reliability Review}

After automatic filtering, geometry standardization, and difficulty annotation, we perform a manual reliability review before constructing the final benchmark pools. The purpose of this step is not to subjectively select visually appealing objects, but to conservatively remove boundary cases that can undermine the reliability of object-centric reconstruction evaluation or difficulty annotation.

Each candidate object is reviewed using the standardized OBJ representation and fixed-view previews rendered with the same PyTorch3D-based pipeline used by the benchmark evaluator. This ensures that the review is performed on the representation that will be used during evaluation, rather than on the original asset preview or an unrelated renderer. The review considers both object validity and benchmark reliability. In particular, we remove objects that are not suitable as single object-centric reconstruction targets, objects that cannot be rendered reliably, and objects whose difficulty estimates are likely dominated by geometry-processing artifacts.

The main rejection categories are:
\begin{equation}
\begin{aligned}
&\text{rendering error}, \\
&\text{multi-object or scene-level asset}, \\
&\text{figure-like, transparent, or single-color asset}, \\
&\text{single-layer sheet-like surface}, \\
&\text{malformed or semantically invalid structure}, \\
&\text{unreliable difficulty signal due to geometry or discretization artifacts}, \\
&\text{other boundary cases}.
\end{aligned}
\end{equation}

To make the review efficient and consistent, we attach risk hints to each candidate using the automatic geometry and difficulty statistics. These hints do not directly determine whether an object is retained, but they help prioritize manual inspection and identify cases that require conservative judgment. Examples include very small ground-truth surface voxel counts, planning difficulty close to the candidate-view limit, large mismatch between saturation difficulty and planning difficulty, and high-saturation objects whose geometry suggests possible ray-casting or voxelization artifacts.

We merge multiple review passes conservatively. If any review marks an object as rejected, the object is removed. If no rejection is present but at least one review marks the object as uncertain, the object is not included in the final retained pool. Only objects that pass the review are used to construct the final clean main pool and slow-saturation diagnostic pool. This conservative policy improves the consistency between the benchmark difficulty semantics, the rendering pipeline, and the final evaluation protocol.

\subsection{Benchmark Splits and Released Metadata}

After manual reliability review, the retained benchmark pool contains $5{,}691$ objects. Among them, $5{,}523$ objects form the final main pool and $168$ objects form the slow-saturation diagnostic pool. The main pool is used for the primary training and hidden-test benchmark, while the slow-saturation pool is released as a separate diagnostic subset for analyzing slow detail reveal behavior. The manual review removes a substantial number of candidates because the upstream Objaverse++ quality and attribute annotations are not perfect, and because some assets suffer from geometry, material, or color degradation during GLB-to-OBJ conversion, texture baking, or PyTorch3D rendering. We therefore treat manual review as a conservative reliability step to ensure that the final benchmark objects are valid, renderable, and associated with stable difficulty annotations.

For the released main benchmark, we construct a public training pool and a hidden test set. The hidden test set is sampled from the final main pool using a planning-difficulty-aware rule. Let $n_c$ denote the number of candidate objects in planning-difficulty bucket $c$, let $N$ denote the total number of hidden-test candidates, and let $T=500$ denote the target hidden-test size. For each non-empty planning-difficulty bucket, we allocate
\begin{equation}
q_c
=
\max
\left(
1,
\mathrm{round}
\left(
\frac{n_c}{N} T
\right)
\right)
\end{equation}
hidden-test slots. We then randomly sample objects within each bucket using a fixed random seed. This rule approximately preserves the natural planning-difficulty distribution while ensuring that medium- and high-difficulty buckets are not omitted from the hidden test set. Because each non-empty bucket receives at least one sample, the final hidden test set contains $528$ objects rather than exactly $500$.

Objects used for paper visualization or qualitative showcase are excluded from hidden-test sampling. After hidden-test construction, all remaining main-pool objects form Main-Train-Public. This yields $4{,}995$ public training objects and $528$ hidden-test objects:
\begin{equation}
5{,}523
=
4{,}995
+
528.
\end{equation}

\begin{table}[t]
\centering
\caption{Dataset construction funnel.}
\label{tab:dataset_funnel}
\begin{tabular}{lrp{7cm}}
\toprule
Stage & Objects & Description \\
\midrule
Objaverse++ annotated assets & $\sim$800K & Initial source pool \\
Quality-prefiltered assets & $\sim$185K & High-quality, single-object, non-scene, non-transparent assets \\
Geometry-filtered candidates & 12K & Conservative geometry filtering and 12-strata candidate selection \\
Final retained benchmark pool & 5,691 & After rendering standardization and manual reliability review with difficulty annotation \\
Main pool & 5,523 & Used to form Main-Train-Public and Main-Hidden-Test \\
Slow-saturation pool & 168 & Released diagnostic subset \\
\bottomrule
\end{tabular}
\end{table}

\begin{table}[t]
\centering
\caption{ObjView-Bench splits and their roles.}
\label{tab:dataset_splits}
\begin{tabular}{lrlp{6cm}}
\toprule
Split & Objects & Public & Role \\
\midrule
Main-Train-Public & 4,995 & Yes & Public training pool for main benchmark \\
Main-Hidden-Test & 528 & No & Server-side hidden test for leaderboard evaluation \\
Slow-Saturation & 168 & Yes & Diagnostic subset for long-tail reveal behavior \\
Main-Analysis-Test & 245 & No & Held-out split for difficulty-aware training analysis \\
Main-Raw-Train & 1,014 & No & Natural-distribution training subset \\
Main-Balanced-Train & 1,014 & No & Difficulty-balanced training subset \\
\bottomrule
\end{tabular}
\end{table}

The hidden-test object identities and geometries are kept server-side for online evaluation and leaderboard use. They are not included in the public release. The public release contains Main-Train-Public metadata and geometry, together with the slow-saturation diagnostic pool. Each released metadata record includes the object identifier, normalization center and scale, the omnidirectional self-occlusion attribute, the observation saturation difficulty, and the planning difficulty.

We also construct separate analysis splits for the difficulty-aware training experiment in Section~\ref{sec:experiment}. These analysis splits are not part of the released leaderboard protocol. They are used only to test whether $d_{\mathrm{plan}}^{(r)}$ can serve as an operational training-distribution axis.

We first construct Main-Analysis-Test from the main pool by round-robin sampling over non-empty $d_{\mathrm{plan}}^{(r)}$ buckets. Specifically, each round samples at most one object from each non-empty bucket, and we use three rounds to obtain $245$ held-out analysis-test objects. After removing these analysis-test objects, we construct two equal-size training subsets from the remaining main-pool objects. Main-Balanced-Train is sampled by the same round-robin strategy over non-empty $d_{\mathrm{plan}}^{(r)}$ buckets, increasing exposure to medium- and high-difficulty objects; this yields $1{,}014$ training objects. Main-Raw-Train is then sampled with the same size, $1{,}014$ objects, while approximately preserving the natural planning-difficulty distribution of the remaining main pool. The comparison between these two training subsets isolates the effect of difficulty-aware sampling while keeping the training-set size fixed.

\subsection{Summary Statistics}

For convenience, we summarize the dataset construction funnel and the final benchmark splits in Tables~\ref{tab:dataset_funnel} and~\ref{tab:dataset_splits}.

\subsection{Dataset Analysis}

\begin{figure*}[t]
  \centering
  \begin{tabular}{cc}
    \includegraphics[width=0.48\textwidth]{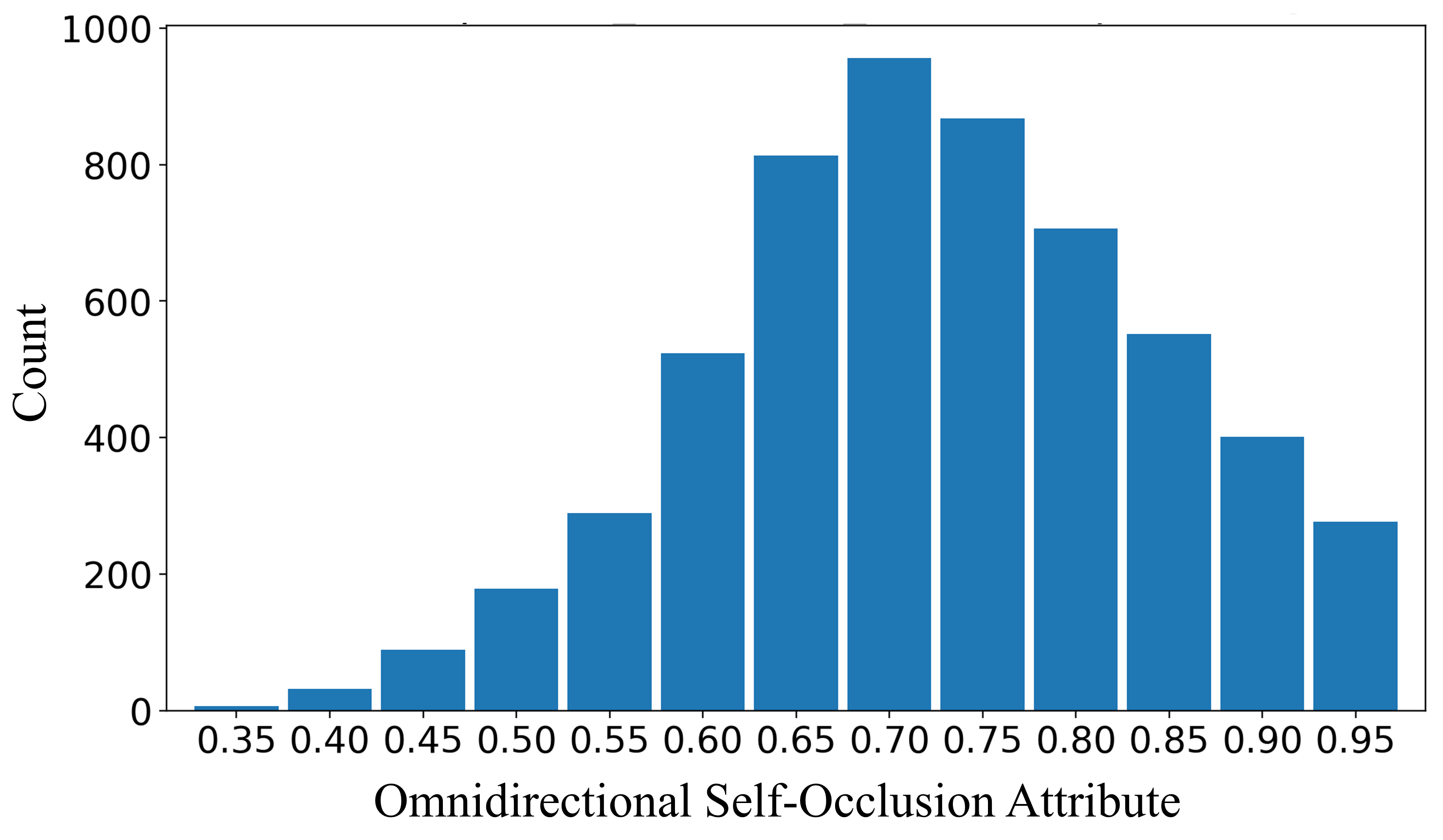} &
    \includegraphics[width=0.48\textwidth]{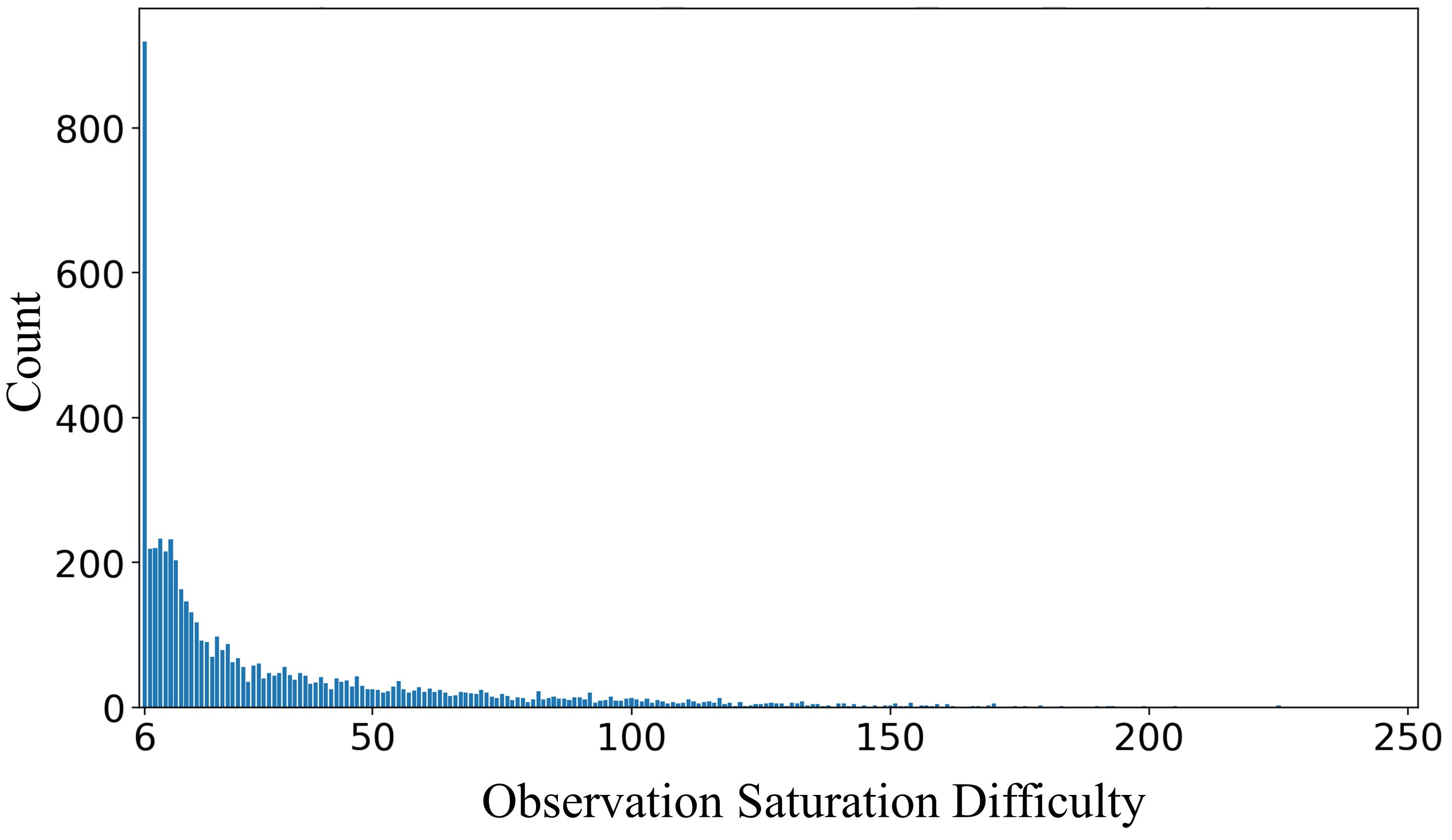} \\
    \small (a) Omnidirectional self-occlusion attribute &
    \small (b) Observation saturation difficulty
  \end{tabular}

  \caption{
  Distributions of the omnidirectional self-occlusion attribute $r_{\mathrm{vis}}^{(r)}$ and the observation saturation difficulty $d_{\mathrm{sat}}^{(r)}$ on the final retained benchmark pool, including both the main pool and the slow-saturation diagnostic pool.
  The planning-difficulty distribution is shown in the main paper, since $d_{\mathrm{plan}}^{(r)}$ is the primary difficulty axis for the main benchmark.
  }
  \label{fig:app_dataset_ab_distributions}
\end{figure*}

\begin{figure*}[t]
  \centering
  \begin{tabular}{cc}
    \includegraphics[width=0.45\textwidth]{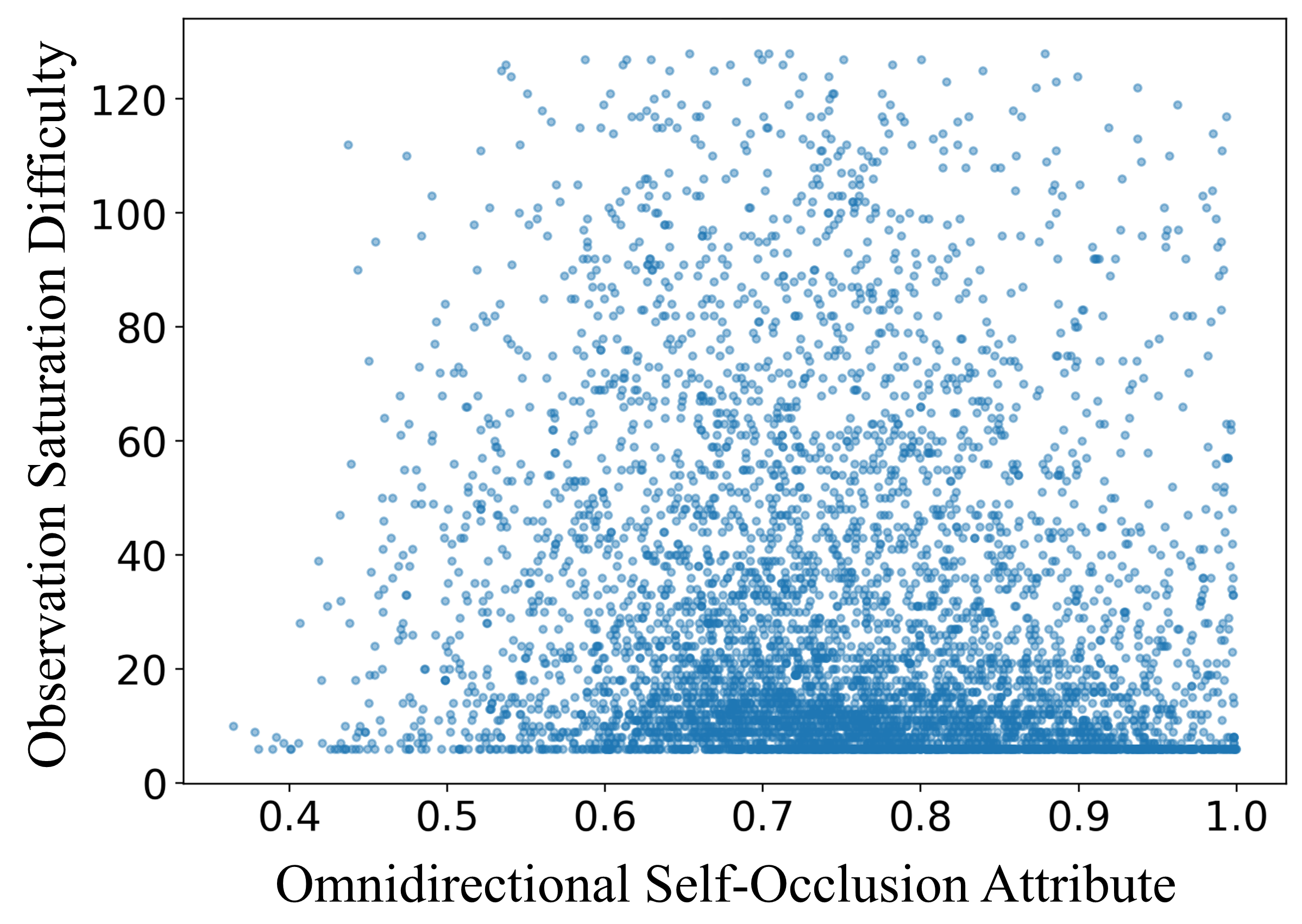} &
    \includegraphics[width=0.45\textwidth]{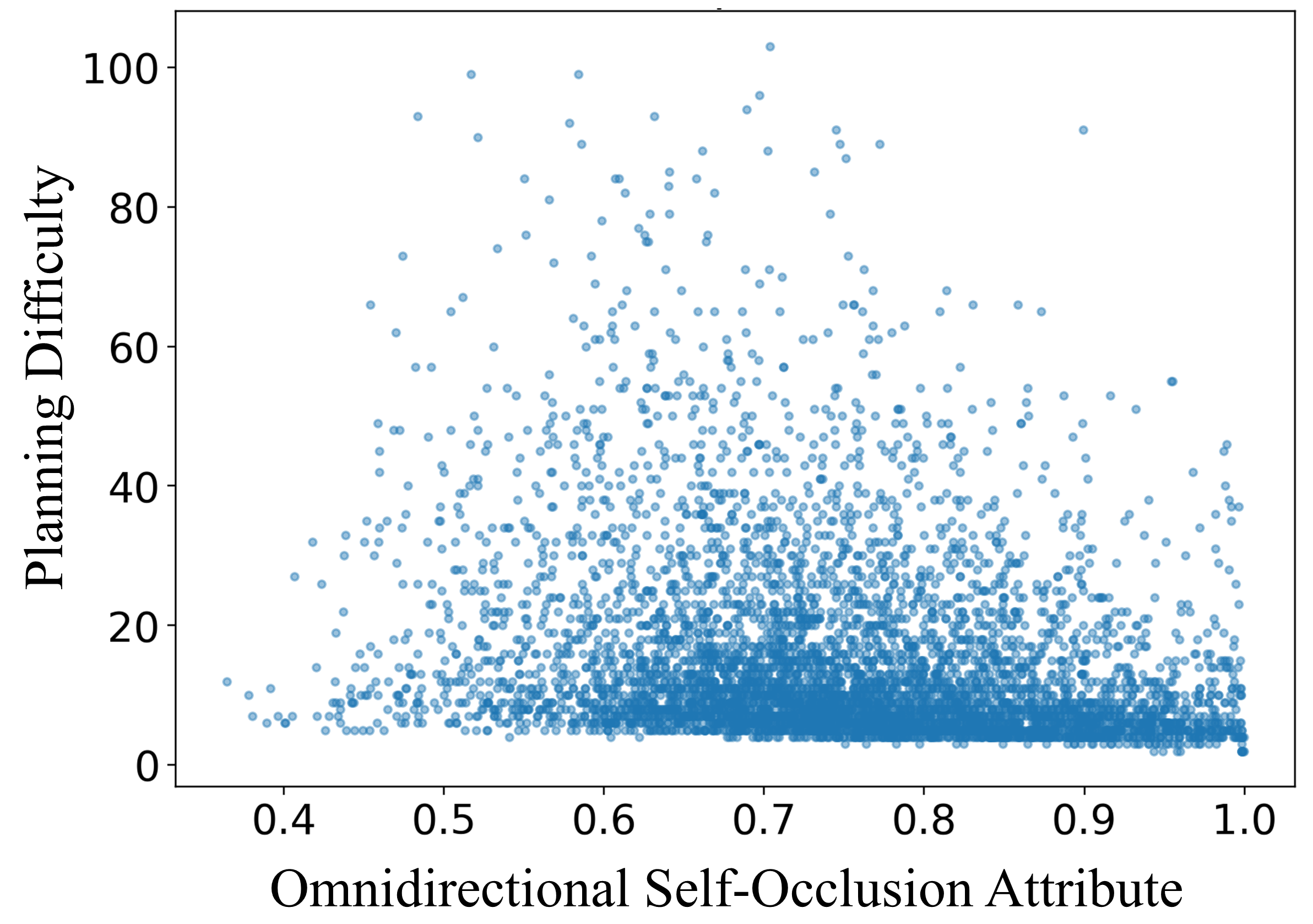} \\
    \small (a) $r_{\mathrm{vis}}^{(r)}$ vs. $d_{\mathrm{sat}}^{(r)}$ &
    \small (b) $r_{\mathrm{vis}}^{(r)}$ vs. $d_{\mathrm{plan}}^{(r)}$
  \end{tabular}

  \begin{tabular}{c}
    \includegraphics[width=0.35\textwidth]{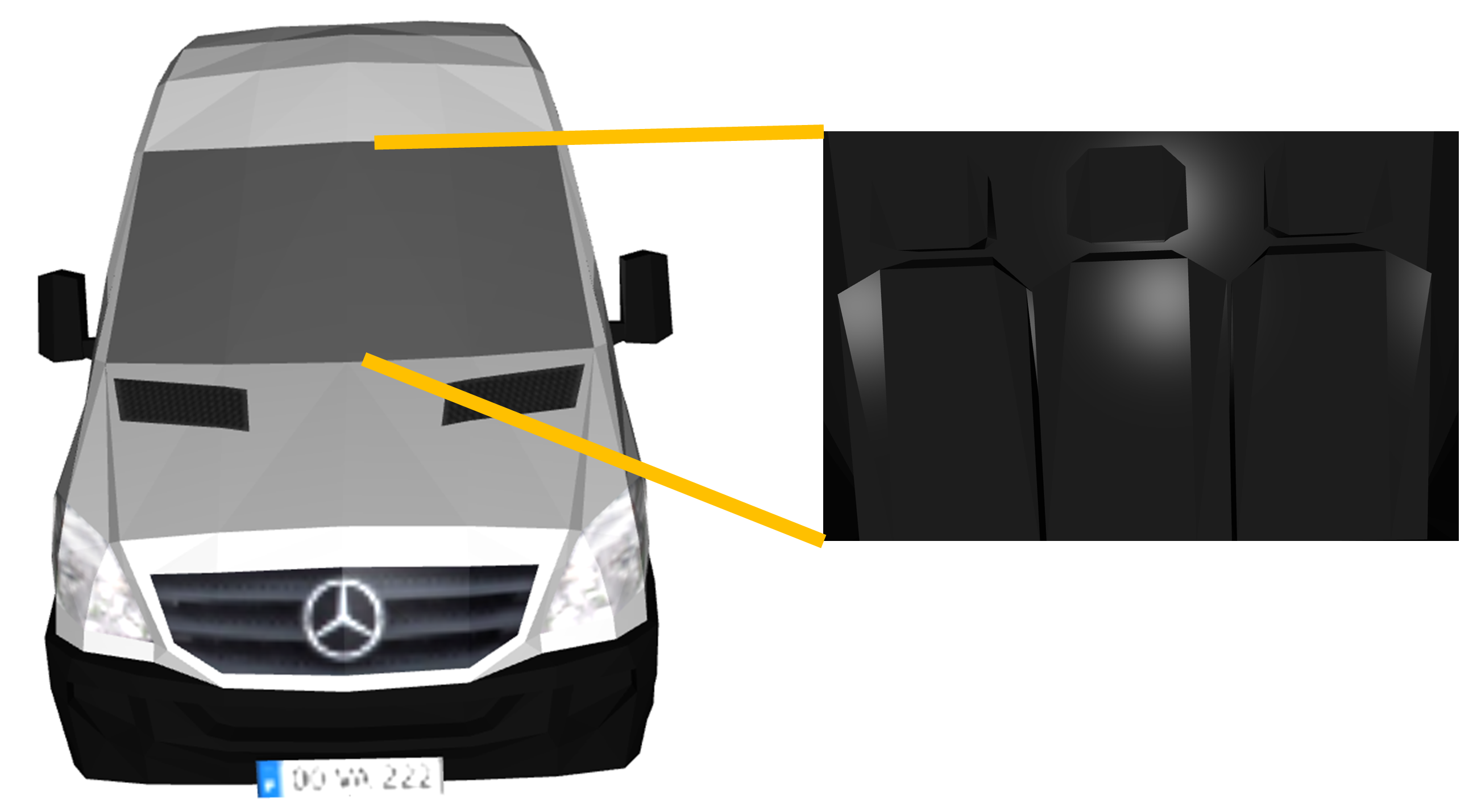} \\
    \small (c) Example of externally unobservable interior surfaces
  \end{tabular}

  \caption{
  The omnidirectional self-occlusion attribute is an object-side descriptive attribute rather than a direct proxy for task-specific difficulty.
  (a,b) Relationships between $r_{\mathrm{vis}}^{(r)}$ and the two task-specific difficulty measures on the main pool.
  Objects with similar $r_{\mathrm{vis}}^{(r)}$ can still have substantially different observation saturation difficulty or planning difficulty.
  (c) Illustrative example in which interior seat surfaces exist in the mesh geometry but are occluded from the external viewing sphere, reducing the externally observable surface ratio before any planning protocol is considered.
  }
  \label{fig:app_self_occlusion_analysis}
\end{figure*}

\subsubsection{Distributions of Omnidirectional Self-Occlusion Attribute and Saturation Difficulty}

We provide additional statistics on the final retained benchmark pool in Figure~\ref{fig:app_dataset_ab_distributions}. 
The omnidirectional self-occlusion attribute $r_{\mathrm{vis}}^{(r)}$ spans a broad range of externally observable surface ratios, indicating that the retained objects include varying degrees of object-side self-occlusion. 
Although \(r_{\mathrm{vis}}^{(r)}\) is related to the observable-surface ceiling, it should not be interpreted as a direct numerical predictor of SC@0.02. 
The self-occlusion attribute is computed in the voxelized difficulty-annotation space using inverse visibility over OctoMap surface voxels at resolution \(r=0.02\). 
In contrast, SC@0.02 is an evaluation metric computed in point-cloud space by measuring nearest-neighbor coverage of sampled ground-truth surface points within a Euclidean tolerance. 
Voxelization, point sampling, fitted saturation, rendering, and the tolerance-based point-cloud coverage rule can therefore lead to small but systematic differences between the distribution of \(r_{\mathrm{vis}}^{(r)}\) and the observed SC@0.02 values. 
We use \(r_{\mathrm{vis}}^{(r)}\) as an object-side visibility attribute for difficulty characterization, rather than as the exact denominator of the evaluation metric.

The observation saturation difficulty $d_{\mathrm{sat}}^{(r)}$ is strongly long-tailed: most objects enter the low-gain saturation regime with relatively few views, while a smaller set of objects continues to reveal observable surface regions under denser view-space sampling.
The large mass at $d_{\mathrm{sat}}^{(r)}=6$ reflects the minimum evaluated Tammes cardinality in our saturation protocol: objects that already satisfy the low-gain criterion at or below the smallest evaluated view-set size are grouped into this first bin.
The planning-difficulty distribution is shown in the main paper because $d_{\mathrm{plan}}^{(r)}$ is the primary difficulty axis for the main benchmark.
A shape-stratified check shows that compact objects tend to have a heavier \(d_{\mathrm{plan}}^{(r)}\) tail, whereas bladed objects are more concentrated at lower planning difficulty. This observation is not used to define difficulty; all benchmark stratification uses the fixed-protocol set-cover \(d_{\mathrm{plan}}^{(r)}\).

\subsubsection{Omnidirectional Self-Occlusion Is Not Task Difficulty}

Figure~\ref{fig:app_self_occlusion_analysis} further clarifies the role of the omnidirectional self-occlusion attribute. 
Although $r_{\mathrm{vis}}^{(r)}$ is useful for interpreting raw surface coverage and defining an externally observable surface reference, it is not a direct proxy for either observation saturation difficulty or planning difficulty. 
Objects with similar $r_{\mathrm{vis}}^{(r)}$ can have substantially different $d_{\mathrm{sat}}^{(r)}$ or $d_{\mathrm{plan}}^{(r)}$, because saturation and planning difficulty depend on how observable regions are distributed across views and how complementary the candidate views are. 
The vehicle example illustrates the geometric meaning of $r_{\mathrm{vis}}^{(r)}$: interior seat surfaces exist in the mesh geometry, but are occluded from the external viewing sphere and therefore reduce the externally observable surface ratio before any planning protocol is considered.

\begin{figure*}[t]
  \centering
  \includegraphics[width=0.95\textwidth]{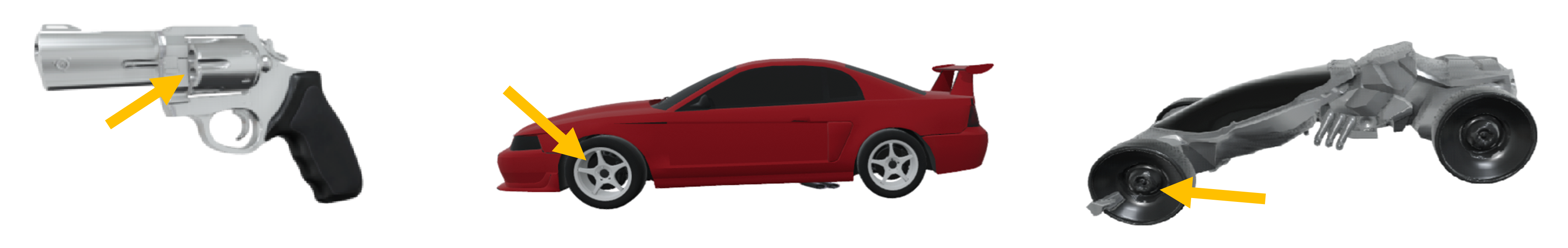}

\caption{
Qualitative examples illustrating that observation saturation difficulty $d_{\mathrm{sat}}^{(r)}$ and planning difficulty $d_{\mathrm{plan}}^{(r)}$ are related but not interchangeable.
Left: a revolver-like object with a localized shallow cavity near the cylinder gap has high saturation difficulty ($d_{\mathrm{sat}}^{(r)}=90$) but relatively low planning difficulty ($d_{\mathrm{plan}}^{(r)}=18$), because the hard-to-reveal region is spatially localized.
Middle: a car-like object with wheel-well and underbody openings has both high saturation difficulty and high planning difficulty ($d_{\mathrm{sat}}^{(r)}=80$, $d_{\mathrm{plan}}^{(r)}=46$): some surfaces are revealed only through restricted openings, while the vehicle also requires multiple complementary exterior views.
Right: a wheel-supported object has moderate saturation difficulty but high planning difficulty ($d_{\mathrm{sat}}^{(r)}=46$, $d_{\mathrm{plan}}^{(r)}=45$). 
Unlike the middle example, the difficulty is not mainly caused by looking through an aperture into hidden interior surfaces; instead, its solid wheel and body surfaces are distributed across multiple viewing directions, requiring many complementary views under the fixed planning protocol.
}
  \label{fig:app_b_c_decoupling_examples}
\end{figure*}

\subsubsection{Decoupling Saturation and Planning Difficulty}

Figure~\ref{fig:app_b_c_decoupling_examples} provides qualitative examples of the distinction between observation saturation difficulty and planning difficulty. 
Observation saturation difficulty captures how slowly observable surfaces are revealed as the view space becomes denser, whereas planning difficulty captures how many complementary views are required under a fixed candidate-view protocol. 
The examples show several ways in which these quantities can decouple. 
A localized shallow cavity can increase $d_{\mathrm{sat}}^{(r)}$ by creating a hard-to-reveal surface region, while still requiring relatively few complementary views once that region is visible. 
Conversely, an object whose coverable surfaces are distributed across many directions can require many complementary views, leading to high $d_{\mathrm{plan}}^{(r)}$, even when its saturation difficulty is only moderate. 
Objects with wheel wells, underbody openings, or other aperture-mediated structures can be high along both axes, whereas objects with solid surfaces distributed across many directions may have high planning difficulty without equally high saturation difficulty.
These examples complement the population-level $d_{\mathrm{sat}}^{(r)}$--$d_{\mathrm{plan}}^{(r)}$ analysis in the main paper and motivate treating the slow-saturation pool as a diagnostic subset rather than merging it into the main planning-difficulty axis.

Overall, these additional analyses characterize both the retained object pool and the difficulty-design choices used in ObjView-Bench.
At the dataset level, the benchmark contains objects with diverse externally observable surface ratios and a long-tailed saturation profile, indicating substantial variation in object-side self-occlusion and observable-surface discovery behavior.
The comparisons among $r_{\mathrm{vis}}^{(r)}$, $d_{\mathrm{sat}}^{(r)}$, and $d_{\mathrm{plan}}^{(r)}$ further show that these quantities capture related but distinct aspects of the problem: omnidirectional visibility describes an object-side surface property, saturation difficulty describes how slowly observable regions are revealed under denser view sampling, and planning difficulty measures the number of complementary views required under a fixed candidate-view protocol.

\subsection{Difficulty and Protocol Analysis}

\subsubsection{Resolution Sensitivity of Saturation Behavior}

\begin{figure*}[t]
  \centering
  \includegraphics[width=0.95\textwidth]{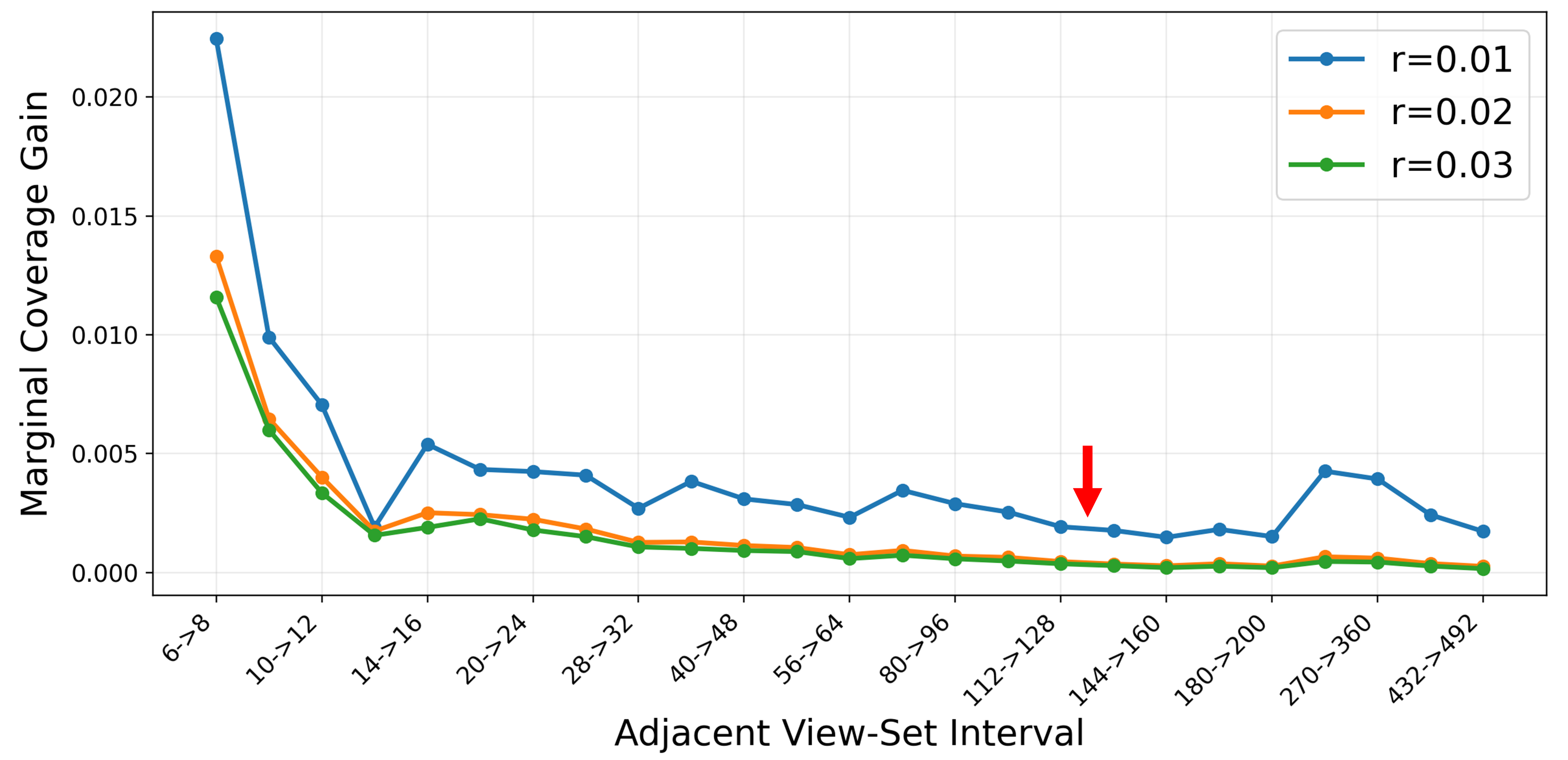}
\caption{
Resolution sensitivity of observation saturation behavior over the final retained benchmark pool objects.
We report the mean marginal normalized coverage gain between adjacent Tammes view-set sizes under three surface discretization resolutions.
The red arrow marks the canonical 128-view protocol.
Finer resolution produces larger residual fluctuations because it preserves more small-scale surface details, but all resolutions show rapidly diminishing gains and low marginal improvement around the 128-view protocol.
Later bumps occur over wider view-count intervals and reflect small persistent contributions from tail objects rather than a change in the overall saturation trend.
}
  \label{fig:app_marginal_gain_curves_by_resolution}
\end{figure*}

We further examine whether the saturation-based motivation for the canonical 128-view protocol depends on the surface discretization resolution.
In this analysis, we vary only the voxel resolution $r$ and keep all other annotation settings unchanged, including the Tammes view-set sequence, camera radius, field of view, image resolution, visibility operator, and saturation-fitting procedure.
This analysis focuses on observation saturation rather than planning difficulty.
We do not expect the absolute fitted saturation count $d_{\mathrm{sat}}^{(r)}$ to be invariant to the voxel resolution, because it is defined over a discretized observable-surface discovery process.
Changing $r$ changes which fine-scale surface regions are represented and how small residual observable regions contribute to the fitted saturation behavior.

For this reason, the key quantity in this analysis is not the absolute value, ranking, or thresholded membership induced by $d_{\mathrm{sat}}^{(r)}$, but the marginal normalized coverage gain as the view-set size increases.
Figure~\ref{fig:app_marginal_gain_curves_by_resolution} shows that the marginal gain decreases rapidly under all three resolutions and remains low around the 128-view protocol over the final retained benchmark pool objects.
At Tammes-128, the mean normalized coverage reaches 98.11\%, 99.68\%, and 99.77\% for $r=0.01$, $0.02$, and $0.03$, respectively.
The corresponding marginal gains from Tammes-112 to Tammes-128 are only 0.192, 0.046, and 0.037 percentage points, respectively.
Thus, all three resolutions have already entered a low-gain regime around the canonical 128-view protocol.
The finest resolution $r=0.01$ exhibits larger residual fluctuations, as expected, since it exposes more fine-scale and low-probability visible surface regions.
The small bumps at larger view-set sizes should also be interpreted together with the non-uniform spacing of the evaluated Tammes cardinalities: later intervals span more additional views, so they can accumulate small residual gains from tail objects that continue to reveal minor surface regions under denser sampling.
This behavior is consistent with the purpose of the slow-saturation diagnostic pool, rather than evidence that the main saturation trend changes.

This resolution sensitivity also motivates our choice of $r=0.02$ as the default annotation resolution.
A finer resolution such as $r=0.01$ preserves more small-scale surface elements, but also introduces stronger residual fluctuations and substantially higher annotation cost.
At this scale, OctoMap ray casting can become more sensitive to voxel-level leakage: tiny gaps, thin structures, or non-watertight mesh artifacts may be discretized as open visibility channels.
As a result, rays can occasionally penetrate regions that should be effectively occluded at the object scale, producing low-probability visible surface fragments and stronger residual fluctuations.
Conversely, a coarser resolution such as $r=0.03$ is more stable but can merge or suppress fine geometric structures that are relevant for object-centric reconstruction.
We therefore use $r=0.02$ as a middle-ground setting that preserves object-level geometric detail while keeping the visibility annotation stable and computationally tractable.

These results support Tammes-128 as a practical candidate-view density across the tested surface resolutions.
The saturation analysis is used to motivate the candidate-view density and to identify slow-saturation tail cases as diagnostic objects.
It is not used as the primary stratification axis of the main benchmark.
All main benchmark stratification and planning-difficulty analysis are based on the fixed-protocol set-cover difficulty $d_{\mathrm{plan}}^{(r)}$ under the documented annotation setting.

\begin{figure*}[t]
  \centering
  \includegraphics[width=0.65\textwidth]{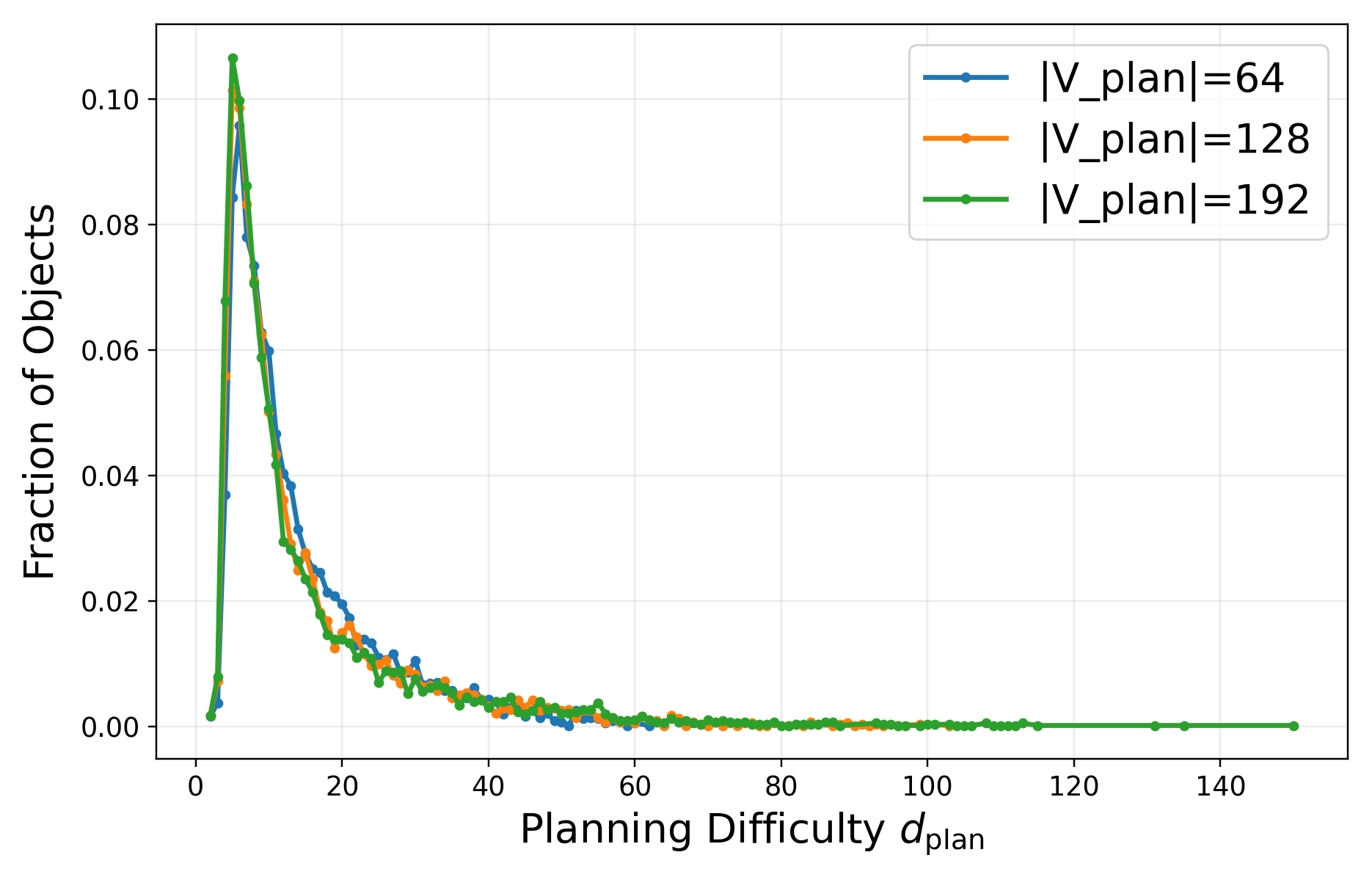}
\caption{
Candidate-view protocol sensitivity of planning difficulty over the main pool objects.
We fix the annotation resolution at $r=0.02$ and vary only the number of Tammes candidate views used to compute the fixed-protocol set-cover difficulty $d_{\mathrm{plan}}^{(r)}$.
The distributions under $|V_{\mathrm{plan}}|=64,128,192$ have similar dominant distributional structures, while denser candidate-view protocols mainly extend the extreme high-difficulty tail.
This indicates that the canonical Tammes-128 protocol preserves the overall planning-difficulty structure while providing a practical trade-off between difficulty resolution and annotation cost.
}
  \label{fig:app_planning_protocol_d_plan_distribution}
\end{figure*}

\begin{table*}[t]
\centering
\caption{
Candidate-view protocol sensitivity of planning difficulty over the main pool objects.
We fix the annotation resolution at $r=0.02$ and vary only the number of Tammes candidate views used to compute $d_{\mathrm{plan}}^{(r)}$.
All rank-based quantities are computed with respect to the canonical Tammes-128 protocol.
Rank displacement is normalized by $N-1$.
}
\label{tab:app_planning_protocol_rank_stability}
\begin{tabular}{lcccc}
\toprule
Comparison & Spearman $\rho$ & Kendall $\tau_b$ & Median $\Delta$rank & Top-20\% overlap \\
\midrule
$|V_{\mathrm{plan}}|=64$ vs. $128$ & 0.9853 & 0.9266 & 3.44\% & 90.59\% \\
$|V_{\mathrm{plan}}|=192$ vs. $128$ & 0.9911 & 0.9508 & 1.92\% & 93.48\% \\
\bottomrule
\end{tabular}
\end{table*}

\subsubsection{Candidate-View Protocol Sensitivity of Planning Difficulty}

Having fixed $r=0.02$ as the default annotation resolution, we next examine the other key protocol choice used by the planning-difficulty definition: the density of the candidate-view set.
Specifically, we keep the voxel resolution, camera model, visibility operator, and set-cover solver unchanged, and vary only the number of Tammes candidate views used to compute $d_{\mathrm{plan}}^{(r)}$.

Figure~\ref{fig:app_planning_protocol_d_plan_distribution} shows that the distributions of $d_{\mathrm{plan}}^{(r)}$ remain highly similar across $|V_{\mathrm{plan}}|=64,128,192$ over the main pool objects.
The main mass of the distribution is nearly unchanged, while denser candidate-view protocols mainly extend the extreme high-difficulty tail.
Consistent with this pattern, the mean $d_{\mathrm{plan}}^{(r)}$ changes only moderately from 14.27 under the 64-view protocol to 15.06 and 15.47 under the 128-view and 192-view protocols, respectively.
This behavior is expected: a denser candidate set can distinguish more fine-grained complementary-view requirements for a small number of hard objects, but does not substantially change the overall planning-difficulty structure.

Table~\ref{tab:app_planning_protocol_rank_stability} further confirms this trend at the object level.
We report complementary rank-stability measures with respect to the canonical Tammes-128 protocol.
Spearman's $\rho$ measures global monotonic agreement between the two difficulty rankings, while Kendall's $\tau_b$ measures pairwise ordering consistency and accounts for tied difficulty values.
The normalized rank displacement measures how far each object moves in the ranked list after changing the candidate-view density, and the top-20\% overlap measures whether the high-difficulty tail contains the same objects across protocols.
Both the 64-view and 192-view protocols show high rank correlation with Tammes-128, small normalized rank displacement, and high overlap among the top-20\% hardest objects.
These results support Tammes-128 as a practical candidate-view protocol that preserves the main planning-difficulty ordering and high-difficulty tail while keeping annotation and set-cover computation tractable.

\begin{figure*}[t]
  \centering
  \includegraphics[width=0.65\textwidth]
  {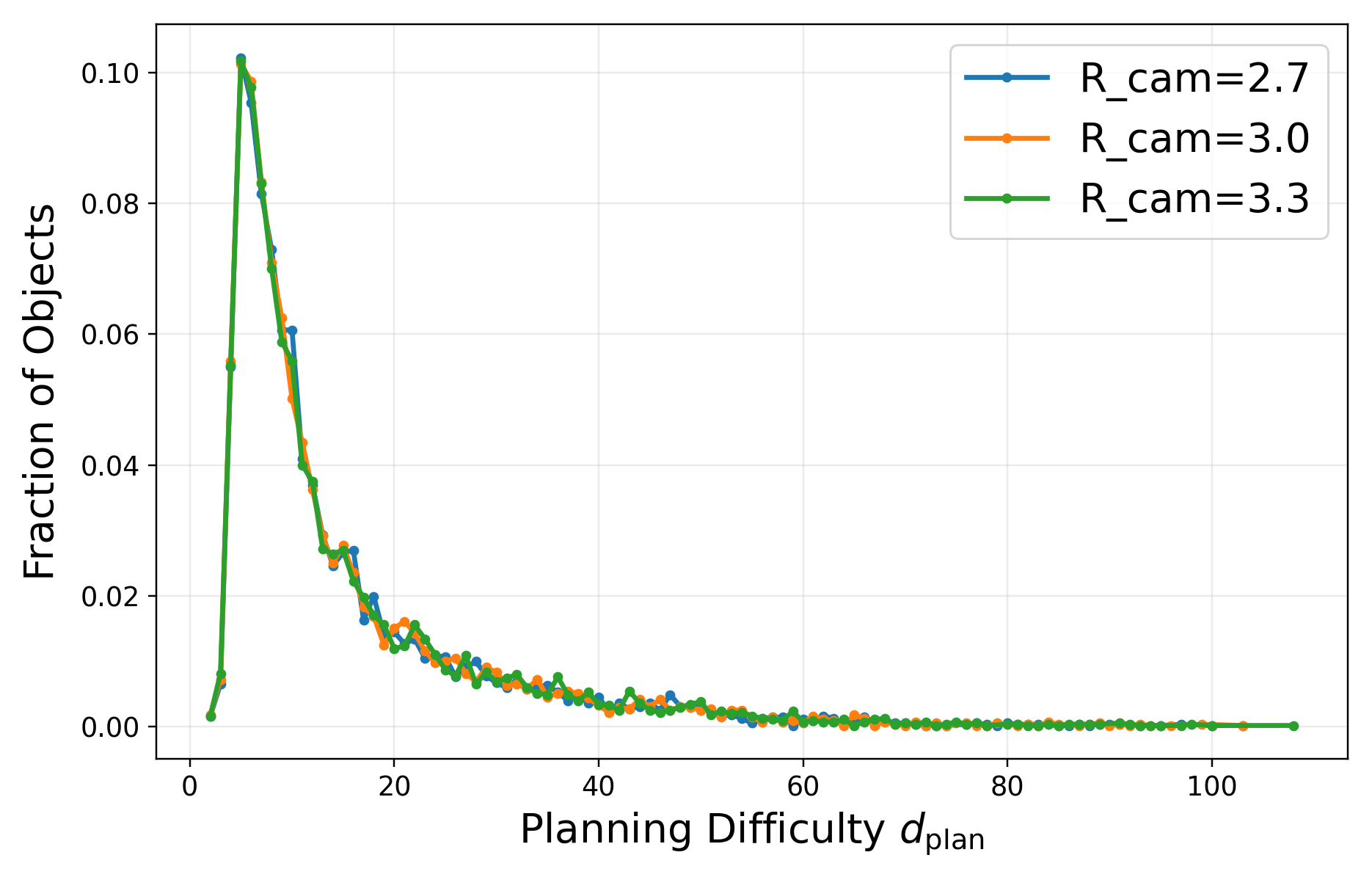}
\caption{
Camera-radius sensitivity of planning difficulty over the main pool objects.
We fix the annotation resolution at $r=0.02$ and the planning candidate set to Tammes-128, and vary only the camera radius used to compute the fixed-protocol set-cover difficulty $d_{\mathrm{plan}}^{(r)}$.
The distributions under $R_{\mathrm{cam}}\in\{2.7,3.0,3.3\}$ retain similar dominant distributional structures, with only minor variation in the high-difficulty tail.
This indicates that the canonical camera setting preserves the overall planning-difficulty structure under moderate camera-distance perturbations.
}
  \label{fig:app_planning_camera_radius_d_plan_distribution}
\end{figure*}

\begin{table*}[t]
\centering
\caption{
Camera-radius sensitivity of planning difficulty over the main pool objects.
We fix the annotation resolution at $r=0.02$, use the same Tammes-128 candidate-view set, and vary only the camera radius used to compute $d_{\mathrm{plan}}^{(r)}$.
All rank-based quantities are computed with respect to the canonical $R_{\mathrm{cam}}=3.0$ protocol.
Rank displacement is normalized by $N-1$.
}
\label{tab:app_planning_camera_radius_rank_stability}
\begin{tabular}{lcccc}
\toprule
Comparison
& Spearman $\rho$
& Kendall $\tau_b$
& Median $\Delta$rank
& Top-20\% overlap \\
\midrule
$R_{\mathrm{cam}}=2.7$ vs. $3.0$ & 0.9932 & 0.9616 & 0.50\% & 95.66\% \\
$R_{\mathrm{cam}}=3.3$ vs. $3.0$ & 0.9934 & 0.9623 & 0.34\% & 95.75\% \\
\bottomrule
\end{tabular}
\end{table*}

\subsubsection{Camera-Radius Sensitivity of Planning Difficulty}

We further examine whether the primary planning-difficulty annotation depends strongly on the exact canonical camera distance.
Although changing the camera radius does not alter the Tammes viewing directions, it changes the projected object scale under a fixed imaging configuration: a closer camera causes the object to occupy more image pixels, whereas a farther camera produces a smaller pixel footprint.
This change may affect the rasterized depth observations, the visibility of fine surface regions, and consequently the fixed-protocol set-cover difficulty $d_{\mathrm{plan}}^{(r)}$.

To isolate this effect, we keep the voxel resolution, Tammes-128 candidate directions, camera orientation convention, visibility operator, and set-cover solver unchanged, and vary only the camera radius using $R_{\mathrm{cam}}\in\{2.7,3.0,3.3\}$.
The camera intrinsics and imaging budget are also fixed, including the $512\times512$ image resolution and $45^\circ$ field of view.
These settings correspond to a $\pm10\%$ perturbation around the canonical radius $R_{\mathrm{cam}}=3.0$.
We additionally verify that all foreground masks remain fully contained within the image under every setting.
Therefore, any observed changes reflect the intended variation in camera distance and projected pixel footprint, rather than differences in image resolution, field of view, or image-boundary truncation.

Figure~\ref{fig:app_planning_camera_radius_d_plan_distribution} shows that the distributions of $d_{\mathrm{plan}}^{(r)}$ remain highly similar across the camera-radius settings over the main pool objects.
The dominant mass of the distribution is nearly unchanged, while the remaining differences are concentrated primarily in the high-difficulty tail.
Consistent with this stability, the mean $d_{\mathrm{plan}}^{(r)}$ remains tightly clustered at 14.95, 15.06, and 15.21 for $R_{\mathrm{cam}}=2.7,3.0,3.3$, respectively.
Thus, moderate changes in camera distance do not substantially alter the overall planning-difficulty structure.

Table~\ref{tab:app_planning_camera_radius_rank_stability} further confirms this trend at the object level.
As in the candidate-view protocol analysis, we report complementary rank-stability measures with respect to the canonical protocol.
Spearman's $\rho$ measures global monotonic agreement between the difficulty rankings, while Kendall's $\tau_b$ measures pairwise ordering consistency and accounts for tied difficulty values.
The normalized rank displacement measures how far each object moves in the ranked list after changing the camera radius, and the top-20\% overlap measures whether the high-difficulty tail contains the same objects across protocols.
Both perturbed camera settings exhibit high rank correlation, small normalized rank displacement, and high overlap among the most difficult objects.

These results indicate that the primary planning-difficulty stratification is stable under moderate camera-radius changes around the canonical configuration.
We do not interpret $d_{\mathrm{plan}}^{(r)}$ as invariant to arbitrary camera models or sensing geometries; rather, this analysis shows that the benchmark ordering is not an artifact of the exact canonical camera radius.

The sensitivity analyses then validate the main protocol choices used for the benchmark.
The saturation-resolution analysis shows that the canonical 128-view protocol lies in a low-marginal-gain regime across surface resolutions, while $r=0.02$ provides a practical balance between geometric detail, annotation stability, and computational cost.
After fixing this annotation resolution, the candidate-view protocol analysis shows that the planning-difficulty structure remains stable across nearby Tammes view densities, with high object-level rank agreement and high overlap in the hardest-object tail.
The camera-radius analysis further shows that this ordering remains stable under moderate changes in camera distance and the resulting projected object scale, while holding the imaging configuration fixed and avoiding image-boundary truncation.
Together, these results support using $d_{\mathrm{sat}}^{(r)}$ to define a slow-saturation diagnostic subset and using $d_{\mathrm{plan}}^{(r)}$ under the documented Tammes-128 protocol as the primary stratification axis for the main benchmark.

\subsection{Multi-Object Clutter and Scene-Level Difficulty Diagnostic}
\label{sec:app_multi_object}

ObjView-Bench is designed primarily for single-object, object-centric reconstruction.
To examine how the proposed difficulty semantics behave when multiple objects share the same scene, we perform a controlled multi-object diagnostic using a moderate-difficulty benchmark object---the cup shown in Fig.~2 of the main paper---replicated into three- and five-instance layouts.
This experiment is not intended to define a complete scene-level benchmark.
Instead, it isolates two effects that arise beyond the single-object setting: the compositional effect of covering multiple object instances from a shared view space, and the additional effect of inter-object occlusion.
The reported results use one reproducible layout seed for each object count and are intended as a mechanism-level diagnostic rather than a population-level multi-object evaluation.

\begin{figure*}[t]
  \centering
  \includegraphics[width=0.60\textwidth]
  {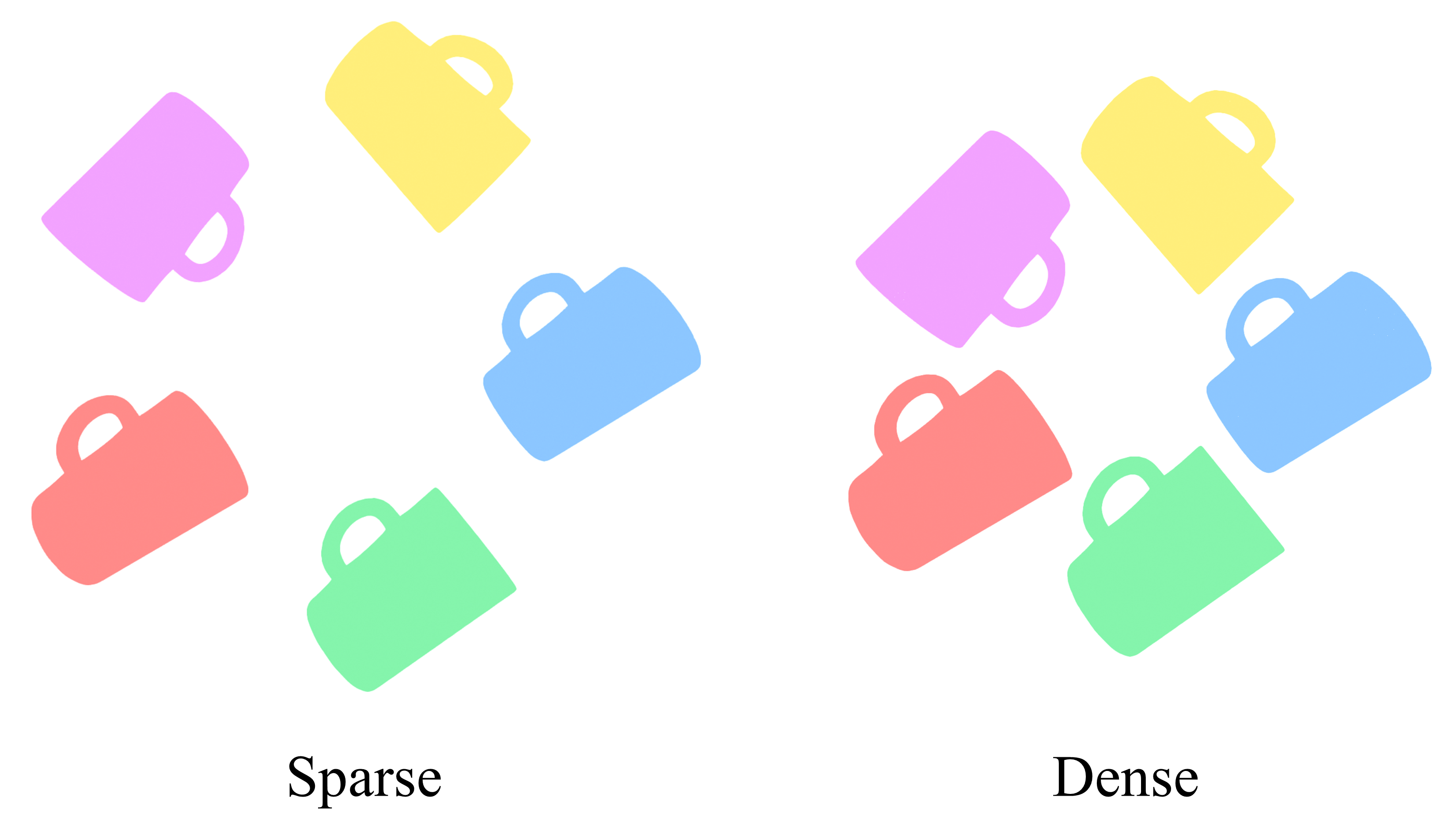}
\caption{
Representative five-object layouts used in the controlled multi-object diagnostic.
The same source object and instance orientations are used in both layouts; only the inter-instance spacing is changed.
The sparse layout (left) places objects farther apart, whereas the dense layout (right) places them more closely and thereby increases the potential for mutual occlusion.
}
  \label{fig:app_multi_object_layouts}
\end{figure*}

Each scene is constructed by scaling the source object by a factor of $0.3$ and placing multiple transformed copies on a regular polygon around the scene origin.
Each instance receives an independently sampled yaw, while a shared random global azimuth determines the polygon orientation.
We evaluate both three- and five-instance scenes under two spacing regimes.
The \emph{dense} and \emph{sparse} layouts use conservative inter-instance XY gaps of $0.03$ and $0.35$, respectively, with the corresponding polygon radius derived from the scaled object footprint.
The dense layout therefore places the instances more closely and creates stronger potential for inter-object occlusion.
All camera positions remain on the canonical radius-$3$ Tammes viewing sphere and look toward the scene origin.
For this diagnostic we use voxel resolution $r=0.01$, image resolution $512\times512$, field of view $45^\circ$, and the same Tammes cardinality sequence and saturation-fitting procedure used by the main difficulty analysis.
Planning difficulty is computed with the Tammes-128 candidate set using the same set-cover formulation.
Figure~\ref{fig:app_multi_object_layouts} illustrates the representative five-object sparse and dense layouts used in this diagnostic.

To separate scene composition from inter-object occlusion, we evaluate two visibility semantics for every multi-object layout.
In the \emph{isolated-union} condition, each instance is ray-cast independently under the same global camera pose and the resulting visible surface sets are combined while retaining instance identity.
This preserves the multi-instance placement and shared view directions but removes mutual occlusion between objects.
In the \emph{joint-scene} condition, all instances are inserted into a single scene and ray-cast jointly, so one object can occlude another.
The difference between these paired conditions therefore isolates the additional effect of inter-object visibility interactions.
No cross-instance occupied-voxel collisions occur in the evaluated layouts.

\begin{table}[t]
\centering
\caption{
Controlled multi-object difficulty diagnostic.
Each multi-object layout is evaluated under isolated-union visibility and joint-scene visibility using identical instance poses.
}
\label{tab:app_multi_object}
\small
\setlength{\tabcolsep}{5pt}
\begin{tabular}{cc|cc|cc|cc}
\toprule
Objects & Layout
& \multicolumn{2}{c|}{$r_{\mathrm{vis}}$}
& \multicolumn{2}{c|}{$d_{\mathrm{sat}}$}
& \multicolumn{2}{c}{$d_{\mathrm{plan}}$} \\
& 
& Iso. & Joint
& Iso. & Joint
& Iso. & Joint \\
\midrule
3 & Sparse & 0.6245 & 0.6241 & 68 & 72 & 30 & 33 \\
3 & Dense  & 0.6248 & 0.6241 & 68 & 76 & 30 & 34 \\
5 & Sparse & 0.6173 & 0.6169 & 64 & 72 & 51 & 56 \\
5 & Dense  & 0.6180 & 0.6170 & 64 & 83 & 48 & 58 \\
\bottomrule
\end{tabular}
\end{table}

For reference, the corresponding single-object configuration has
$d_{\mathrm{sat}}=67$ and $d_{\mathrm{plan}}=24$.
Simply composing multiple independently visible instances does not multiply either difficulty measure.
For example, the three-instance isolated-union scenes have
$d_{\mathrm{plan}}=30$, substantially below three times the single-object value.
Under the shared global view space, a single camera direction can contribute useful observations to several instances simultaneously, so scene-level planning difficulty need not scale linearly with the number of objects.
Likewise, isolated-union saturation difficulty remains close to the single-object value, reaching 68 for the three-instance scenes and 64 for the five-instance scenes.
Thus, scene-level difficulty is not obtained by summing per-object difficulty values; it also depends on how the instances are spatially arranged and how their useful viewing directions align within the shared candidate-view space.

Joint-scene rendering consistently increases both saturation and planning difficulty relative to the corresponding isolated-union control.
For three instances, joint visibility increases $d_{\mathrm{plan}}$ from 30 to 33 in the sparse layout and to 34 in the dense layout, while $d_{\mathrm{sat}}$ increases from 68 to 72 and 76, respectively.
The effect becomes stronger for five instances: the sparse scene increases
$d_{\mathrm{plan}}$ from 51 to 56 and $d_{\mathrm{sat}}$ from 64 to 72, whereas the dense scene increases
$d_{\mathrm{plan}}$ from 48 to 58 and $d_{\mathrm{sat}}$ from 64 to 83.
The largest change occurs in the dense five-instance scene, consistent with stronger inter-object occlusion producing more restrictive reveal and complementary-view requirements.

Interestingly, these difficulty increases occur with only small changes in the fitted observable-surface ratio.
For the three-instance scenes, the isolated-to-joint drop in $r_{\mathrm{vis}}$ is approximately $4.6\times10^{-4}$ in the sparse layout and $7.4\times10^{-4}$ in the dense layout.
For five instances, the corresponding drops are approximately $3.7\times10^{-4}$ and $1.0\times10^{-3}$.
Thus, inter-object clutter can substantially increase the number and specificity of views required to reveal and cover a scene even when the overall externally observable surface fraction changes only slightly.

This controlled diagnostic suggests that the proposed difficulty concepts remain informative in this scene-level setting, but are not directly compositional.
Multi-object scenes introduce an additional interaction term through shared view geometry and inter-object occlusion: independently observable objects may share useful camera directions, while joint visibility can make particular surface regions accessible only from more specific viewpoints.
A complete scene-level benchmark would therefore require explicit scene-composition variables, broader object diversity, and systematic control of layout density and occlusion.
We treat the present experiment as evidence that scene-level difficulty cannot be inferred by simply summing single-object difficulty values, rather than as a complete multi-object benchmark.

\section{Benchmark and Evaluation Protocol Details}
\label{app:benchmark}

\subsection{Rendering API}

The rendering API defines the observation primitive used by all online evaluation episodes. 
Given a normalized object and a camera pose, it returns an RGB-D observation in a fixed object-centric coordinate frame. 
The renderer is bound to one standardized OBJ asset at a time and does not apply additional centering or scaling during evaluation; object normalization is performed during dataset construction as described in Appendix~\ref{app:dataset}.

Camera poses are represented using a common 7-dimensional pose format:
\begin{equation}
p
=
\left(
c_x,c_y,c_z,
\ell_x,\ell_y,\ell_z,
\rho
\right),
\end{equation}
where $(c_x,c_y,c_z)$ is the camera center in the object frame, $(\ell_x,\ell_y,\ell_z)$ is the look-at target, and $\rho$ is the roll angle around the viewing direction. 
This over-parameterized pose representation is used consistently by the rendering, feasibility, interaction, and evaluation components.

For each accepted pose, the rendering API returns an RGB-D frame:
\begin{equation}
\mathcal{F}
=
(I, D, M, K, p),
\end{equation}
where $I$ is an RGB image, $D$ is a depth map, $M$ is a foreground visibility mask, $K$ denotes the camera intrinsics, and $p$ is the executed camera pose. 
Depth values are stored as standard pinhole-camera $z$-depth in the exported RGB-D camera frame, whose axes are defined as $+X$ right, $+Y$ down, and $+Z$ forward. 
The benchmark uses image resolution $512 \times 512$ and field of view $45^\circ$ for the official evaluation renders.

The rendering API supports three execution modes. 
In \texttt{online} mode, RGB-D observations are rendered directly with the PyTorch3D-based renderer. 
In \texttt{cache} mode, observations are loaded from precomputed render caches using tolerance-based pose matching. 
In \texttt{auto} mode, the renderer first attempts to load a cached observation and falls back to online rendering when no matching cache entry is available. 
The render cache is indexed by pose values rather than by method-specific action identifiers, so cached and online observations share the same pose-level semantics.

We use render caching only as an engineering optimization. 
The benchmark protocol remains defined by the submitted camera pose and the returned RGB-D observation, regardless of whether the observation is loaded from cache or generated online. 
The cache worker precomputes RGB-D observations for the standardized start views and canonical Tammes view sets used by evaluation and observable-reference construction, while the online rendering path allows methods that submit valid non-cached poses to be evaluated under the same observation interface.

\subsection{Feasibility API}

The feasibility API defines the executable view space for each evaluation episode. 
It is independent of the planning algorithm: given a submitted camera pose, it returns whether the pose is feasible under the current view-space constraint. 
The same 7-dimensional \texttt{camera\_lookat\_roll} pose format used by the rendering API is used here.

A feasibility specification is provided as a JSON constraint file. 
Each constraint file can contain two components. 
The first is a structured continuous rule over pose fields, such as camera-coordinate inequalities. 
The second is an optional blocklist of specific poses that should be treated as infeasible under a specified numerical tolerance. 
A pose is considered feasible only if it satisfies the structured rule and is not matched by the blocklist:
\begin{equation}
\mathrm{feasible}(p)
=
\mathrm{rule}(p)
\wedge
\neg \mathrm{blocked}(p).
\end{equation}

All official view-space constraints include a basic validity condition that the camera center must lie outside the normalized object unit sphere:
\begin{equation}
\|c\|_2 \geq 1.
\end{equation}
This condition reflects the dataset normalization convention that object geometry lies inside the unit-radius object frame, and prevents submitted camera centers from being placed inside the object volume.

The main benchmark uses two view-space settings. 
The \texttt{whole} setting represents the unconstrained object-centric viewing sphere and applies no additional directional restriction beyond the basic radius constraint. 
The \texttt{quarter} setting represents a constrained reachable view space and additionally requires:
\begin{equation}
c_z \geq 0,
\qquad
c_y \geq 0.
\end{equation}
Thus, \texttt{quarter} preserves only one quadrant of the external viewing sphere. 
The implementation also supports other constraint files, such as hemispherical or one-eighth view spaces, but the main experiments report the \texttt{whole} and \texttt{quarter} settings because they capture the contrast between idealized and strongly constrained deployment conditions.

During an episode, every submitted move action is checked by the feasibility API before rendering. 
If the pose is feasible, the runner accepts the move, renders the corresponding RGB-D observation, and updates the evaluator. 
If the pose is infeasible, the move is not executed and the episode terminates with an infeasible-action terminal reason. 
This policy ensures that all reported observations correspond to executable poses under the selected view-space constraint.

\subsection{Evaluation API}

The evaluation API is an episode-level metric accumulator. 
It is called only after the benchmark runner has accepted a pose, obtained an RGB-D observation, and decided that the observation should become part of the episode history. 
It does not control the protocol, decide pose feasibility, interact with the algorithm, or determine terminal reasons. 
Its role is to fuse accepted observations and update the official metrics used for reporting.

For each accepted RGB-D frame, the evaluator back-projects the valid depth pixels into a point cloud in the camera frame and transforms the points into the object-centric world frame using the executed camera pose. 
The accumulated observation point cloud after step $t$ is denoted by $P_t$. 
To reduce duplicate samples from overlapping views, the evaluator fuses point clouds with voxel downsampling at voxel size $0.005$ in the normalized object coordinate frame.

Surface coverage is computed against the complete ground-truth surface point cloud $P_{\mathrm{gt}}$. 
For a distance threshold $\tau$, we define:
\begin{equation}
\mathrm{SC}_{\tau}(P_t)
=
\frac{
\left|
\left\{
q \in P_{\mathrm{gt}}
\;\middle|\;
\min_{p \in P_t}
\|q-p\|_2
\leq \tau
\right\}
\right|
}{
|P_{\mathrm{gt}}|
}.
\end{equation}
The evaluator reports $\mathrm{SC}@0.01$, $\mathrm{SC}@0.02$, and $\mathrm{SC}@0.03$.

Normalized surface coverage is computed using a view-space-specific observable reference. 
For each object and view-space constraint, the evaluator constructs an observable reference point cloud $P_{\mathrm{obs}}^{\mathrm{ref}}$ by fusing RGB-D observations rendered from a dense \texttt{tammes\_360} view set under the same feasibility constraint. 
This reference estimates the surface that is observable under the current view-space restriction. 
For threshold $\tau$, normalized surface coverage is:
\begin{equation}
\mathrm{NSC}_{\tau}(P_t)
=
\frac{
\left|
\left\{
q \in P_{\mathrm{obs}}^{\mathrm{ref}}
\;\middle|\;
\min_{p \in P_t}
\|q-p\|_2
\leq \tau
\right\}
\right|
}{
|P_{\mathrm{obs}}^{\mathrm{ref}}|
}.
\end{equation}
Thus, $\mathrm{NSC}$ measures completion relative to the observable surface under the same view-space constraint, rather than relative to all ground-truth surface points. 
This makes $\mathrm{NSC}$ comparable across \texttt{whole} and \texttt{quarter} settings, whose observable ceilings can differ.

The evaluator also reports Chamfer Distance between the accumulated observation point cloud and the full ground-truth surface point cloud:
\begin{equation}
\begin{aligned}
\mathrm{CD}(P_t, P_{\mathrm{gt}})
=
&
\frac{1}{|P_t|}
\sum_{p \in P_t}
\min_{q \in P_{\mathrm{gt}}}
\|p-q\|_2
\\
+
&
\frac{1}{|P_{\mathrm{gt}}|}
\sum_{q \in P_{\mathrm{gt}}}
\min_{p \in P_t}
\|q-p\|_2 .
\end{aligned}
\end{equation}
$\mathrm{CD}$ is reported as a complementary geometry-error metric, while $\mathrm{NSC}@0.02$ is used as the primary reconstruction-quality metric in the main tables.

Path cost is accumulated from the executed camera centers, but each transition is measured with a collision-avoiding distance around the normalized unit object. 
Let $R=1$ denote the radius of the protected object sphere, and let $c_i$ be the camera center of the $i$-th acquired observation. 
For two consecutive camera centers $p$ and $q$, we first check whether the straight line segment intersects the protected sphere:
\begin{equation}
d_{\mathrm{seg}}(p,q)
=
\min_{\lambda \in [0,1]}
\left\|
(1-\lambda)p + \lambda q
\right\|_2 .
\end{equation}
If the segment does not intersect the sphere, the transition cost is the Euclidean distance:
\begin{equation}
d_{\mathrm{move}}(p,q)
=
\|p-q\|_2,
\qquad
d_{\mathrm{seg}}(p,q) \geq R .
\end{equation}

If the segment intersects the protected sphere, we use the shortest tangent--arc--tangent path around the sphere. 
Let
\begin{equation}
r_p = \|p\|_2,
\qquad
r_q = \|q\|_2,
\end{equation}
and
\begin{equation}
\theta
=
\arccos
\left(
\frac{p^\top q}{r_p r_q}
\right).
\end{equation}
The tangent lengths from $p$ and $q$ to the sphere are
\begin{equation}
\ell_p
=
\sqrt{r_p^2 - R^2},
\qquad
\ell_q
=
\sqrt{r_q^2 - R^2}.
\end{equation}
The tangent angles are
\begin{equation}
\gamma_p
=
\arccos
\left(
\frac{R}{r_p}
\right),
\qquad
\gamma_q
=
\arccos
\left(
\frac{R}{r_q}
\right).
\end{equation}
In the plane spanned by $p$ and $q$, we represent the direction of $p$ by angle $0$ and the direction of $q$ by angle $\theta$. 
The two possible tangent points from $p$ have angles
\begin{equation}
\mathcal{A}_p
=
\{
+\gamma_p,\,
-\gamma_p
\},
\end{equation}
and the two possible tangent points from $q$ have angles
\begin{equation}
\mathcal{A}_q
=
\{
\theta+\gamma_q,\,
\theta-\gamma_q
\}.
\end{equation}
The shortest surface-arc angle is then
\begin{equation}
\delta(p,q)
=
\min_{a \in \mathcal{A}_p,\ b \in \mathcal{A}_q}
\min
\left(
|a-b|_{2\pi},
2\pi - |a-b|_{2\pi}
\right),
\end{equation}
where $|a-b|_{2\pi}$ denotes the absolute angular difference after wrapping angles to $[0,2\pi)$.

The collision-avoiding transition cost is:
\begin{equation}
d_{\mathrm{move}}(p,q)
=
\ell_p
+
\ell_q
+
R\,\delta(p,q),
\qquad
d_{\mathrm{seg}}(p,q) < R .
\end{equation}
The accumulated path cost after step $t$ is:
\begin{equation}
\mathrm{PathCost}_t
=
\sum_{i=1}^{t}
d_{\mathrm{move}}
\left(
c_{i-1}, c_i
\right).
\end{equation}
This cost is computed in the normalized object coordinate frame. 
It avoids assigning artificially short travel distances to transitions whose straight-line segment would pass through the normalized object volume.

The evaluator separately records algorithm-reported runtime and benchmark wall-clock time. 
The main result tables use algorithm runtime as the method-side computational cost, while rendering, metric computation, interaction overhead, and benchmark bookkeeping are treated as evaluator overhead.

Finally, the evaluator maintains map-stabilization statistics used by the automatic stopping protocol. 
For each stabilization voxel size $\delta \in \{0.01,0.02,0.03\}$, the accumulated observation point cloud is voxelized to obtain a voxel set $M_t^\delta$ with size $N_t^\delta$. 
The relative map-growth rate is:
\begin{equation}
g_t^\delta
=
\frac{
N_t^\delta - N_{t-1}^\delta
}{
\max(N_{t-1}^\delta, 1)
}.
\end{equation}
A map-stabilization checkpoint $\mathrm{MS}@\delta$ is triggered when $g_t^\delta < 0.01$ for three consecutive accepted observations. 
The main paper uses $\mathrm{MS}@0.02$ as the default automatic stopping checkpoint.

\subsection{Interaction API}

The interaction API defines the boundary between the benchmark process and the evaluated algorithm. 
It is implemented as a shared-folder protocol with a small JSON-RPC surface. 
The benchmark writes released episode information, current state, observation manifests, RPC responses, and logs; the algorithm writes RPC requests and final actions. 
This separation prevents methods from accessing hidden geometry, ground-truth coverage sets, or evaluator internals.

At the beginning of an episode, the benchmark publishes an interface description and the episode configuration. 
At each step, it publishes a current-step record and an observation manifest:
\begin{equation}
\mathcal{M}_t
=
\left(
\mathrm{episode\_id},
t,
\mathrm{visited\_view\_num},
p_t,
\mathrm{observation\_path}
\right),
\end{equation}
where $p_t$ is the current executed pose. 
The observation manifest points either to a cached RGB-D frame or to an RGB-D frame materialized in the current session directory. 
Algorithms therefore consume observations through the same manifest interface regardless of whether the frame was rendered online or loaded from cache.

The algorithm responds by submitting exactly one final action for the current step. 
The supported action types are:
\begin{equation}
a_t
\in
\{
\mathrm{move}(p),
\mathrm{stop}(s)
\},
\end{equation}
where $\mathrm{move}(p)$ proposes a new camera pose and $\mathrm{stop}(s)$ terminates the episode with an algorithm-side stop reason. 
The interaction layer validates the action format and associates the action with the expected episode and step index before the episode runner processes it.

In addition to final actions, the interaction API exposes a small set of safe RPC methods:
\begin{equation}
\{
\texttt{describe\_interface},
\texttt{is\_feasible},
\texttt{pose\_to\_matrix},
\texttt{submit\_action}
\}.
\end{equation}
The \texttt{is\_feasible} method allows an algorithm to query the feasibility of candidate poses under the current view-space constraint, and \texttt{pose\_to\_matrix} converts the benchmark pose format into camera extrinsics for method-side processing. 
These queries do not expose ground-truth geometry or hidden-test metadata. 
An additional oracle-gain query exists only for oracle-family analysis methods and is disabled for regular benchmark methods.

To avoid race conditions from partially written files, the interaction protocol uses ready-signal files. 
After publishing a new observation, the benchmark writes a \texttt{ready\_benchmark} signal. 
After writing an action or request, the algorithm writes the corresponding ready signal. 
The benchmark only reads an action after the algorithm-ready signal is present and clears the signal after processing. 
This handshake makes the interaction deterministic and robust to asynchronous algorithm processes.

All requests, responses, submitted actions, and observation-publication events are appended to an interaction history. 
The final episode summary therefore records not only the evaluated trajectory and metrics, but also the sequence of algorithm decisions and feasibility queries used to obtain it.

\subsection{Episode Runner}

The episode runner is the single-episode execution core of ObjView-Bench. 
It evaluates one rollout for a fixed object, start view, view-space constraint, and method. 
An episode is therefore identified by:
\begin{equation}
\mathrm{episode}
=
(\mathrm{object},\ \mathrm{start\ view},\ \mathrm{viewspace},\ \mathrm{method}).
\end{equation}
Budget settings are not separate episode dimensions. 
Instead, fixed-budget and automatic-stopping results are recorded as checkpoints within the same online rollout.

Each episode begins with a standardized initial observation. 
The runner renders the initial pose, records it in the evaluator as step $0$, publishes the corresponding observation manifest, and waits for the algorithm to submit an action. 
After initialization, the runner repeatedly performs the following loop:
\begin{equation}
\mathrm{observe}
\rightarrow
\mathrm{wait\ for\ action}
\rightarrow
\mathrm{validate}
\rightarrow
\mathrm{feasibility\ check}
\rightarrow
\mathrm{render}
\rightarrow
\mathrm{evaluate}.
\end{equation}
If the algorithm submits a valid and feasible move action, the runner renders the new RGB-D observation, updates the evaluator, publishes the next observation manifest, and continues. 
If the algorithm submits a stop action, the episode terminates with an algorithm-side stop reason. 
If the submitted action is malformed or infeasible under the current view-space constraint, the episode terminates with a benchmark-side invalid-action reason.

We distinguish between the number of algorithm-selected next-best-view actions and the total number of acquired observations. 
Let $K$ denote the number of accepted algorithm-selected move actions after the standardized initial observation. 
The reported number of views is:
\begin{equation}
\#\mathrm{Views}
=
K + 1,
\end{equation}
where the additional view is the initial observation. 
Thus, a fixed-budget checkpoint $K=5$ corresponds to $6$ acquired observations, and $K=30$ corresponds to $31$ acquired observations.

The runner records fixed-budget checkpoints whenever the accepted move count reaches a specified budget:
\begin{equation}
K \in \{5,10,30\}.
\end{equation}
The main paper reports representative $K=5$ and $K=30$ settings, while the complete benchmark outputs also retain the other checkpoints for analysis. 
The runner also records map-stabilization checkpoints $\mathrm{MS}@0.01$, $\mathrm{MS}@0.02$, and $\mathrm{MS}@0.03$ using the evaluator-side stabilization test described above.

For free-budget execution, the runner allows the method to proceed until a natural stop condition, a map-stabilization checkpoint, or a safety cap is reached. 
The default safety cap is:
\begin{equation}
\mathrm{max\ visited\ views}
=
129,
\end{equation}
corresponding to one standardized initial observation plus at most $128$ accepted NBV actions. 
This cap prevents unbounded episodes while preserving the full Tammes-128 action horizon used by the benchmark.

At termination, the runner records a terminal checkpoint using the current accumulated observation state. 
The episode summary contains the terminal reason, accepted NBV action count, total visited view count, fixed-budget checkpoints, map-stabilization checkpoints, terminal metrics, runtime accumulators, and the algorithm action history. 
These summaries are later aggregated across objects, start views, view-space constraints, and methods.

\subsection{Suite Runner and Result Aggregation}

The suite runner extends the single-episode runner to full benchmark evaluation.
It enumerates evaluation jobs over objects, view-space constraints, and methods:
\begin{equation}
\mathcal{J}
=
\mathcal{O}
\times
\mathcal{V}
\times
\mathcal{M},
\end{equation}
where $\mathcal{O}$ is the evaluated object set, $\mathcal{V}$ is the set of view-space constraints, and $\mathcal{M}$ is the set of evaluated methods.

The benchmark implementation also supports a shared start-view set, \texttt{benchmark\_start\_3}. 
Its three standardized initial views are chosen to be feasible under the constrained view-space settings, allowing the same initialization protocol to be used for both \texttt{whole} and \texttt{quarter} evaluation. 
In the experiments reported in this paper, we use the first start view, \texttt{start\_view\_id}=0, for all methods and objects to keep the full method-by-protocol evaluation computationally tractable. 
Since all methods share the same initial observation within each object and view-space setting, the reported comparisons are controlled with respect to initialization. 
The remaining standardized start views are retained in the benchmark implementation for future multi-start evaluation and robustness analysis.

For each job, the suite runner creates a session directory, resolves the object geometry and ground-truth point cloud paths, selects the corresponding feasibility constraint file, and launches a single episode through the episode launcher. 
The episode launcher starts the method process, constructs the rendering, feasibility, evaluation, and interaction APIs, runs the episode runner, and writes the resulting episode summary. 
The suite runner itself does not compute metrics or advance the rollout; it only schedules independent episodes and manages their output paths.

Each completed episode produces a summary file containing the method identity, family, view-space constraint, start-view metadata, terminal reason, accepted NBV action count, visited view count, evaluator summary, fixed-budget checkpoints, map-stabilization checkpoints, and terminal checkpoint. 
Because fixed-budget and automatic settings are stored as checkpoints within the same rollout, result aggregation does not require rerunning the same episode for different budget rules.

The result aggregation script reads all episode summaries and extracts rows for requested checkpoints. 
For a checkpoint $q$, the aggregation unit is:
\begin{equation}
(\mathrm{method},\ \mathrm{family},\ \mathrm{viewspace},\ q).
\end{equation}
With a single shared start view in the reported experiments, metrics are averaged over evaluated objects for each group. 
For each group, we report the mean values of reconstruction quality, path cost, view count, and runtime, and retain dispersion statistics such as standard deviation or standard error for full tables when needed.

The main paper reports representative deployment settings, including fixed-budget checkpoints and the automatic stopping checkpoint. 
The full benchmark output keeps all per-episode summaries and checkpoint-level rows, enabling additional post-hoc tables over methods, families, view-space constraints, metric thresholds, and start-view settings without changing the underlying rollout protocol.

\subsection{Sensing Noise Analysis}
\label{app:sensing_noise}

\subsubsection{Dense Accumulation under Perturbations}

ObjView-Bench primarily evaluates view-planning behavior under clean RGB-D observations.
This design is intentional: adding sensing noise to the main benchmark would introduce additional confounding factors beyond view selection, including depth-noise modeling, pose uncertainty, point-cloud fusion, registration, map filtering, and stopping under noisy intermediate observations.
These factors are important for deployment, but they can obscure whether a method succeeds or fails because of view-planning decisions or because of the particular noisy sensing and fusion assumptions.
We therefore keep the main benchmark clean in order to isolate the object-centric view-planning problem, while treating noisy sensing as a deployment-oriented extension.

To nevertheless examine how the evaluation metrics respond to sensing perturbations, we include a lightweight fixed-view sanity check.
This experiment does not define a noisy active-planning benchmark.
Instead, it uses the same rendering and evaluation infrastructure, fixes the observation protocol to all Tammes-128 views, and evaluates the accumulated observation point cloud against the same Tammes-360 externally observable reference used by the main benchmark.
No planner, stopping policy, online registration, ICP correction, or closed-loop re-rendering under perturbed poses is involved.
For depth perturbations, we add zero-mean Gaussian noise to valid foreground depth values before back-projection.
For pose perturbations, we keep the rendered depth image fixed but perturb the camera-to-world transform used for back-projection, simulating calibration or execution pose error.
All accumulated observed point clouds are voxel-downsampled using the same observation voxel size of 0.005 before evaluation.

Let $P_{\mathrm{vis}}$ denote the Tammes-360 externally observable reference point cloud and $P_{\mathrm{obs}}$ denote the accumulated noisy Tammes-128 observation point cloud.
For a distance threshold $\tau$, we compute the observable-surface recall as
\begin{equation}
\mathrm{Recall}_{\tau}
=
\frac{
\left|
\left\{
q \in P_{\mathrm{vis}}
\;\middle|\;
\min_{p \in P_{\mathrm{obs}}}
\|q-p\|_2
\leq \tau
\right\}
\right|
}{
|P_{\mathrm{vis}}|
}.
\end{equation}
This follows the same externally observable reference convention as NSC in the main benchmark.

Since recall-style coverage does not penalize off-surface or misregistered observed points, we also compute
\begin{equation}
\mathrm{Precision}_{\tau}
=
\frac{
\left|
\left\{
p \in P_{\mathrm{obs}}
\;\middle|\;
\min_{q \in P_{\mathrm{vis}}}
\|p-q\|_2
\leq \tau
\right\}
\right|
}{
|P_{\mathrm{obs}}|
}.
\end{equation}
Finally, we report the harmonic mean,
\begin{equation}
\mathrm{F1}_{\tau}
=
\frac{
2 \cdot \mathrm{Precision}_{\tau} \cdot \mathrm{Recall}_{\tau}
}{
\mathrm{Precision}_{\tau} + \mathrm{Recall}_{\tau}
}.
\end{equation}
We report results at $\tau \in \{0.01,0.02,0.03\}$, with $\tau=0.02$ matching the main benchmark tolerance.

Noise magnitudes are specified in normalized object coordinates.
For depth perturbations, we use $\sigma_z \in \{0,0.005,0.010,0.020\}$, corresponding to $0$, $0.25$, $0.5$, and $1.0$ times the main tolerance $\tau=0.02$.
For pose perturbations, we use translation noise $\sigma_t \in \{0,0.002,0.005,0.010\}$ and rotation noise $\sigma_r \in \{0^\circ,0.10^\circ,0.25^\circ,0.50^\circ\}$.
For intuition, we also report an approximate displacement-scale proxy,
\[
\sigma_{\mathrm{disp}} \approx \sigma_t + 3\sin(\sigma_r),
\]
where $\sigma_r$ is converted to radians and 3 is the canonical camera radius in normalized coordinates.
This proxy is used only to indicate the approximate geometric scale of the pose perturbation relative to the main tolerance $\tau=0.02$; it is not intended as a statistically equivalent combination of translational and rotational noise.
The four pose settings correspond to approximately $0$, $0.36$, $0.91$, and $1.81$ times $\tau$, respectively.
The two perturbation types are not intended to be matched one-to-one in physical severity: depth noise is a per-pixel range perturbation, whereas pose noise is a frame-level extrinsic perturbation.

\begin{table*}[t]
\centering
\scriptsize
\setlength{\tabcolsep}{3.0pt}
\caption{
Fixed-view sensing perturbation sanity check under different evaluation tolerances over all Main-Hidden-Test objects.
All Tammes-128 observations are accumulated and evaluated against the Tammes-360 externally observable reference.
Each entry reports R / P / F1 in percentage, where R denotes Recall and P denotes Precision.
Depth perturbations use Gaussian depth noise $\sigma_z$.
Pose perturbations use translation noise $\sigma_t$ and rotation noise $\sigma_r$ for back-projection.
All translational and depth perturbation magnitudes are specified in normalized object coordinates.
}
\label{tab:app_fixed_view_noise_sanity}
\begin{tabular}{llccc}
\toprule
Perturbation & Noise parameters 
& $\tau=0.01$ 
& $\tau=0.02$ 
& $\tau=0.03$ \\
& 
& R / P / F1
& R / P / F1
& R / P / F1 \\
\midrule
Clean 
& -- 
& 99.93 / 99.99 / 99.96
& 99.98 / 100.00 / 99.99
& 99.99 / 100.00 / 100.00 \\
\midrule
Depth 
& $\sigma_z=0.005$
& 99.95 / 91.66 / 95.62
& 99.99 / 99.99 / 99.99
& 99.99 / 100.00 / 100.00 \\

Depth 
& $\sigma_z=0.010$
& 99.96 / 61.47 / 76.08
& 99.99 / 95.14 / 97.50
& 100.00 / 99.81 / 99.90 \\

Depth 
& $\sigma_z=0.020$
& 99.98 / 39.26 / 56.29
& 99.99 / 70.84 / 82.89
& 100.00 / 89.56 / 94.48 \\
\midrule
Pose 
& $\sigma_t=0.002,\ \sigma_r=0.10^\circ$
& 99.93 / 86.91 / 92.94
& 99.98 / 99.90 / 99.94
& 99.99 / 100.00 / 100.00 \\

Pose 
& $\sigma_t=0.005,\ \sigma_r=0.25^\circ$
& 99.91 / 49.45 / 66.07
& 99.99 / 85.11 / 91.92
& 99.99 / 97.61 / 98.78 \\

Pose 
& $\sigma_t=0.010,\ \sigma_r=0.50^\circ$
& 99.87 / 31.10 / 47.34
& 99.98 / 58.05 / 73.37
& 99.99 / 77.80 / 87.47 \\
\bottomrule
\end{tabular}
\end{table*}

The fixed-view results over all Main-Hidden-Test objects in Table~\ref{tab:app_fixed_view_noise_sanity} show that observable recall remains close to saturated under dense Tammes-128 sensing, even when depth or pose perturbations are introduced.
This observable recall follows the same externally observable reference convention as NSC in the main benchmark.
Thus, the result indicates that NSC-style coverage is relatively insensitive to these perturbations under a dense fixed-view protocol.
This is expected: recall-style coverage only asks whether each externally observable reference point has at least one nearby observed point, and dense fixed-view sensing provides many opportunities for such coverage.

However, Precision and F1 degrade much more clearly as perturbation magnitude increases.
For example, at the main tolerance $\tau=0.02$, depth noise with $\sigma_z=0.020$ keeps observable recall at $99.99\%$, but reduces Precision and F1 to $70.84\%$ and $82.89\%$, respectively.
Similarly, pose noise with $\sigma_t=0.010$ and $\sigma_r=0.50^\circ$ keeps observable recall at $99.98\%$, but reduces Precision and F1 to $58.05\%$ and $73.37\%$.
Under this dense fixed-view protocol, the dominant measured effect of sensing perturbations is the introduction of off-surface, thickened, or misregistered geometry, rather than a substantial loss of observable-surface recall.

This fixed-view analysis shows that sensing perturbations require geometric-consistency metrics beyond NSC-style coverage.
Under dense Tammes-128 accumulation, observable-surface recall can remain nearly saturated even when the reconstructed geometry becomes substantially degraded, whereas Precision and F1 expose off-surface, thickened, and misregistered observations.

This protocol should therefore be interpreted as a dense-observation sensing diagnostic rather than a noisy active-planning benchmark.
It motivates additional metric design, but does not yet characterize how sensing perturbations interact with sequential map updates, registration, and observation order.
We examine this interaction next through a controlled sequential-fusion diagnostic.

\subsubsection{Mechanism-Level Sequential Fusion}

The preceding sensing-perturbation analysis characterizes how imperfect depth and pose estimates affect accumulated geometry under a fixed observation protocol.
However, a full noisy-planner comparison introduces an additional source of variation: different planners may induce different observation orders, and these orders interact with registration and map fusion.
A direct noisy rerun of all baselines would therefore measure not only view-planning behavior, but also method-specific choices for registration, filtering, and fusion.
Conversely, an adaptation-aware comparison would require these components and their tuning budgets to be standardized.
We therefore perform an additional controlled diagnostic that isolates one mechanism shared across planners: the interaction between pose error, observation order, and sequential registration.

We evaluate all 528 Main-Hidden-Test objects using the Tammes-128 view set under a fixed strong pose-perturbation condition with translational standard deviation $\sigma_t=0.01$ and rotational standard deviation $\sigma_r=0.5^\circ$.
The initial observation is retained as an unperturbed anchor, while the remaining camera poses are independently perturbed before back-projection and fusion.
Starting from the same initial view, we construct two deterministic traversal orders.
The \emph{nearest} order greedily selects the remaining camera position closest to the current view, whereas the \emph{farthest} order selects the most distant one.
These trajectories are designed to induce different local-overlap regimes: the nearest traversal favors view continuity and higher consecutive-frame overlap, whereas the farthest traversal favors rapid acquisition of complementary geometry at the cost of reduced local overlap for registration.

For each traversal, we compare three conditions: clean poses, perturbed poses without registration, and perturbed poses with point-to-point ICP registration against the accumulated reconstruction.
For the ICP condition, the estimated correction is applied only when the registration passes the validity and residual-error checks; otherwise, the uncorrected noisy observation is fused.
We use point-to-point ICP with a voxel-downsampled registration resolution of $0.01$, a maximum correspondence distance of $0.025$, at most 50 iterations, and accept a correction only when at least three correspondences are found and the inlier MSE is below $10^{-4}$.
We use the same Tammes-360 externally observable reference, observation voxel size, and geometric matching definitions as in the preceding sensing-perturbation analysis.
We report F1 at $\tau=0.02$ after increasing numbers of accumulated observations.
F1 is used rather than coverage recall alone because pose errors may introduce off-surface or duplicated geometry that remains unpenalized by a one-sided coverage metric.

\begin{table}[t]
\centering
\caption{
Sequential-fusion diagnostic under strong pose perturbation.
Values are F1@0.02 (\%) averaged over all 528 hidden-test objects.
Nearest and farthest denote greedy traversals according to camera-center distance.
ICP denotes point-to-point registration of each incoming observation against the accumulated reconstruction.
}
\label{tab:app_sequential_fusion}
\small
\setlength{\tabcolsep}{4pt}
\begin{tabular}{c|ccc|ccc}
\toprule
& \multicolumn{3}{c|}{Nearest traversal}
& \multicolumn{3}{c}{Farthest traversal} \\
Views
& Clean & Noise & Noise+ICP
& Clean & Noise & Noise+ICP \\
\midrule
2   & 70.23 & 65.92 & 69.88 & 92.84 & 78.95 & 79.09 \\
5   & 77.61 & 68.63 & 76.78 & 96.26 & 81.54 & 84.31 \\
10  & 86.13 & 74.43 & 84.86 & 98.38 & 83.31 & 88.19 \\
30  & 96.48 & 78.59 & 92.76 & 99.79 & 81.81 & 91.40 \\
60  & 99.68 & 78.63 & 94.59 & 99.93 & 78.86 & 91.43 \\
128 & 99.99 & 73.54 & 92.63 & 99.99 & 73.54 & 89.53 \\
\bottomrule
\end{tabular}
\end{table}

Table~\ref{tab:app_sequential_fusion} first exposes the expected clean-data trade-off between complementary coverage and local overlap.
The farthest traversal discovers complementary surface regions much faster, reaching 92.84\% F1 after only two views and 98.38\% after ten, compared with 70.23\% and 86.13\% for the nearest traversal.
In contrast, the nearest traversal favors stronger local continuity between successive observations, providing a more favorable condition for sequential registration.

Under pose perturbation, ICP recovers a substantial fraction of the geometric degradation for both trajectories, but the benefit becomes systematically larger for the nearest traversal as accumulation proceeds.
At 30 views, for example, ICP improves F1 from 78.59\% to 92.76\% for the nearest traversal, compared with 81.81\% to 91.40\% for the farthest traversal.
At 60 views, the corresponding F1 values reach 94.59\% and 91.43\%.
After all 128 views have been accumulated, the clean and unregistered-noise conditions become identical across the two traversal orders because both have incorporated the same complete observation set.
The ICP results remain order-dependent, however, with 92.63\% F1 for the nearest traversal and 89.53\% for the farthest traversal.
This isolates an order-dependent effect of sequential registration: although the acquired view set is identical, the final reconstruction can differ because each ICP update is conditioned on the previously accumulated map.

Interestingly, F1 does not improve monotonically once surface coverage is close to saturation.
For the ICP trajectories, F1 decreases from 94.59\% to 92.63\% between 60 and 128 views for the nearest traversal, and from 91.43\% to 89.53\% for the farthest traversal.
This decrease is consistent with accumulated registration error or additional misaligned geometry outweighing the small remaining coverage gains once most observable surface regions have already been recovered.
This further illustrates why sensing robustness cannot be characterized by coverage alone.

Together with the dense Tammes-128 perturbation analysis, this diagnostic identifies two complementary requirements for evaluating imperfect sensing.
First, reconstruction quality should include geometric-consistency measures such as Precision or F1 in addition to NSC-style coverage, because one-sided coverage can remain high despite substantial geometric degradation.
Second, planner comparisons require an explicit protocol for registration, fusion, filtering, and method-specific adaptation.
The order-dependent ICP results further show that simply injecting noise into the existing clean benchmark would conflate view-planning quality with the robustness of the surrounding reconstruction pipeline.

We therefore treat this experiment as a mechanism-level diagnostic rather than a noisy-baseline leaderboard.
The primary benchmark retains clean observations to isolate view-planning behavior, while a comprehensive noisy active-planning benchmark with standardized perturbation, registration, fusion, metric, and adaptation protocols remains an important future extension.

\section{Baseline Implementation Details}
\label{app:baselines}

\subsection{Common Algorithm Interface}

All baselines are wrapped as online algorithms under the same benchmark interaction protocol described in Appendix~\ref{app:benchmark}. 
At each step, the benchmark releases the current RGB-D observation and the current camera pose through an observation manifest. 
The algorithm may maintain any private internal state, such as a voxel map, accumulated point cloud, completed shape prediction, candidate-view scores, or a finite planned trajectory. 
However, the benchmark evaluator does not access this internal state. 
All methods are scored only from the executed RGB-D observations recorded by the official evaluator.

Each algorithm receives the current episode state and returns either a move action or a stop action:
\begin{equation}
a_t
\in
\{
\mathrm{move}(p_t),
\mathrm{stop}(s_t)
\},
\end{equation}
where $p_t$ is a proposed camera pose in the benchmark \texttt{camera\_lookat\_roll} format and $s_t$ is an algorithm-side stop reason. 
For methods that operate over a fixed discrete candidate-view set, the algorithm first selects a candidate index and the wrapper converts it to the corresponding benchmark pose. 
For methods that generate continuous or object-dependent pose candidates, such as boundary-based view prediction, the wrapper directly submits the generated pose.

All submitted poses are checked by the benchmark feasibility API before execution. 
The wrapper may query feasibility during planning, but the final submitted action is still validated by the benchmark runner. 
If a submitted pose is infeasible, already invalid, or malformed, the runner terminates the episode with a benchmark-side failure reason. 
This policy ensures that all reported observations correspond to executable poses under the selected view-space constraint.

We distinguish two execution modes. 
Iterative methods replan after every newly acquired observation:
\begin{equation}
p_t
=
\pi
\left(
\mathcal{H}_t
\right),
\end{equation}
where $\mathcal{H}_t$ is the observation history up to step $t$. 
Finite-plan methods instead compute an ordered list of future views:
\begin{equation}
\Pi_t
=
(p_t, p_{t+1}, \ldots, p_{t+k}),
\end{equation}
and then execute this list until the plan ends, a candidate becomes infeasible, or the benchmark reaches a checkpoint or safety cap. 
This distinction is recorded as method metadata but does not change the external evaluation protocol: both modes submit one action at a time to the same online runner.

For the primary benchmark, all candidate-set-compatible methods use the canonical Tammes-128 view set.
We additionally evaluate Tammes-360 as a candidate-view discretization sensitivity analysis in Appendix~\ref{app:candidate_sensitivity}.
Methods with native dynamic candidate generation retain their original action interfaces.

Algorithm runtime is measured on the method side and accumulated by the benchmark runner. 
For methods that invoke external services, such as planning networks or shape completion modules, the reported method runtime includes the service inference time used to produce the submitted action. 
Rendering, metric computation, and benchmark bookkeeping are not counted as method-side planning time.

\subsection{Simple and Classical Baselines}

We include simple and classical baselines to provide non-learned reference points for the benchmark.
These methods do not require training on ObjView-Bench and therefore help separate the effect of learned priors from the effect of candidate-view search, coverage optimization, and trajectory ordering.
All methods in this group receive the same online inputs as other baselines: the current RGB-D observation, the executed camera pose, the observation history available through the benchmark manifest, and the current view-space feasibility constraint.
They output either an executable camera pose or a stop action through the common interaction interface.

\paragraph{Random and coverage-ordering baselines.}
Random+TSP is used as the simple non-learned coverage baseline in the reported experiments.
Its input is the feasible candidate-view set under the current view-space constraint.
It samples a finite set of candidate views and orders the selected views with a TSP-style path optimizer before execution.
This separates view-set selection from trajectory ordering and provides a simple reference for how far non-learned coverage over a dense candidate set can go under the same budget and reachability constraints.
The ordered plan is then executed one action at a time through the online benchmark interface.

Our implementation also includes farthest-point-style coverage variants that select spatially dispersed candidate views from the feasible view set.
These variants are retained in the codebase as additional simple baselines, but are not included in the main reported comparisons.
For both random and farthest-point variants, infeasible candidates are filtered using the benchmark feasibility API before ordering or execution.

\paragraph{Voxel information-gain baselines.}
We implement a voxel information-gain framework that supports multiple classical view-utility criteria.
The input to these methods is the accumulated RGB-D observation history.
Each method maintains an internal occupancy-style map in the normalized object frame, using a $64^3$ voxel grid over the workspace $[-1,1]^3$, and updates this map from the accumulated depth observations.
At each step, candidate views are scored by ray casting through the current voxel map and estimating the expected information gain of executing each view.
The selected candidate is then submitted as a benchmark move action and is evaluated only through the resulting RGB-D observation.

The implementation supports several information-gain criteria.
The CPU path supports six criteria, denoted OA, UV, RSE, APORA, and Kr in the codebase~\citep{delmerico2018comparison,daudelin2017adaptable,kriegel2015efficient}.
For dense candidate-view evaluation, we additionally provide a CUDA backend for the subset whose ray-wise scoring functions are simple to parallelize: OA, UV, RSE, and Kr.

In the reported experiments, we use the rear-side entropy variant, denoted VoxelIG-RSE, as the representative classical voxel information-gain baseline.
In the primary benchmark, VoxelIG-RSE searches over the canonical Tammes-128 candidate set at viewing radius $3.0$.
Before scoring and execution, candidates are filtered by the benchmark feasibility API, allowing the same implementation to run under both \texttt{whole} and \texttt{quarter} view-space constraints.
VoxelIG-RSE favors views that are expected to reduce uncertainty on currently unobserved or partially observed regions behind the visible surface.

We also implement a movement-aware variant, VoxelIG-RSE+Mov, which augments the information-gain score with a movement-cost penalty:
\begin{equation}
U(v)
=
\widehat{G}(v)
-
\lambda
\widehat{D}(v),
\end{equation}
where $\widehat{G}(v)$ is the normalized information-gain score, $\widehat{D}(v)$ is the normalized movement cost from the current camera pose to candidate view $v$, and $\lambda$ controls the trade-off between reconstruction gain and travel cost.

\paragraph{Oracle analysis baselines.}
For diagnostic analysis, we also include oracle baselines that access ground-truth coverage information to select candidate views.
Their input includes hidden coverage information that is unavailable to deployable methods, so they are not treated as ordinary baselines and are not included as deployable methods in the main comparison.
Instead, they provide upper-bound or sanity-check references for interpreting the gap between feasible online methods and ground-truth-aware coverage selection.

\subsection{Learned View-Prediction Baselines}

We adapt learned view-prediction baselines to the same online interaction protocol as the non-learned methods.
All learned baselines receive only released observations and benchmark-side pose information.
They do not access ground-truth geometry, hidden coverage sets, or evaluator internals.
At inference time, each learned method maintains its own method-specific state representation and submits one benchmark action at a time.

\paragraph{Discrete direct next-view prediction.}
NBV-Net is adapted as a direct next-view classification baseline over the canonical Tammes-128 candidate set.
The algorithm maintains an internal occupancy-style map from the accumulated RGB-D observations and exports a dense $64^3$ grid over the normalized workspace $[-1,1]^3$ at each step.
Unknown occupancy is initialized to $0.5$.
The planning-network service receives this grid and predicts a single next-view class over the $128$ candidate views:
\begin{equation}
\hat{i}_t
=
\arg\max_{i \in \{1,\ldots,128\}}
f_{\theta}
\left(
G_t
\right)_i,
\end{equation}
where $G_t$ denotes the current dense occupancy-grid input.

The predicted view index is converted to the corresponding benchmark camera pose at viewing radius $3.0$ and submitted as a move action if it is feasible and has not been visited.
If the predicted index is already visited, missing, out of range, or infeasible under the current view-space constraint, the method terminates with its native action-interface failure reason.
We do not replace NBV-Net's single-class output with a masked ranking policy, because doing so would change its deployment-time action semantics.

\paragraph{Learned set-cover prediction.}
MA-SCVP is adapted as a learned set-cover-style planner over the Tammes-128 candidate set.
Following the default MA-SCVP setting, the method is initialized from two observed views before predicting a residual set of complementary views.
Because the benchmark initial pose is not guaranteed to coincide exactly with one of the Tammes-128 candidate views, we first map the standardized initial pose to its nearest Tammes-128 candidate for the view-state representation.
The second initialization view is selected by a farthest-point rule over feasible Tammes-128 candidates, rather than by an NBV-Net prediction.
This makes the two-view initialization deterministic and stable under reachability constraints; using an unconstrained learned next-view predictor for the second view can otherwise introduce infeasible or unstable starts in constrained view spaces.

After the two-view initialization, MA-SCVP maintains a dense $64^3$ occupancy-grid representation over $[-1,1]^3$, together with a $128$-dimensional view-state vector.
The planning network predicts per-view scores for the residual view set:
\begin{equation}
s_t
=
f_{\theta}
\left(
G_t,\ m_t
\right)
\in
[0,1]^{128}.
\end{equation}
We decode a predicted residual view set by thresholding the scores with a fixed decode threshold 0.3.
The resulting set is filtered to remove already visited and infeasible views.
The remaining views are then ordered with a TSP-style path optimizer, and the method executes the resulting finite plan one action at a time.

MA-SCVP therefore differs from direct next-view methods in execution mode: it predicts a set of complementary views rather than a single next view.
Its native termination occurs when the decoded and filtered plan has been exhausted.

\paragraph{Boundary-based view prediction.}
BENBV-Net uses a dynamic, object-dependent candidate generation procedure instead of a fixed Tammes-128 action classifier.
At each step, the wrapper aggregates the observed partial point cloud, estimates surface normals, detects boundary points with PCL, and clusters them into a fixed number of boundary candidates.
In our implementation, we use $20$ boundary candidates, $4096$ sampled partial points, $k=30$ for local neighborhood processing, and a boundary angle threshold of $120^\circ$.

Each boundary candidate is mapped to a continuous camera-lookat pose, and the BENBV network scores the resulting candidate set:
\begin{equation}
s_t
=
f_{\theta}
\left(
P_t,\ S_t,\ C_t
\right)
\in
\mathbb{R}^{20},
\end{equation}
where $P_t$ denotes the partial point cloud input, and $S_t$ and $C_t$ denote the boundary-candidate state and context features used by the model.
The wrapper ranks all boundary candidates by score, queries the benchmark feasibility API for the corresponding continuous poses, and submits the highest-scoring feasible pose.

This boundary-based formulation provides a different action parameterization from fixed-view learned baselines.
It can adapt candidate poses to the observed object geometry, while still being evaluated through the same feasibility checks, RGB-D rendering, and official reconstruction metrics as all other methods.

\subsection{Shape Completion and Completion-Assisted Planning}

We include completion-assisted baselines to evaluate whether explicit shape prediction can improve view planning under budget and reachability constraints. 
These methods decouple geometry inference from view selection: a shape completion module first predicts a completed point cloud from the current partial observation, and a downstream planner then selects views using the predicted geometry. 
All completion-assisted methods are still evaluated online through the same benchmark runner. 
The completed shape is used only inside the algorithm for planning and is never used by the evaluator.

In the primary benchmark, completion-assisted planners search over the canonical Tammes-128 candidate-view set at viewing radius $3.0$.
Candidate views are filtered by the benchmark feasibility API before execution, allowing the same planners to run under both \texttt{whole} and \texttt{quarter} constraints.
The Tammes-360 sensitivity analysis is reported in Appendix~\ref{app:candidate_sensitivity}.

\paragraph{PoinTr-C completion module.}
We use PoinTr-C as the shared shape completion module. 
At each planning step, the wrapper constructs the accumulated partial point cloud from the executed RGB-D observations and sends it to the completion service. 
The partial point cloud is sampled or padded to $2{,}048$ input points, and the completion module predicts an $8{,}192$-point dense completed point cloud:
\begin{equation}
\widehat{P}_{\mathrm{comp},t}
=
f_{\theta}
\left(
P_t^{2048}
\right),
\qquad
\widehat{P}_{\mathrm{comp},t}
\in
\mathbb{R}^{8192 \times 3}.
\end{equation}
Here $P_t^{2048}$ denotes the accumulated partial observation point cloud at step $t$ after sampling to the model input size. 
The predicted point cloud is transformed into the same normalized object frame as the benchmark observations and is used to estimate unobserved target regions for planning. 
Completion inference time is counted as method-side runtime because it is part of the algorithm's decision process.

\paragraph{PoinTr-C+NBV.}
PoinTr-C+NBV uses the completed point cloud to define a one-step completion-aware next-best-view utility. 
The algorithm voxelizes both the accumulated observation point cloud and the predicted completed point cloud in the normalized object frame. 
Predicted voxels that are not yet observed are treated as planning targets:
\begin{equation}
U_t
=
\widehat{V}_{\mathrm{comp},t}
\setminus
V_{\mathrm{obs},t},
\end{equation}
where $\widehat{V}_{\mathrm{comp},t}$ is the voxelized completed shape and $V_{\mathrm{obs},t}$ is the voxelized observed surface. 
For each feasible Tammes-128 candidate view, the planner estimates how many target voxels would become visible from that view and selects a high-gain candidate. 
When multiple candidates have similar estimated gain, the planner prefers the candidate with lower movement cost from the current camera pose. 
In implementation, we select among candidates whose gain is within a fixed ratio of the best estimated gain and choose the shortest-motion candidate among them.

\paragraph{PoinTr-C+SCP.}
PoinTr-C+SCP formulates completion-assisted planning as a set-cover problem. 
Like MA-SCVP, it uses a two-view bootstrap before constructing the finite plan. 
The first view is the standardized benchmark initial view. 
The second view is selected deterministically by a farthest-point rule over feasible Tammes-128 candidates, which provides a stable initialization under constrained view spaces.

After the bootstrap observations, the algorithm invokes PoinTr-C and constructs a residual planning target from the completed shape. 
Rather than directly treating all predicted points as targets, we apply a lightweight filtering and voxelization procedure. 
The completed point cloud is first denoised with statistical outlier removal, then voxelized at planning resolution $0.03125$ within the normalized workspace $[-1,1]^3$. 
The current observed partial point cloud is voxelized with the same planning grid, and the planning reference is formed from the union of current and predicted geometry. 
The residual target universe is then defined as predicted-but-not-yet-observed voxels:
\begin{equation}
U_t
=
\widehat{V}_{\mathrm{comp},t}
\setminus
V_{\mathrm{obs},t}.
\end{equation}
This target construction reduces the influence of completion outliers and prevents already observed geometry from dominating the set-cover objective.

For each feasible Tammes-128 candidate view, the planner estimates the subset of this residual universe visible from that view using the same inverse-visibility coverage model as the difficulty annotation pipeline:
\begin{equation}
C_i
\subseteq
U_t.
\end{equation}
The planner then solves:
\begin{equation}
\begin{aligned}
\min_{\mathbf{x}}
\quad
&
\sum_i x_i
\\
\mathrm{s.t.}
\quad
&
\sum_{i:u\in C_i} x_i \geq 1,
\qquad
\forall u \in U_t,
\\
&
x_i \in \{0,1\}.
\end{aligned}
\end{equation}
The optimization is solved with a time limit of $10$ seconds. 
The selected views are ordered with a TSP-style path optimizer before execution. 
This method produces a finite plan and terminates when the selected plan is exhausted or no executable candidate remains.

\paragraph{PoinTr-C+MCP.}
PoinTr-C+MCP uses the same two-view bootstrap and residual target construction as PoinTr-C+SCP. 
After the standardized initial view, a second feasible view is selected by the same deterministic farthest-point rule over feasible Tammes-128 candidates. 
The accumulated partial point cloud is then completed by PoinTr-C, denoised, voxelized at planning resolution $0.03125$, and compared against the currently observed voxels to obtain the predicted-but-not-yet-observed target universe $U_t$. 
For each feasible Tammes-128 candidate view, we compute an estimated coverage set $C_i \subseteq U_t$.

Unlike SCP, MCP solves a budgeted maximum-coverage problem. 
Given a remaining view budget $B_t$, the planner selects at most $B_t$ views to maximize predicted target coverage:
\begin{equation}
\begin{aligned}
\max_{\mathbf{x}}
\quad
&
\left|
\bigcup_{i:x_i=1} C_i
\right|
\\
\mathrm{s.t.}
\quad
&
\sum_i x_i \leq B_t,
\\
&
x_i \in \{0,1\}.
\end{aligned}
\end{equation}
This formulation is explicitly budget-aware: it does not require covering all predicted residual targets, but instead chooses the subset of views expected to provide the highest coverage within the current budget.

If the optimized selected set contains fewer views than the available budget, we fill the remaining slots with a farthest-point-style selection over feasible, unvisited Tammes-128 candidates. 
This fallback prevents the finite plan from ending early merely because the predicted residual target has become small or sparse, and allows the method to continue exploring under fixed-budget evaluation. 
The final selected set is ordered with the same TSP-style path optimizer and executed one action at a time through the online interaction interface.

All completion-assisted planners use benchmark feasibility checks before execution. 
Therefore, predicted targets or candidate views that are not reachable under a constrained view space do not directly become executed actions. 
This makes the completion-assisted family a useful test of whether explicit shape inference plus classical planning remains robust when the executable view space is restricted.

\subsection{Training Data Generation}

We generate training data for learned baselines from cached object-centric observations and oracle coverage rollouts. 
The data-generation pipeline reuses CUDA-accelerated visibility routines used by the annotation pipeline. 
This avoids repeatedly rendering objects or recomputing expensive ray-casting queries during training-data export, making large-scale case generation practical.

\paragraph{Oracle rollouts from cached view coverage.}
For fixed-candidate learned baselines, we first precompute per-view observation caches over the Tammes-128 candidate set. 
Each cached view stores the surface voxels observed from that view under the same object-centric geometry and visibility convention used by the benchmark construction pipeline. 
Given these per-view coverage sets, we generate oracle rollouts for each object and each possible Tammes-128 start view.

At each rollout step, the generator maintains the set of already visited views and the set of covered voxels. 
It computes the marginal gain of every unvisited candidate view:
\begin{equation}
\Delta_i
=
\left|
C_i
\setminus
C_{\mathrm{obs}}
\right|,
\end{equation}
where $C_i$ is the cached coverage set of candidate view $i$ and $C_{\mathrm{obs}}$ is the currently covered voxel set. 
The oracle next-best-view label is the candidate with maximum marginal gain:
\begin{equation}
i^\star
=
\arg\max_i
\Delta_i .
\end{equation}
The rollout then executes this oracle view, updates the covered set, and continues until full coverage, zero marginal gain, or the maximum rollout length is reached.

In addition to the oracle next-view label, each rollout step also records the residual set-cover solution over the remaining uncovered voxels. 
This provides supervision for learned set-cover-style baselines: the residual target is the uncovered universe, and the label is the set of remaining candidate views selected by the set-cover solver.

\paragraph{Case export for NBV-Net and MA-SCVP.}
The oracle rollout data are converted into method-specific training cases. 
For each sampled step, we reconstruct the current partial observation state by merging the cached point clouds of the views already visited in the oracle rollout. 
We then export a dense occupancy grid, the accumulated partial point cloud, the visited-view state, and the corresponding supervision labels.

For direct next-view prediction, the target label is the oracle next-best-view index $i^\star$. 
For learned set-cover prediction, the target label is the residual set-cover view set. 
Concretely, each exported case contains the current occupancy grid, the accumulated observed point cloud, a $128$-dimensional visited-view state vector, an oracle next-view label, a per-view score vector derived from oracle marginal gains, and a residual set-cover view-id list.

We use method-specific case sampling to reduce redundancy while preserving each baseline's native supervision semantics. 
For NBV-Net, cases are sampled from complete start-view rollout groups, so each selected start view contributes a temporally coherent sequence of direct next-view prediction cases. 
For MA-SCVP, we follow the long-tail sampling strategy used in the original set-cover prediction pipeline. 
The selected MA-SCVP cases provide the occupancy grid, visited-view state, and residual set-cover view set used for learned view-set supervision.

The exported physical states are shared when possible. 
Instead of storing separate method-specific copies, the data generation pipeline writes one common state representation per selected rollout step and then uses method-specific case manifests to determine which fields are consumed by each baseline. 
This reduces storage duplication between NBV-Net and MA-SCVP while keeping their supervision targets distinct.

\paragraph{BENBV-Net training data.}
BENBV-Net uses a different data-generation pipeline because its action space is object-dependent rather than a fixed Tammes-128 classifier. 
For each object, we sample a start view from the Tammes-128 view set and run a coverage-driven rollout. 
At each step, the current partial point cloud is converted into a point-normal representation. 
We estimate normals, detect boundary points, and cluster the boundary points into $20$ candidate regions. 
Each candidate region defines a local target point and a corresponding view direction, yielding a dynamic set of boundary-based candidate camera poses.

For every generated boundary candidate, we estimate its visibility gain against the remaining uncovered surface. 
The candidate with the largest gain becomes the training label. 
Each BENBV training sequence stores sampled partial points, boundary-candidate features, context features, per-candidate supervision scores, the best candidate index, current coverage, candidate gain, overlap statistics, and a valid-candidate mask. 
In our implementation, each step uses $4096$ sampled partial points, $20$ boundary candidates, and $k=30$ local neighbors for boundary processing. 
Each rollout is capped at $128$ steps and may terminate earlier when coverage reaches $0.99$, when no valid candidate remains, or when the best candidate gain is below $10$ voxels.

\subsection{Training Hyperparameters and Model-Specific Details}

Network training was performed on NVIDIA RTX A6000 GPUs with 48 GB of VRAM.

\textbf{PointTr-C}. As a purely geometric method focused on shape completion, PoinTr-C is designed to infer and reconstruct the complete underlying surface from partial point cloud observations.
We preprocess our dataset following the protocol of the original PoinTr ShapeNet55 pipeline on Main-Train-Public. 
Specifically, each complete point cloud is converted into an \texttt{.npy} file containing 8192 points; point clouds with more than 8192 points are downsampled using farthest point sampling. 
The processed samples are then split into 9/1 for train/val.
We next follow the curriculum training strategy in Pred-NBV~\citep{dhami2023pred}, which gradually increases perturbation difficulty by expanding the rotation and translation ranges. 
The eight-stage rotation--translation schedule is $(25^\circ, 0.0)$, $(25^\circ, 0.1)$, $(45^\circ, 0.1)$, $(45^\circ, 0.25)$, $(45^\circ, 0.5)$, $(90^\circ, 0.5)$, $(180^\circ, 0.5)$, and $(360^\circ, 0.5)$, with stage durations of 15, 15, 20, 20, 25, 30, 35, and 40 epochs respectively, for a total of 200 epochs.
We train this shared completion module with a batch size of 96 and a learning rate of 0.0005.

\textbf{BENBV-Net}.
For each object, we randomly select four initial views from the Tammes-128 view set and run the BENBV-Net data-generation rollout on Main-Train-Public.
This produces 140,164 training cases, which are split 8/2 into training and validation sets.
We train BENBV-Net for 30 epochs with a batch size of 32 and a learning rate of 0.0005. 
We observe that BENBV-Net obtains low top-1 view prediction accuracy, while still achieving competitive performance under the final benchmark evaluation. 
This discrepancy suggests that top-1 accuracy alone can provide a biased assessment of NBV performance.

\textbf{NBV-Net and MA-SCVP.}
Motivated by the above observation, we further adapt the learned baselines that directly predict candidate viewpoints. 
For both NBV-Net and MA-SCVP, we use the previously discussed training data generation on Main-Train-Public, which contains 4995 objects. 
For NBV-Net, this process yields 630,155 samples, while for MA-SCVP it yields 628,801 samples. In both cases, the samples are split into training and validation sets using an 8:2 sample-level split.
For these methods, we augment the candidate-view representation with fixed positional embeddings derived from the Tammes-128 view ordering due to the hardness of 128 candidate-view discretization. 
The positional embeddings provide explicit information about the spatial arrangement of candidate viewpoints, helping the model distinguish geometrically nearby but distinct view directions and improving training stability.
Such positional information becomes important for dense candidate-view settings, where different view indices may be geometrically close but still correspond to distinct observation directions. 
For all three direct view-prediction baselines, we use a batch size of 8 and a learning rate of 0.0004.

\subsection{Implementation Notes for Attempted Baseline}

We also attempted to adapt PC-NBV~\citep{zeng2020pc} as a point-cloud-based view-scoring baseline over the Tammes-128 candidate set. 
For this adaptation, we used the same oracle-rollout training data generation pipeline as NBV-Net and MA-SCVP on Main-Train-Public, but replaced the occupancy-grid input with the accumulated partial point cloud. 
We also added the same fixed candidate-view positional encoding used by the direct learned view-prediction baselines, so that the model receives explicit information about the spatial arrangement of Tammes-128 candidate views.

However, under our current training setup, the model did not converge to a stable policy with reliable validation. 
Although validation metrics initially improved, reaching optimal loss and top-1 accuracy~(0.04) at epoch 7, training abruptly diverged at epoch 8 with a severe spike in validation loss (from 4.90 to 60.78). 
By epoch 9, both the loss values and model parameters collapsed to NaN. 
This divergence indicates that the training setting is unstable for PC-NBV when scaled to the large dataset with the introduction of position embeddings.
We therefore do not include PC-NBV in the main reported comparisons. 
This attempted adaptation is still informative: it suggests that direct point-cloud view-scoring baselines can be sensitive to the combination of object diversity, long-tail geometry, and candidate-view discretization in ObjView-Bench. 
We leave a more extensive re-training and architecture-specific tuning of PC-NBV to future work.

\section{Extended Benchmark Results}
\label{app:extended_benchmark_results}

\subsection{Full Results Across Evaluation Settings}
\label{app:full_results}

Tables~\ref{tab:app_budgeted_plan_results_transposed}--\ref{tab:app_iterative_results_transposed} report complete per-method results across acquisition budgets, stopping settings, and coverage thresholds under the canonical Tammes-128 protocol.
Candidate-set-compatible methods use Tammes-128, while BENBV-Net retains its native dynamic boundary-based action interface.
Each reported method--setting combination includes all 528 hidden-test objects.

Results are grouped by execution mode to preserve each method's native evaluation semantics.
Random+TSP and PoinTr-C+MCP are evaluated at their requested fixed budgets.
MA-SCVP and PoinTr-C+SCP are evaluated at native automatic termination.
RSE, RSE-Mov, NBV-Net, BENBV-Net, and PoinTr-C+NBV are evaluated at both fixed-budget and map-stabilization checkpoints.

Compared with the main-text summaries, these tables additionally include $K=10$, map-stabilization settings at voxel resolutions 0.01 and 0.03, and coverage metrics at distance thresholds $\tau \in \{0.01,0.02,0.03\}$.
They also report Chamfer Distance (CD), path cost, acquired views, and planning-runtime bins.
When an iterative method terminates before a requested checkpoint, its terminal state is used rather than discarding the episode.
Consequently, identical entries across checkpoints can reflect early termination rather than execution of the full requested budget.

All numerical entries are means over the evaluated objects.
Views include the standardized initial observation, while $K$ counts additional acquisitions.
Path cost is measured in the normalized object-centric coordinate frame.
Runtime bins are reported in seconds: $<0.1$, 0.1--0.5, 0.5--1, 1--3, 3--5, and $>5$.
The bins provide a coarse summary of planning cost and remain dependent on hardware and system load.
The candidate-discretization analysis in Section~\ref{app:candidate_sensitivity} focuses on the primary coverage threshold $\tau=0.02$ and the main-text budget and stopping settings.

\begin{table*}[t]
\centering
\scriptsize
\caption{Full results for budgeted finite-plan methods under the matched Tammes-128 candidate protocol, shown in a transposed per-method layout.
Each entry averages 528 hidden-test episodes.
Views include the initial observation; $K$ counts additional acquisitions.
Time bins are in seconds.
}
\label{tab:app_budgeted_plan_results_transposed}

\begin{tabular}{lcccccc}
\toprule
\multicolumn{7}{c}{\textbf{Random+TSP}} \\
\midrule
& \multicolumn{3}{c}{\texttt{whole}}
& \multicolumn{3}{c}{\texttt{quarter}} \\
\cmidrule(lr){2-4} \cmidrule(lr){5-7}
Metric & $K=5$ & $K=10$ & $K=30$
       & $K=5$ & $K=10$ & $K=30$ \\
\midrule
Views & 6.0 & 11.0 & 31.0 & 6.0 & 11.0 & 31.0 \\
NSC@0.01 $\uparrow$ & 0.905 & 0.971 & 0.994 & 0.926 & 0.966 & 0.994 \\
NSC@0.02 $\uparrow$ & 0.939 & 0.984 & 0.997 & 0.952 & 0.981 & 0.998 \\
NSC@0.03 $\uparrow$ & 0.957 & 0.990 & 0.998 & 0.965 & 0.988 & 0.999 \\
SC@0.01 $\uparrow$ & 0.825 & 0.889 & 0.912 & 0.666 & 0.701 & 0.737 \\
SC@0.02 $\uparrow$ & 0.883 & 0.928 & 0.943 & 0.747 & 0.778 & 0.808 \\
SC@0.03 $\uparrow$ & 0.915 & 0.948 & 0.959 & 0.800 & 0.827 & 0.851 \\
CD $\downarrow$ & 0.0015 & 0.0011 & 0.0010 & 0.0043 & 0.0034 & 0.0029 \\
Path $\downarrow$ & 14.08 & 22.00 & 42.21 & 9.67 & 14.16 & 29.29 \\
Time bin & $<0.1$ & $<0.1$ & $<0.1$ & $<0.1$ & $<0.1$ & $<0.1$ \\
\bottomrule
\end{tabular}

\vspace{0.8em}

\begin{tabular}{lcccccc}
\toprule
\multicolumn{7}{c}{\textbf{PoinTr-C+MCP}} \\
\midrule
& \multicolumn{3}{c}{\texttt{whole}}
& \multicolumn{3}{c}{\texttt{quarter}} \\
\cmidrule(lr){2-4} \cmidrule(lr){5-7}
Metric & $K=5$ & $K=10$ & $K=30$
       & $K=5$ & $K=10$ & $K=30$ \\
\midrule
Views & 6.0 & 11.0 & 31.0 & 6.0 & 11.0 & 31.0 \\
NSC@0.01 $\uparrow$ & 0.950 & 0.981 & 0.994 & 0.967 & 0.983 & 0.994 \\
NSC@0.02 $\uparrow$ & 0.973 & 0.991 & 0.997 & 0.983 & 0.993 & 0.998 \\
NSC@0.03 $\uparrow$ & 0.983 & 0.995 & 0.998 & 0.990 & 0.996 & 0.999 \\
SC@0.01 $\uparrow$ & 0.868 & 0.899 & 0.913 & 0.706 & 0.723 & 0.737 \\
SC@0.02 $\uparrow$ & 0.917 & 0.936 & 0.944 & 0.785 & 0.797 & 0.808 \\
SC@0.03 $\uparrow$ & 0.942 & 0.955 & 0.960 & 0.833 & 0.843 & 0.851 \\
CD $\downarrow$ & 0.0011 & 0.0010 & 0.0010 & 0.0032 & 0.0031 & 0.0029 \\
Path $\downarrow$ & 19.72 & 27.73 & 52.68 & 15.34 & 20.32 & 34.36 \\
Time bin & 0.5--1 & 0.5--1 & 0.5--1
         & 0.1--0.5 & 0.1--0.5 & 0.1--0.5 \\
\bottomrule
\end{tabular}
\end{table*}

\begin{table*}[t]
\centering
\scriptsize
\caption{Full results for native finite set-cover methods under the matched Tammes-128 candidate protocol, shown in a transposed per-method layout.
Results are evaluated at native automatic termination and averaged over 528 hidden-test episodes per constraint.
Views include the initial observation; time bins are in seconds.
}
\label{tab:app_setcover_plan_results_transposed}

\begin{tabular}{lcc}
\toprule
\multicolumn{3}{c}{\textbf{MA-SCVP}} \\
\midrule
Metric & \texttt{whole} & \texttt{quarter} \\
\midrule
Views & 49.5 & 12.5 \\
NSC@0.01 $\uparrow$ & 0.980 & 0.938 \\
NSC@0.02 $\uparrow$ & 0.990 & 0.967 \\
NSC@0.03 $\uparrow$ & 0.994 & 0.979 \\
SC@0.01 $\uparrow$ & 0.900 & 0.687 \\
SC@0.02 $\uparrow$ & 0.937 & 0.772 \\
SC@0.03 $\uparrow$ & 0.956 & 0.823 \\
CD $\downarrow$ & 0.0010 & 0.0032 \\
Path $\downarrow$ & 57.19 & 18.19 \\
Time bin & 0.1--0.5 & 0.1--0.5 \\
\bottomrule
\end{tabular}
\hspace{2em}
\begin{tabular}{lcc}
\toprule
\multicolumn{3}{c}{\textbf{PoinTr-C+SCP}} \\
\midrule
Metric & \texttt{whole} & \texttt{quarter} \\
\midrule
Views & 15.9 & 14.8 \\
NSC@0.01 $\uparrow$ & 0.980 & 0.982 \\
NSC@0.02 $\uparrow$ & 0.991 & 0.994 \\
NSC@0.03 $\uparrow$ & 0.995 & 0.997 \\
SC@0.01 $\uparrow$ & 0.898 & 0.723 \\
SC@0.02 $\uparrow$ & 0.937 & 0.799 \\
SC@0.03 $\uparrow$ & 0.956 & 0.844 \\
CD $\downarrow$ & 0.0010 & 0.0030 \\
Path $\downarrow$ & 30.75 & 21.73 \\
Time bin & 0.5--1 & 0.1--0.5 \\
\bottomrule
\end{tabular}
\end{table*}

\begin{table*}[p]
\centering
\scriptsize
\setlength{\tabcolsep}{3.6pt}
\caption{Full results for iterative replanning methods under the matched Tammes-128 protocol, shown in a transposed per-method layout.
BENBV-Net retains its native dynamic boundary-derived candidates rather than using a fixed Tammes-128 action selector.
Each entry averages all 528 hidden-test episodes, using the terminal state when a requested checkpoint is not reached.
Views include the initial observation; time bins are in seconds.
}
\label{tab:app_iterative_results_transposed}

\resizebox{\textwidth}{!}{
\begin{tabular}{lcccccc|cccccc}
\toprule
\multicolumn{13}{c}{\textbf{RSE}} \\
\midrule
& \multicolumn{6}{c|}{\texttt{whole}}
& \multicolumn{6}{c}{\texttt{quarter}} \\
\cmidrule(lr){2-7} \cmidrule(lr){8-13}
Metric & $K=5$ & $K=10$ & $K=30$
& MS@0.01 & MS@0.02 & MS@0.03
& $K=5$ & $K=10$ & $K=30$
& MS@0.01 & MS@0.02 & MS@0.03 \\
\midrule
Views & 6.0 & 11.0 & 31.0 & 11.6 & 9.8 & 8.9 & 6.0 & 11.0 & 31.0 & 11.6 & 9.8 & 9.0 \\
NSC@0.01 $\uparrow$ & 0.961 & 0.982 & 0.992 & 0.984 & 0.982 & 0.979 & 0.971 & 0.985 & 0.994 & 0.987 & 0.984 & 0.983 \\
NSC@0.02 $\uparrow$ & 0.979 & 0.991 & 0.997 & 0.993 & 0.991 & 0.990 & 0.986 & 0.993 & 0.998 & 0.994 & 0.993 & 0.993 \\
NSC@0.03 $\uparrow$ & 0.988 & 0.995 & 0.998 & 0.996 & 0.995 & 0.995 & 0.991 & 0.996 & 0.999 & 0.997 & 0.996 & 0.996 \\
SC@0.01 $\uparrow$ & 0.880 & 0.900 & 0.912 & 0.902 & 0.899 & 0.897 & 0.711 & 0.726 & 0.738 & 0.728 & 0.725 & 0.724 \\
SC@0.02 $\uparrow$ & 0.923 & 0.936 & 0.943 & 0.937 & 0.936 & 0.935 & 0.789 & 0.800 & 0.809 & 0.801 & 0.799 & 0.799 \\
SC@0.03 $\uparrow$ & 0.946 & 0.955 & 0.959 & 0.956 & 0.955 & 0.954 & 0.836 & 0.845 & 0.852 & 0.846 & 0.844 & 0.844 \\
CD $\downarrow$ & 0.0011 & 0.0011 & 0.0010 & 0.0010 & 0.0010 & 0.0011 & 0.0031 & 0.0031 & 0.0029 & 0.0030 & 0.0030 & 0.0031 \\
Path $\downarrow$ & 25.70 & 47.06 & 104.85 & 47.94 & 41.71 & 38.35 & 20.36 & 34.98 & 79.08 & 36.14 & 31.79 & 29.78 \\
Time bin & 0.5--1 & 1--3 & 3--5 & 1--3 & 1--3 & 1--3 & 0.5--1 & 1--3 & 3--5 & 1--3 & 1--3 & 0.5--1 \\
\bottomrule
\end{tabular}
}


\resizebox{\textwidth}{!}{
\begin{tabular}{lcccccc|cccccc}
\toprule
\multicolumn{13}{c}{\textbf{RSE-Mov}} \\
\midrule
& \multicolumn{6}{c|}{\texttt{whole}}
& \multicolumn{6}{c}{\texttt{quarter}} \\
\cmidrule(lr){2-7} \cmidrule(lr){8-13}
Metric & $K=5$ & $K=10$ & $K=30$
& MS@0.01 & MS@0.02 & MS@0.03
& $K=5$ & $K=10$ & $K=30$
& MS@0.01 & MS@0.02 & MS@0.03 \\
\midrule
Views & 6.0 & 11.0 & 31.0 & 13.7 & 12.1 & 11.5 & 6.0 & 11.0 & 31.0 & 12.9 & 11.4 & 10.9 \\
NSC@0.01 $\uparrow$ & 0.839 & 0.977 & 0.991 & 0.983 & 0.981 & 0.980 & 0.922 & 0.984 & 0.994 & 0.987 & 0.986 & 0.984 \\
NSC@0.02 $\uparrow$ & 0.891 & 0.988 & 0.996 & 0.992 & 0.991 & 0.990 & 0.951 & 0.993 & 0.998 & 0.995 & 0.994 & 0.993 \\
NSC@0.03 $\uparrow$ & 0.921 & 0.993 & 0.998 & 0.995 & 0.995 & 0.994 & 0.966 & 0.996 & 0.999 & 0.997 & 0.996 & 0.996 \\
SC@0.01 $\uparrow$ & 0.764 & 0.895 & 0.911 & 0.902 & 0.899 & 0.898 & 0.670 & 0.725 & 0.738 & 0.728 & 0.726 & 0.725 \\
SC@0.02 $\uparrow$ & 0.836 & 0.932 & 0.942 & 0.937 & 0.936 & 0.935 & 0.753 & 0.799 & 0.809 & 0.801 & 0.800 & 0.799 \\
SC@0.03 $\uparrow$ & 0.879 & 0.952 & 0.959 & 0.955 & 0.955 & 0.954 & 0.807 & 0.844 & 0.852 & 0.846 & 0.845 & 0.845 \\
CD $\downarrow$ & 0.0020 & 0.0011 & 0.0010 & 0.0011 & 0.0011 & 0.0011 & 0.0036 & 0.0031 & 0.0029 & 0.0030 & 0.0030 & 0.0030 \\
Path $\downarrow$ & 10.37 & 30.16 & 84.93 & 39.00 & 33.89 & 31.85 & 10.15 & 24.62 & 63.41 & 28.82 & 25.53 & 24.07 \\
Time bin & 0.5--1 & 1--3 & 3--5 & 1--3 & 1--3 & 1--3 & 0.5--1 & 1--3 & 3--5 & 1--3 & 1--3 & 1--3 \\
\bottomrule
\end{tabular}
}


\resizebox{\textwidth}{!}{
\begin{tabular}{lcccccc|cccccc}
\toprule
\multicolumn{13}{c}{\textbf{NBV-Net}} \\
\midrule
& \multicolumn{6}{c|}{\texttt{whole}}
& \multicolumn{6}{c}{\texttt{quarter}} \\
\cmidrule(lr){2-7} \cmidrule(lr){8-13}
Metric & $K=5$ & $K=10$ & $K=30$
& MS@0.01 & MS@0.02 & MS@0.03
& $K=5$ & $K=10$ & $K=30$
& MS@0.01 & MS@0.02 & MS@0.03 \\
\midrule
Views & 6.0 & 9.2 & 9.9 & 9.2 & 8.4 & 8.0 & 1.2 & 1.2 & 1.2 & 1.2 & 1.2 & 1.2 \\
NSC@0.01 $\uparrow$ & 0.961 & 0.977 & 0.977 & 0.977 & 0.976 & 0.975 & 0.415 & 0.415 & 0.415 & 0.415 & 0.415 & 0.415 \\
NSC@0.02 $\uparrow$ & 0.978 & 0.988 & 0.988 & 0.988 & 0.987 & 0.987 & 0.472 & 0.472 & 0.472 & 0.472 & 0.472 & 0.472 \\
NSC@0.03 $\uparrow$ & 0.987 & 0.993 & 0.993 & 0.993 & 0.993 & 0.992 & 0.515 & 0.515 & 0.515 & 0.515 & 0.515 & 0.515 \\
SC@0.01 $\uparrow$ & 0.879 & 0.894 & 0.895 & 0.895 & 0.894 & 0.893 & 0.313 & 0.313 & 0.313 & 0.313 & 0.313 & 0.313 \\
SC@0.02 $\uparrow$ & 0.921 & 0.931 & 0.932 & 0.932 & 0.931 & 0.931 & 0.387 & 0.387 & 0.387 & 0.387 & 0.387 & 0.387 \\
SC@0.03 $\uparrow$ & 0.945 & 0.951 & 0.952 & 0.951 & 0.951 & 0.951 & 0.443 & 0.443 & 0.443 & 0.443 & 0.443 & 0.443 \\
CD $\downarrow$ & 0.0011 & 0.0011 & 0.0011 & 0.0011 & 0.0011 & 0.0011 & 0.0726 & 0.0726 & 0.0726 & 0.0726 & 0.0726 & 0.0726 \\
Path $\downarrow$ & 24.77 & 39.16 & 41.73 & 38.96 & 35.76 & 33.69 & 0.90 & 0.90 & 0.90 & 0.90 & 0.90 & 0.90 \\
Time bin & 0.5--1 & 1--3 & 1--3 & 1--3 & 1--3 & 1--3 & 0.1--0.5 & 0.1--0.5 & 0.1--0.5 & 0.1--0.5 & 0.1--0.5 & 0.1--0.5 \\
\bottomrule
\end{tabular}
}


\resizebox{\textwidth}{!}{
\begin{tabular}{lcccccc|cccccc}
\toprule
\multicolumn{13}{c}{\textbf{BENBV-Net}} \\
\midrule
& \multicolumn{6}{c|}{\texttt{whole}}
& \multicolumn{6}{c}{\texttt{quarter}} \\
\cmidrule(lr){2-7} \cmidrule(lr){8-13}
Metric & $K=5$ & $K=10$ & $K=30$
& MS@0.01 & MS@0.02 & MS@0.03
& $K=5$ & $K=10$ & $K=30$
& MS@0.01 & MS@0.02 & MS@0.03 \\
\midrule
Views & 6.0 & 10.6 & 28.0 & 11.8 & 10.4 & 9.9 & 5.8 & 10.0 & 23.9 & 10.9 & 9.8 & 9.3 \\
NSC@0.01 $\uparrow$ & 0.927 & 0.970 & 0.988 & 0.975 & 0.971 & 0.970 & 0.915 & 0.947 & 0.966 & 0.950 & 0.947 & 0.945 \\
NSC@0.02 $\uparrow$ & 0.956 & 0.985 & 0.995 & 0.988 & 0.986 & 0.985 & 0.945 & 0.967 & 0.979 & 0.969 & 0.967 & 0.966 \\
NSC@0.03 $\uparrow$ & 0.971 & 0.991 & 0.997 & 0.993 & 0.992 & 0.992 & 0.960 & 0.977 & 0.985 & 0.978 & 0.977 & 0.976 \\
SC@0.01 $\uparrow$ & 0.847 & 0.888 & 0.907 & 0.893 & 0.889 & 0.887 & 0.661 & 0.687 & 0.709 & 0.690 & 0.687 & 0.685 \\
SC@0.02 $\uparrow$ & 0.900 & 0.929 & 0.940 & 0.932 & 0.930 & 0.929 & 0.743 & 0.766 & 0.784 & 0.768 & 0.766 & 0.764 \\
SC@0.03 $\uparrow$ & 0.929 & 0.950 & 0.958 & 0.952 & 0.951 & 0.951 & 0.797 & 0.816 & 0.831 & 0.818 & 0.816 & 0.815 \\
CD $\downarrow$ & 0.0019 & 0.0010 & 0.0010 & 0.0010 & 0.0010 & 0.0010 & 0.0040 & 0.0037 & 0.0035 & 0.0037 & 0.0037 & 0.0037 \\
Path $\downarrow$ & 15.64 & 27.63 & 66.98 & 30.84 & 27.49 & 26.23 & 11.08 & 19.25 & 44.11 & 21.20 & 19.10 & 18.14 \\
Time bin & 0.5--1 & 1--3 & 3--5 & 1--3 & 1--3 & 1--3 & 0.5--1 & 1--3 & 1--3 & 1--3 & 1--3 & 0.5--1 \\
\bottomrule
\end{tabular}
}


\resizebox{\textwidth}{!}{
\begin{tabular}{lcccccc|cccccc}
\toprule
\multicolumn{13}{c}{\textbf{PoinTr-C+NBV}} \\
\midrule
& \multicolumn{6}{c|}{\texttt{whole}}
& \multicolumn{6}{c}{\texttt{quarter}} \\
\cmidrule(lr){2-7} \cmidrule(lr){8-13}
Metric & $K=5$ & $K=10$ & $K=30$
& MS@0.01 & MS@0.02 & MS@0.03
& $K=5$ & $K=10$ & $K=30$
& MS@0.01 & MS@0.02 & MS@0.03 \\
\midrule
Views & 6.0 & 11.0 & 31.0 & 14.4 & 12.0 & 11.2 & 6.0 & 11.0 & 31.0 & 12.6 & 10.7 & 9.9 \\
NSC@0.01 $\uparrow$ & 0.864 & 0.956 & 0.994 & 0.975 & 0.964 & 0.960 & 0.932 & 0.976 & 0.994 & 0.979 & 0.973 & 0.971 \\
NSC@0.02 $\uparrow$ & 0.908 & 0.975 & 0.998 & 0.986 & 0.979 & 0.976 & 0.956 & 0.987 & 0.998 & 0.988 & 0.985 & 0.984 \\
NSC@0.03 $\uparrow$ & 0.930 & 0.982 & 0.999 & 0.991 & 0.985 & 0.983 & 0.967 & 0.992 & 0.999 & 0.992 & 0.989 & 0.989 \\
SC@0.01 $\uparrow$ & 0.786 & 0.874 & 0.913 & 0.893 & 0.882 & 0.878 & 0.678 & 0.716 & 0.738 & 0.719 & 0.713 & 0.711 \\
SC@0.02 $\uparrow$ & 0.852 & 0.919 & 0.944 & 0.931 & 0.924 & 0.921 & 0.759 & 0.791 & 0.809 & 0.794 & 0.789 & 0.787 \\
SC@0.03 $\uparrow$ & 0.887 & 0.941 & 0.960 & 0.951 & 0.945 & 0.942 & 0.810 & 0.838 & 0.852 & 0.839 & 0.836 & 0.834 \\
CD $\downarrow$ & 0.0028 & 0.0013 & 0.0010 & 0.0011 & 0.0012 & 0.0012 & 0.0039 & 0.0032 & 0.0029 & 0.0032 & 0.0032 & 0.0032 \\
Path $\downarrow$ & 17.87 & 35.23 & 100.65 & 46.31 & 38.61 & 35.86 & 14.33 & 26.10 & 67.88 & 29.52 & 25.25 & 23.62 \\
Time bin & 1--3 & 3--5 & $>5$ & 3--5 & 3--5 & 3--5 & 1--3 & 1--3 & $>5$ & 3--5 & 1--3 & 1--3 \\
\bottomrule
\end{tabular}
}

\end{table*}

The complete tables reveal additional budget, metric-tolerance, and stopping-sensitivity patterns beyond the main-text summaries.
The following per-method observations concern the evaluated implementations and protocols.

\paragraph{Random+TSP.}
The additional $K=10$ checkpoint shows how coverage develops between the small- and large-budget settings.
In the \texttt{whole} setting, NSC@0.02 increases from 0.939 to 0.984 and 0.997 at $K=5$, $K=10$, and $K=30$, respectively.
The corresponding path costs are 14.08, 22.00, and 42.21.
Thus, the coverage gain per additional acquisition decreases as coverage approaches saturation.

\paragraph{PoinTr-C+MCP.}
At $K=10$, PoinTr-C+MCP already reaches NSC@0.02 of 0.991 under \texttt{whole} and 0.993 under \texttt{quarter}.
Increasing the budget to $K=30$ further improves normalized coverage, but raises path cost from 27.73 to 52.68 under \texttt{whole} and from 20.32 to 34.36 under \texttt{quarter}.
The stricter NSC@0.01 metric retains a larger coverage deficit at small budgets than NSC@0.03.

\paragraph{MA-SCVP.}
At native automatic termination, MA-SCVP achieves NSC values of 0.980, 0.990, and 0.994 at thresholds 0.01, 0.02, and 0.03 under \texttt{whole}.
The corresponding values under \texttt{quarter} are 0.938, 0.967, and 0.979.
The larger variation across thresholds under \texttt{quarter} shows that the assessed final coverage is more sensitive to evaluation tolerance in this constrained setting.
These entries describe the final executed plan rather than separate stopping decisions at each coverage threshold.

\paragraph{PoinTr-C+SCP.}
PoinTr-C+SCP executes 15.9 views under \texttt{whole} and 14.8 under \texttt{quarter}.
Its NSC values increase from 0.980 to 0.995 across evaluation thresholds 0.01--0.03 under \texttt{whole}, and from 0.982 to 0.997 under \texttt{quarter}.
The method constructs its finite plan over feasible candidates; these results do not involve iterative replanning after each acquisition.
As with MA-SCVP, coverage thresholds change the evaluation of the terminal reconstruction rather than its native termination rule.

\paragraph{RSE.}
RSE's map-stabilization checkpoint changes with the voxel resolution used by the stopping criterion.
Under \texttt{whole}, mean acquired views decrease from 11.6 at $\mathrm{MS}@0.01$ to 9.8 at $\mathrm{MS}@0.02$ and 8.9 at $\mathrm{MS}@0.03$.
NSC@0.02 changes from 0.993 to 0.991 and 0.990 across these checkpoints.
The tables therefore expose the additional acquisition and motion cost associated with the finer stabilization criterion.

\paragraph{RSE-Mov.}
RSE-Mov exhibits a pronounced improvement between $K=5$ and $K=10$ under \texttt{whole}, with NSC@0.02 increasing from 0.891 to 0.988.
Its map-stabilization checkpoints occur at 13.7, 12.1, and 11.5 views for resolutions 0.01, 0.02, and 0.03, respectively.
At $\mathrm{MS}@0.02$, its unrounded NSC@0.02 is approximately 0.990847, slightly below RSE's 0.991411, while its path cost is lower at 33.89 versus 41.71.
Their identical three-decimal NSC entries should therefore not be interpreted as identical reconstruction quality.

\paragraph{NBV-Net.}
Under \texttt{whole}, NBV-Net executes 9.2 views on average at the $K=10$ evaluation and 9.9 at the $K=30$ evaluation because some trajectories terminate before reaching the requested budgets.
The two settings both round to 0.988 NSC@0.02 despite having different mean trajectory lengths.
Under \texttt{quarter}, all requested fixed-budget and map-stabilization entries use the terminal fallback.
The repeated values across these columns therefore reflect early termination, not insensitivity to the requested budget or stopping resolution.

\paragraph{BENBV-Net.}
BENBV-Net retains its native dynamic boundary-derived candidates throughout these evaluations.
Under \texttt{whole}, its map-stabilization checkpoints occur at 11.8, 10.4, and 9.9 acquired views for resolutions 0.01, 0.02, and 0.03, respectively.
Under \texttt{quarter}, the corresponding counts are 10.9, 9.8, and 9.3.
Its fixed-budget mean view counts can also fall below the requested cap, reaching 28.0 under \texttt{whole} and 23.9 under \texttt{quarter} at $K=30$.
These entries retain episodes that terminate before the requested checkpoint.

\paragraph{PoinTr-C+NBV.}
The intermediate $K=10$ checkpoint reaches NSC@0.02 of 0.975 under \texttt{whole} and 0.987 under \texttt{quarter}, compared with 0.908 and 0.956 at $K=5$.
Under \texttt{whole}, changing the stabilization criterion from $\mathrm{MS}@0.01$ to $\mathrm{MS}@0.03$ reduces mean acquired views from 14.4 to 11.2 and path cost from 46.31 to 35.86.
NSC@0.02 decreases from 0.986 to 0.976 across the same checkpoints.
This provides an additional example of how stopping resolution changes both the attained coverage and deployment cost.

\subsection{Sensitivity to Candidate-View Discretization}
\label{app:candidate_sensitivity}

This section examines the sensitivity of the reported budget- and reachability-related findings to candidate-view discretization.
We use the canonical Tammes-128 protocol adopted in the main text as the reference condition and compare it with a denser Tammes-360 candidate discretization for compatible planners.
Both conditions are evaluated over the same 528 hidden-test objects.
Tables~\ref{tab:app_candidate_quality}--\ref{tab:app_candidate_reachability} report the complete method comparisons, including absolute performance and numerical ordering.

\paragraph{Candidate discretization settings.}
Under the canonical condition, all candidate-set-compatible methods use Tammes-128.
Under the denser condition, Random+TSP, RSE, RSE-Mov, PoinTr-C+NBV, PoinTr-C+SCP, and PoinTr-C+MCP use Tammes-360.
NBV-Net and MA-SCVP retain their trained Tammes-128 action spaces in both conditions, and their results are reused.

BENBV-Net retains its native dynamic boundary-based action interface and is marked with $\dagger$ throughout the tables.
Tammes-128 is used only for its rollout initialization, while subsequent candidates are generated from observed boundary geometry.
Its results are also reused across conditions.
The panel labels therefore specify the candidate discretization of compatible planners rather than a common action space imposed on every method.

The two conditions share the hidden-test objects, standardized initialization, budget definitions, view-space constraints, and constraint-specific Tammes-360 observable-surface reference.
The comparison changes the planning candidate discretization while retaining the evaluation reference.
It measures sensitivity to both candidate count and the available viewing directions rather than isolating candidate count alone.

\paragraph{Evaluation settings and table interpretation.}
Fixed-budget comparisons use $K=5$ and $K=30$ additional acquisitions.
Automatic evaluation uses $\mathrm{MS}@0.02$ for iterative methods and native termination for finite set-cover methods.
Budgeted finite-plan methods are evaluated only at their requested budgets, while native finite set-cover methods appear only under Automatic.
When an iterative method terminates before a requested checkpoint, its terminal state is retained.

Table~\ref{tab:app_candidate_quality} ranks the methods evaluated under each setting by mean NSC@0.02.
Table~\ref{tab:app_candidate_path} first requires mean NSC@0.02 $\geq 0.95$ and then ranks the qualifying methods by mean path cost.
Methods below this aggregate quality threshold remain visible but receive no path rank.
The threshold is applied to the method-level mean and does not require every individual episode to exceed 0.95.
Views include the initial observation, and NSC is reported to six decimal places to expose small numerical differences.
These complete rankings serve as supplementary diagnostics rather than the family-representative summaries used in the main text.

\begin{table*}[p]
\centering
\scriptsize
\setlength{\tabcolsep}{4pt}
\caption{Sensitivity of whole-space quality rankings to candidate-view discretization.
Panel A reports the canonical Tammes-128 condition used in the main text; Panel B uses denser Tammes-360 candidates for compatible planners.
Methods are ranked by mean NSC@0.02 within each setting.
Each entry averages 528 hidden-test episodes.
Views include the initial observation, and time bins are in seconds.
NBV-Net and MA-SCVP retain their trained action spaces, while BENBV-Net$^\dagger$ retains its native dynamic interface; their results are reused across panels.}
\label{tab:app_candidate_quality}

\begin{tabular}{lrllrrrr}
\toprule
\multicolumn{8}{c}{\textbf{Panel A: Canonical candidate discretization (Tammes-128)}} \\
\midrule
Setting & Rank & Method & Family
& Views & NSC@0.02 & Path & Time bin \\
\midrule
$K=5$ & 1 & RSE & Classical & 6.0 & 0.979472 & 25.70 & 0.5--1 \\
$K=5$ & 2 & NBV-Net & Learned & 6.0 & 0.978313 & 24.77 & 0.5--1 \\
$K=5$ & 3 & PoinTr-C+MCP & Hybrid & 6.0 & 0.973472 & 19.72 & 0.5--1 \\
$K=5$ & 4 & BENBV-Net$^\dagger$ & Learned & 6.0 & 0.956392 & 15.64 & 0.5--1 \\
$K=5$ & 5 & Random+TSP & Classical & 6.0 & 0.938983 & 14.08 & $<0.1$ \\
$K=5$ & 6 & PoinTr-C+NBV & Hybrid & 6.0 & 0.907701 & 17.87 & 1--3 \\
$K=5$ & 7 & RSE-Mov & Classical & 6.0 & 0.890913 & 10.37 & 0.5--1 \\
\midrule
$K=30$ & 1 & PoinTr-C+NBV & Hybrid & 31.0 & 0.997657 & 100.65 & $>5$ \\
$K=30$ & 2 & Random+TSP & Classical & 31.0 & 0.997296 & 42.21 & $<0.1$ \\
$K=30$ & 3 & PoinTr-C+MCP & Hybrid & 31.0 & 0.997187 & 52.68 & 0.5--1 \\
$K=30$ & 4 & RSE & Classical & 31.0 & 0.996760 & 104.85 & 3--5 \\
$K=30$ & 5 & RSE-Mov & Classical & 31.0 & 0.995873 & 84.93 & 3--5 \\
$K=30$ & 6 & BENBV-Net$^\dagger$ & Learned & 28.0 & 0.994973 & 66.98 & 3--5 \\
$K=30$ & 7 & NBV-Net & Learned & 9.9 & 0.987945 & 41.73 & 1--3 \\
\midrule
Automatic & 1 & RSE & Classical & 9.8 & 0.991411 & 41.71 & 1--3 \\
Automatic & 2 & PoinTr-C+SCP & Hybrid & 15.9 & 0.991254 & 30.75 & 0.5--1 \\
Automatic & 3 & RSE-Mov & Classical & 12.1 & 0.990847 & 33.89 & 1--3 \\
Automatic & 4 & MA-SCVP & Learned & 49.5 & 0.990456 & 57.19 & 0.1--0.5 \\
Automatic & 5 & NBV-Net & Learned & 8.4 & 0.987454 & 35.76 & 1--3 \\
Automatic & 6 & BENBV-Net$^\dagger$ & Learned & 10.4 & 0.985903 & 27.49 & 1--3 \\
Automatic & 7 & PoinTr-C+NBV & Hybrid & 12.0 & 0.979172 & 38.61 & 3--5 \\
\bottomrule
\end{tabular}

\vspace{0.8em}

\begin{tabular}{lrllrrrr}
\toprule
\multicolumn{8}{c}{\textbf{Panel B: Denser candidate discretization (Tammes-360)}} \\
\midrule
Setting & Rank & Method & Family
& Views & NSC@0.02 & Path & Time bin \\
\midrule
$K=5$ & 1 & RSE & Classical & 6.0 & 0.979525 & 25.66 & 0.5--1 \\
$K=5$ & 2 & NBV-Net & Learned & 6.0 & 0.978313 & 24.77 & 0.5--1 \\
$K=5$ & 3 & PoinTr-C+MCP & Hybrid & 6.0 & 0.976531 & 20.28 & 1--3 \\
$K=5$ & 4 & BENBV-Net$^\dagger$ & Learned & 6.0 & 0.956392 & 15.64 & 0.5--1 \\
$K=5$ & 5 & Random+TSP & Classical & 6.0 & 0.944962 & 14.16 & $<0.1$ \\
$K=5$ & 6 & PoinTr-C+NBV & Hybrid & 6.0 & 0.885928 & 17.19 & 1--3 \\
$K=5$ & 7 & RSE-Mov & Classical & 6.0 & 0.734758 & 5.75 & 1--3 \\
\midrule
$K=30$ & 1 & PoinTr-C+MCP & Hybrid & 31.0 & 0.997787 & 51.35 & 1--3 \\
$K=30$ & 2 & Random+TSP & Classical & 31.0 & 0.996438 & 40.54 & $<0.1$ \\
$K=30$ & 3 & PoinTr-C+NBV & Hybrid & 31.0 & 0.995689 & 99.11 & $>5$ \\
$K=30$ & 4 & BENBV-Net$^\dagger$ & Learned & 28.0 & 0.994973 & 66.98 & 3--5 \\
$K=30$ & 5 & RSE & Classical & 31.0 & 0.994842 & 90.94 & $>5$ \\
$K=30$ & 6 & RSE-Mov & Classical & 31.0 & 0.994724 & 68.98 & $>5$ \\
$K=30$ & 7 & NBV-Net & Learned & 9.9 & 0.987945 & 41.73 & 1--3 \\
\midrule
Automatic & 1 & PoinTr-C+SCP & Hybrid & 15.5 & 0.992507 & 30.17 & 0.5--1 \\
Automatic & 2 & RSE & Classical & 9.6 & 0.990593 & 40.70 & 1--3 \\
Automatic & 3 & MA-SCVP & Learned & 49.5 & 0.990456 & 57.19 & 0.1--0.5 \\
Automatic & 4 & NBV-Net & Learned & 8.4 & 0.987454 & 35.76 & 1--3 \\
Automatic & 5 & BENBV-Net$^\dagger$ & Learned & 10.4 & 0.985903 & 27.49 & 1--3 \\
Automatic & 6 & RSE-Mov & Classical & 14.1 & 0.984623 & 31.92 & 3--5 \\
Automatic & 7 & PoinTr-C+NBV & Hybrid & 11.7 & 0.959838 & 36.21 & 3--5 \\
\bottomrule
\end{tabular}
\end{table*}

\paragraph{Quality ordering and numerical differences.}
At $K=5$, the complete quality ordering is identical under the two discretizations, with RSE, NBV-Net, and PoinTr-C+MCP occupying the first three positions.
However, unchanged ordering does not imply unchanged performance.
RSE-Mov achieves NSC@0.02 of 0.890913 under the canonical Tammes-128 condition, compared with 0.734758 under the denser Tammes-360 condition.
PoinTr-C+NBV similarly achieves 0.907701 versus 0.885928.
Conversely, Random+TSP achieves 0.938983 under the canonical condition versus 0.944962 under the denser condition, while PoinTr-C+MCP achieves 0.973472 versus 0.976531.
The response to candidate discretization is therefore method-dependent rather than uniformly favorable to either candidate set.

At $K=30$, both conditions have the same top-three method set: PoinTr-C+NBV, Random+TSP, and PoinTr-C+MCP.
PoinTr-C+NBV ranks first under the canonical condition and third under the denser condition, while PoinTr-C+MCP exhibits the reverse ordering.
The canonical top-three NSC@0.02 values range from 0.997187 to 0.997657.
Thus, numerical reordering occurs among methods that all attain near-saturated normalized coverage.

Under Automatic, RSE ranks first and PoinTr-C+SCP second under the canonical condition, while their positions are reversed under the denser condition.
RSE-Mov achieves 0.990847 under the canonical condition versus 0.984623 under the denser condition, changing its rank from sixth to third.
MA-SCVP's unchanged value of 0.990456 places it fourth under the canonical condition and third under the denser condition.
Its difference from canonical third-place RSE-Mov is approximately 0.00039.
This numerical closeness motivates interpreting the full quality--cost profile rather than emphasizing the third-place boundary.
PoinTr-C+NBV also changes from 0.959838 under the denser condition to 0.979172 under the canonical condition, showing that sensitivity is not confined to nearly saturated leading methods.

\begin{table*}[p]
\centering
\scriptsize
\setlength{\tabcolsep}{4pt}
\caption{Sensitivity of whole-space path-efficiency rankings to candidate-view discretization.
Panel A reports the canonical Tammes-128 condition; Panel B uses denser Tammes-360 candidates for compatible planners.
Methods with mean NSC@0.02 $\geq 0.95$ are ranked by mean path cost within each setting.
Methods below the threshold are retained but left unranked.
Each entry averages 528 hidden-test episodes.
Views include the initial observation, and time bins are in seconds.
Results for NBV-Net, MA-SCVP, and BENBV-Net$^\dagger$ are reused across panels, with BENBV-Net retaining its native dynamic interface.}
\label{tab:app_candidate_path}

\begin{tabular}{lrllrrrr}
\toprule
\multicolumn{8}{c}{\textbf{Panel A: Canonical candidate discretization (Tammes-128)}} \\
\midrule
Setting & Rank & Method & Family
& Views & NSC@0.02 & Path & Time bin \\
\midrule
$K=5$ & 1 & BENBV-Net$^\dagger$ & Learned & 6.0 & 0.956392 & 15.64 & 0.5--1 \\
$K=5$ & 2 & PoinTr-C+MCP & Hybrid & 6.0 & 0.973472 & 19.72 & 0.5--1 \\
$K=5$ & 3 & NBV-Net & Learned & 6.0 & 0.978313 & 24.77 & 0.5--1 \\
$K=5$ & 4 & RSE & Classical & 6.0 & 0.979472 & 25.70 & 0.5--1 \\
$K=5$ & -- & PoinTr-C+NBV & Hybrid & 6.0 & 0.907701 & 17.87 & 1--3 \\
$K=5$ & -- & RSE-Mov & Classical & 6.0 & 0.890913 & 10.37 & 0.5--1 \\
$K=5$ & -- & Random+TSP & Classical & 6.0 & 0.938983 & 14.08 & $<0.1$ \\
\midrule
$K=30$ & 1 & NBV-Net & Learned & 9.9 & 0.987945 & 41.73 & 1--3 \\
$K=30$ & 2 & Random+TSP & Classical & 31.0 & 0.997296 & 42.21 & $<0.1$ \\
$K=30$ & 3 & PoinTr-C+MCP & Hybrid & 31.0 & 0.997187 & 52.68 & 0.5--1 \\
$K=30$ & 4 & BENBV-Net$^\dagger$ & Learned & 28.0 & 0.994973 & 66.98 & 3--5 \\
$K=30$ & 5 & RSE-Mov & Classical & 31.0 & 0.995873 & 84.93 & 3--5 \\
$K=30$ & 6 & PoinTr-C+NBV & Hybrid & 31.0 & 0.997657 & 100.65 & $>5$ \\
$K=30$ & 7 & RSE & Classical & 31.0 & 0.996760 & 104.85 & 3--5 \\
\midrule
Automatic & 1 & BENBV-Net$^\dagger$ & Learned & 10.4 & 0.985903 & 27.49 & 1--3 \\
Automatic & 2 & PoinTr-C+SCP & Hybrid & 15.9 & 0.991254 & 30.75 & 0.5--1 \\
Automatic & 3 & RSE-Mov & Classical & 12.1 & 0.990847 & 33.89 & 1--3 \\
Automatic & 4 & NBV-Net & Learned & 8.4 & 0.987454 & 35.76 & 1--3 \\
Automatic & 5 & PoinTr-C+NBV & Hybrid & 12.0 & 0.979172 & 38.61 & 3--5 \\
Automatic & 6 & RSE & Classical & 9.8 & 0.991411 & 41.71 & 1--3 \\
Automatic & 7 & MA-SCVP & Learned & 49.5 & 0.990456 & 57.19 & 0.1--0.5 \\
\bottomrule
\end{tabular}

\vspace{0.8em}

\begin{tabular}{lrllrrrr}
\toprule
\multicolumn{8}{c}{\textbf{Panel B: Denser candidate discretization (Tammes-360)}} \\
\midrule
Setting & Rank & Method & Family
& Views & NSC@0.02 & Path & Time bin \\
\midrule
$K=5$ & 1 & BENBV-Net$^\dagger$ & Learned & 6.0 & 0.956392 & 15.64 & 0.5--1 \\
$K=5$ & 2 & PoinTr-C+MCP & Hybrid & 6.0 & 0.976531 & 20.28 & 1--3 \\
$K=5$ & 3 & NBV-Net & Learned & 6.0 & 0.978313 & 24.77 & 0.5--1 \\
$K=5$ & 4 & RSE & Classical & 6.0 & 0.979525 & 25.66 & 0.5--1 \\
$K=5$ & -- & PoinTr-C+NBV & Hybrid & 6.0 & 0.885928 & 17.19 & 1--3 \\
$K=5$ & -- & RSE-Mov & Classical & 6.0 & 0.734758 & 5.75 & 1--3 \\
$K=5$ & -- & Random+TSP & Classical & 6.0 & 0.944962 & 14.16 & $<0.1$ \\
\midrule
$K=30$ & 1 & Random+TSP & Classical & 31.0 & 0.996438 & 40.54 & $<0.1$ \\
$K=30$ & 2 & NBV-Net & Learned & 9.9 & 0.987945 & 41.73 & 1--3 \\
$K=30$ & 3 & PoinTr-C+MCP & Hybrid & 31.0 & 0.997787 & 51.35 & 1--3 \\
$K=30$ & 4 & BENBV-Net$^\dagger$ & Learned & 28.0 & 0.994973 & 66.98 & 3--5 \\
$K=30$ & 5 & RSE-Mov & Classical & 31.0 & 0.994724 & 68.98 & $>5$ \\
$K=30$ & 6 & RSE & Classical & 31.0 & 0.994842 & 90.94 & $>5$ \\
$K=30$ & 7 & PoinTr-C+NBV & Hybrid & 31.0 & 0.995689 & 99.11 & $>5$ \\
\midrule
Automatic & 1 & BENBV-Net$^\dagger$ & Learned & 10.4 & 0.985903 & 27.49 & 1--3 \\
Automatic & 2 & PoinTr-C+SCP & Hybrid & 15.5 & 0.992507 & 30.17 & 0.5--1 \\
Automatic & 3 & RSE-Mov & Classical & 14.1 & 0.984623 & 31.92 & 3--5 \\
Automatic & 4 & NBV-Net & Learned & 8.4 & 0.987454 & 35.76 & 1--3 \\
Automatic & 5 & PoinTr-C+NBV & Hybrid & 11.7 & 0.959838 & 36.21 & 3--5 \\
Automatic & 6 & RSE & Classical & 9.6 & 0.990593 & 40.70 & 1--3 \\
Automatic & 7 & MA-SCVP & Learned & 49.5 & 0.990456 & 57.19 & 0.1--0.5 \\
\bottomrule
\end{tabular}
\end{table*}

\begin{table*}[t]
\centering
\scriptsize
\setlength{\tabcolsep}{4pt}
\caption{Sensitivity of the whole-to-quarter comparison to candidate-view discretization.
Panel A reports the canonical Tammes-128 condition; Panel B uses denser Tammes-360 candidates for compatible planners.
Each method is evaluated at its maximum-view deployment setting selected under \texttt{whole}, with the same setting used under \texttt{quarter}.
W/Q denotes \texttt{whole}/\texttt{quarter}.
$\Delta$NSC and $\Delta$SC are computed as W minus Q from unrounded values.
Each entry averages 528 hidden-test episodes.
Results for NBV-Net, MA-SCVP, and BENBV-Net$^\dagger$ are reused across panels, with BENBV-Net retaining its native dynamic interface.}
\label{tab:app_candidate_reachability}

\resizebox{\textwidth}{!}{
\begin{tabular}{llrrrrr}
\toprule
\multicolumn{7}{c}{\textbf{Panel A: Canonical candidate discretization (Tammes-128)}} \\
\midrule
Method & Family & Views W/Q & NSC@0.02 W/Q
& $\Delta$NSC & SC@0.02 W/Q & $\Delta$SC \\
\midrule
Random+TSP & Classical & 31.0/31.0 & 0.997296/0.997847 & -0.000551 & 0.9434/0.8076 & 0.1358 \\
RSE & Classical & 31.0/31.0 & 0.996760/0.998155 & -0.001396 & 0.9433/0.8086 & 0.1346 \\
RSE-Mov & Classical & 31.0/31.0 & 0.995873/0.998144 & -0.002271 & 0.9424/0.8086 & 0.1338 \\
NBV-Net & Learned & 9.9/1.2 & 0.987945/0.472436 & 0.515509 & 0.9318/0.3869 & 0.5450 \\
BENBV-Net$^\dagger$ & Learned & 28.0/23.9 & 0.994973/0.979299 & 0.015674 & 0.9404/0.7838 & 0.1567 \\
MA-SCVP & Learned & 49.5/12.5 & 0.990456/0.966996 & 0.023461 & 0.9375/0.7720 & 0.1655 \\
PoinTr-C+NBV & Hybrid & 31.0/31.0 & 0.997657/0.998115 & -0.000458 & 0.9445/0.8085 & 0.1360 \\
PoinTr-C+SCP & Hybrid & 15.9/14.8 & 0.991254/0.993774 & -0.002521 & 0.9366/0.7985 & 0.1381 \\
PoinTr-C+MCP & Hybrid & 31.0/31.0 & 0.997187/0.997885 & -0.000697 & 0.9437/0.8079 & 0.1358 \\
\midrule
\multicolumn{7}{c}{\textbf{Panel B: Denser candidate discretization (Tammes-360)}} \\
\midrule
Method & Family & Views W/Q & NSC@0.02 W/Q
& $\Delta$NSC & SC@0.02 W/Q & $\Delta$SC \\
\midrule
Random+TSP & Classical & 31.0/31.0 & 0.996438/0.996793 & -0.000356 & 0.9424/0.8052 & 0.1371 \\
RSE & Classical & 31.0/31.0 & 0.994842/0.997905 & -0.003064 & 0.9416/0.8134 & 0.1282 \\
RSE-Mov & Classical & 31.0/31.0 & 0.994724/0.998064 & -0.003341 & 0.9412/0.8133 & 0.1279 \\
NBV-Net & Learned & 9.9/1.2 & 0.987945/0.472436 & 0.515509 & 0.9318/0.3869 & 0.5450 \\
BENBV-Net$^\dagger$ & Learned & 28.0/23.9 & 0.994973/0.979299 & 0.015674 & 0.9404/0.7838 & 0.1567 \\
MA-SCVP & Learned & 49.5/12.5 & 0.990456/0.966996 & 0.023461 & 0.9375/0.7720 & 0.1655 \\
PoinTr-C+NBV & Hybrid & 31.0/31.0 & 0.995689/0.998460 & -0.002771 & 0.9420/0.8132 & 0.1288 \\
PoinTr-C+SCP & Hybrid & 15.5/20.1 & 0.992507/0.996710 & -0.004202 & 0.9374/0.8066 & 0.1308 \\
PoinTr-C+MCP & Hybrid & 31.0/31.0 & 0.997787/0.998934 & -0.001147 & 0.9440/0.8132 & 0.1308 \\
\bottomrule
\end{tabular}
}
\end{table*}

\paragraph{Path efficiency and executed views.}
The path-efficiency ordering is more stable than the quality ordering in these comparisons.
At $K=5$, both the qualifying method set and its complete path ordering are identical under the two discretizations.
At $K=30$, the top-three set remains NBV-Net, Random+TSP, and PoinTr-C+MCP, with the first two exchanging positions.
Under Automatic, the complete path ordering remains unchanged, with BENBV-Net, PoinTr-C+SCP, and RSE-Mov retaining the first three positions.

Stable ordering can nevertheless coexist with appreciable changes in motion cost.
At $K=30$, RSE's path cost is 104.85 under the canonical condition versus 90.94 under the denser condition.
RSE-Mov similarly has path costs of 84.93 and 68.98, respectively.
Both methods execute 31 views under either condition, so the path differences reflect changes in their trajectories rather than acquisition count.
A smaller candidate set therefore does not necessarily produce a shorter execution path.

Automatic plan sizes also depend on candidate discretization.
Under \texttt{whole}, RSE-Mov executes 12.1 views under the canonical condition versus 14.1 under the denser condition, but its path cost is 33.89 versus 31.92.
PoinTr-C+SCP executes 15.9 versus 15.5 views, with path costs of 30.75 and 30.17.
These examples show that candidate discretization can affect termination, plan size, and spatial ordering differently.

Path ranks must be interpreted together with executed view counts and attained quality.
For example, NBV-Net's $K=30$ result includes early termination and averages only 9.9 acquired views.
Its short path therefore does not imply that it executes a more efficient 31-view trajectory.
The comparison retains these native termination outcomes instead of forcing all methods to exhaust the requested budget.

Runtime bins are included for completeness.
Their changes should not be attributed solely to candidate count, since the reported runtime includes method-specific processing and remains sensitive to hardware and system load.

\paragraph{Reachability comparison.}
Table~\ref{tab:app_candidate_reachability} compares \texttt{whole} and \texttt{quarter} under both candidate discretizations.
For each method and candidate condition, we select the setting with the largest mean executed view count under \texttt{whole}, then evaluate the same setting under \texttt{quarter}.
For the reported methods, this selects $K=30$ for budgeted and iterative methods, and Automatic for native finite set-cover methods.
All differences are computed as \texttt{whole} minus \texttt{quarter} from unrounded aggregate values.

Under the canonical Tammes-128 condition, classical and hybrid planners achieve \texttt{quarter} NSC@0.02 values between approximately 0.994 and 0.998 at the selected settings.
Their raw SC@0.02 nevertheless decreases by approximately 0.134--0.138 relative to \texttt{whole}.
The denser condition exhibits the same qualitative distinction between near-saturated constrained-reference coverage and reduced full-object coverage.
Slightly higher NSC under \texttt{quarter} must be interpreted relative to its smaller, constraint-specific observable reference.

The learned-interface behaviors remain visible in both panels.
NBV-Net's mean acquired views decrease from 9.9 to 1.2 under the reachability constraint, accompanied by an NSC@0.02 decrease of approximately 0.516.
MA-SCVP's executed plan decreases from 49.5 to 12.5 views, with NSC@0.02 decreasing by approximately 0.0235.
BENBV-Net retains more extensive acquisition under the constraint, although its NSC@0.02 also decreases.
These identical entries show how the existing learned interfaces compare with compatible planners under either discretization; they are not independent reruns of the learned methods.

Candidate discretization still affects individual constrained plans.
For PoinTr-C+SCP, the \texttt{quarter} plan contains 14.8 views under the canonical condition and 20.1 under the denser condition.
Its corresponding NSC@0.02 values are 0.993774 and 0.996710.
Thus, preservation of the broad reachability finding does not imply invariant plan size or identical final coverage.

\paragraph{Implications.}
The principal budget- and reachability-related findings are supported under both candidate discretizations, despite method-dependent changes in absolute performance and fine-grained rankings.
Small and large acquisition budgets continue to produce different quality--cost comparisons, automatic termination retains method-specific behavior, and reachability restrictions continue to separate constrained-reference coverage from full-object coverage.
Candidate discretization can nevertheless materially affect individual trajectories, plan sizes, and heuristic behavior.
The canonical Tammes-128 condition provides the primary results used in the main text, while the denser Tammes-360 condition serves as a sensitivity analysis.
BENBV-Net remains explicitly identified as a native dynamic-interface method in both conditions.

\section{Slow-Saturation Case Study}
\label{app:slow_saturation_case_study}

\begin{figure}[t]
  \centering
  \includegraphics[width=0.90\linewidth]{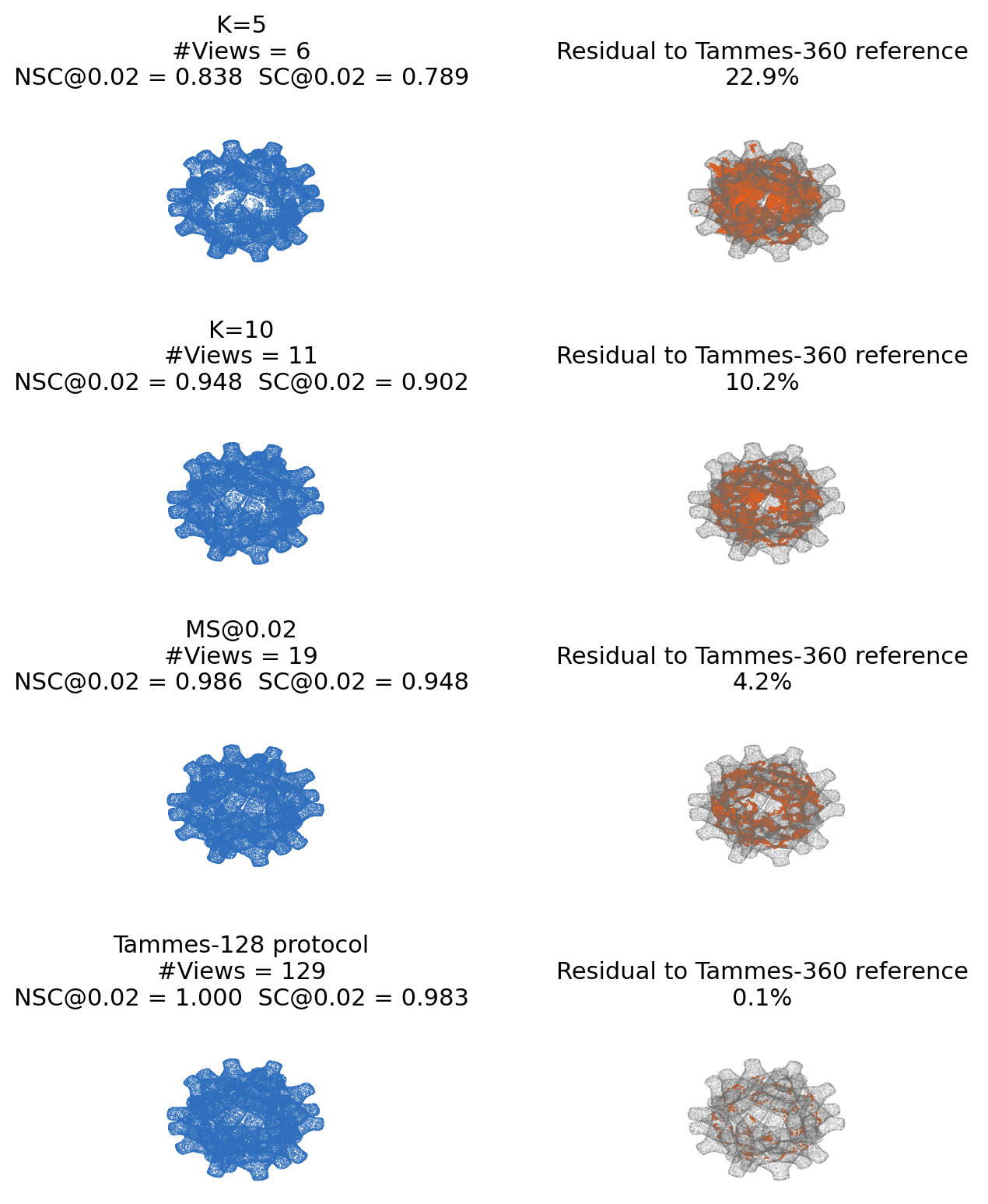}
\caption{
Qualitative visualization of the representative slow-saturation object with $d_{\mathrm{sat}}=162$ from Figure~\ref{fig:difficulty_statistics}.
The rollout is a ground-truth oracle rollout under the canonical \texttt{whole} view-space setting and is used only for diagnostic visualization.
The left column shows the cumulative observed surface at fixed-budget checkpoints, the MS@0.02 stopping checkpoint, and the exhausted Tammes-128 protocol.
The right column shows the residual observable surface with respect to a denser Tammes-360 reference.
Blue points denote the surface observed at the checkpoint, gray points denote the Tammes-360 reference, and orange points denote the residual surface not yet observed at that checkpoint.
Even when MS@0.02 reaches high normalized surface coverage, small residual regions remain visible under denser view-space sampling.
}
  \label{fig:slow_saturation_case}
\end{figure}

Figure~\ref{fig:slow_saturation_case} visualizes the representative object with observation saturation difficulty $d_{\mathrm{sat}}=162$ from Figure~\ref{fig:difficulty_statistics}. 
The visualization is generated under the canonical \texttt{whole} view-space protocol using the same ground-truth oracle rollout used in the slow-saturation diagnostic analysis, rather than from a learned or deployable planner. 
The left column shows the cumulative observed surface at several benchmark checkpoints, while the right column shows the residual observable surface with respect to a denser Tammes-360 reference. 
Although the oracle rollout reaches high NSC@0.02 by MS@0.02, a visible residual tail remains. 
This illustrates that map stabilization captures short-horizon observed-map convergence, but does not necessarily imply complete long-horizon observable-surface saturation. 
After exhausting the Tammes-128 protocol, the residual to the Tammes-360 reference becomes very small, indicating that the benchmark protocol covers nearly all observable surface for this case while still exposing slow-saturation behavior along the way.

\section{Real-Robot and Real-RGB-D Evaluation}
\label{app:robot}

\subsection{Robot Setup and View-Space Instantiation}

We conduct a small real-robot reachability study to ground the reachable-view protocols used in ObjView-Bench. 
The study is not intended as a full real-robot reconstruction benchmark. 
Instead, its purpose is to measure how much of the ideal object-centric candidate-view sphere remains executable under real robot kinematics, collision constraints, and tabletop placement variation.

We use a UR5e 6-DoF robot arm equipped with a Gemini 336 RGB-D camera. 
Each trial starts from a fixed home pose that provides a global view of the tabletop workspace, as shown in Figure~\ref{fig:teaser}. 
This initial pose is chosen to observe the object and the surrounding table region reliably before any candidate-view feasibility checks are performed.

For each object placement, we first acquire an RGB-D observation from the home pose and extract a workspace-filtered object point cloud. 
We estimate the object center from the segmented point cloud:
\begin{equation}
c_{\mathrm{obj}}
=
\frac{1}{|\mathcal{P}|}
\sum_{p \in \mathcal{P}} p,
\end{equation}
where $\mathcal{P}$ denotes the denoised observed object points in the robot/world frame. 
To obtain a robust object scale, we use the $95$-th percentile of point distances from the estimated center:
\begin{equation}
s_{\mathrm{obj}}
=
\mathrm{percentile}_{95}
\left(
\left\{
\left\|
p - c_{\mathrm{obj}}
\right\|_2
\;\middle|\;
p \in \mathcal{P}
\right\}
\right).
\end{equation}
This robust scale estimate reduces sensitivity to segmentation outliers and sparse depth artifacts.

We then instantiate an object-centric candidate-view sphere around the estimated object center. 
The physical viewing radius is selected from the estimated object size and the minimum reliable working distance of the depth camera. 
The goal is to choose a radius large enough to keep the object within the camera field of view while respecting the valid depth-sensing range of the camera; in our setup, we use a minimum working distance of $23$ cm.
For each normalized benchmark candidate direction, we map the candidate camera center to the real robot frame as:
\begin{equation}
x_i
=
c_{\mathrm{obj}}
+
R_{\mathrm{view}}
\,
\hat{v}_i,
\end{equation}
where $\hat{v}_i$ is the unit candidate-view direction and $R_{\mathrm{view}}$ is the selected physical viewing radius. 
The camera is oriented to look at $c_{\mathrm{obj}}$.

For each candidate pose, we evaluate feasibility using cuRobo~\citep{sundaralingam2023curobo} with multiple roll values for inverse kinematics and smooth motion planning. 
A view is counted as reachable only if a collision-free robot motion to the corresponding camera pose can be found while satisfying the robot joint limits, tabletop collision constraints, and camera pose constraint. 
This produces a binary reachability label for each of the $128$ candidate views.

The supplementary video shows the robot setup, object placements, and representative feasible and infeasible candidate views used in this measurement.

\subsection{Reachable-View Measurement}
\label{app:robot:reachability}

\begin{table}[!t]
\centering
\caption{
Real-robot reachable-view counts under a tabletop setup.
Each entry reports the number of feasible views among the $128$ benchmark candidate views.
Object positions are described with respect to the robot base and are visualized in the supplementary video.
}
\label{tab:real_robot_reachability}
\begin{tabular}{lcccccc}
\toprule
Object & Center & Far-Left & Far-Right & Close-Left & Close-Right & Avg. \\
\midrule
Cleaner & 36/128 & 27/128 & 33/128 & 36/128 & 57/128 & 37.8 \\
Cereal  & 62/128 & 31/128 & 33/128 & 49/128 & 74/128 & 49.8 \\
Cup     & 49/128 & 28/128 & 37/128 & 48/128 & 71/128 & 46.6 \\
Duck    & 58/128 & 26/128 & 33/128 & 46/128 & 67/128 & 46.0 \\
Drill   & 53/128 & 35/128 & 33/128 & 44/128 & 68/128 & 46.6 \\
\midrule
Avg.    & 51.6 & 29.4 & 33.8 & 44.6 & 67.4 & 45.4 \\
\bottomrule
\end{tabular}
\end{table}

To ground the reachable-view protocol in a real robotic setting, we measured how many of the $128$ benchmark candidate views are feasible for a tabletop robot setup. 
We placed five representative objects at five table locations and checked each candidate view using robot kinematic and collision constraints. 
Table~\ref{tab:real_robot_reachability} reports the number of reachable views out of $128$.

Across all object--location pairs, the robot can reach $45.4$ out of $128$ views on average, corresponding to approximately $35.4\%$ of the idealized candidate-view sphere. 
The reachable fraction varies substantially across placements, ranging from $26/128$ to $74/128$. 
This supports our use of reachable-view constraints as a deployment-relevant evaluation condition: in typical tabletop settings, reachability is not a small perturbation of the ideal view sphere, but a substantial contraction of the executable view space.
The benchmark \textit{quarter} view-space constraint should therefore be interpreted as a controlled abstraction of this reachability contraction, rather than an exact model of a particular robot placement.
It provides a standardized and reproducible stress test whose degree of view-space contraction is of the same order as that observed in our physical measurements.

\subsection{Learned-Planner Executability under Physical Reachability}
\label{app:robot:executability}

We next examine how physical reachability interacts with two representative learned planning interfaces: NBV-Net, which predicts a single discrete next-view index, and MA-SCVP, which predicts a finite set of candidate views.
We evaluate both methods over all $25$ object--placement configurations from the real-robot reachability study, using every physically reachable Tammes-128 candidate view as a possible initial observation.
This yields $1{,}134$ reachable-view starts in total.
Unlike the regular spherical \texttt{quarter} constraint used in simulation, these physical reachable sets are sparse and irregular because they are shaped jointly by robot kinematics, collision constraints, tabletop geometry, and camera working-distance limits.
We use this experiment as an action-interface executability diagnostic rather than as a full real-robot reconstruction benchmark.

For NBV-Net, each rollout starts from a physically reachable candidate view and the network predicts a single next-view index at each step.
If the predicted view is infeasible or already visited, we terminate the rollout rather than introduce an additional fallback policy not defined by the original method.
For MA-SCVP, the learned planner uses its farthest-point bootstrap and then predicts a finite residual view set.
The predicted set is filtered using the corresponding physical reachability mask, and the remaining feasible views form the executable plan.
Rates and view counts are first computed within each object--placement configuration and then macro-averaged across the $25$ configurations, preventing configurations with larger reachable sets from dominating the aggregate statistics.

Table~\ref{tab:robot_action_interface_diagnostic} summarizes the resulting behavior.
NBV-Net executes only $0.18$ additional views on average beyond the initial observation, yielding $1.18$ total acquired views, with $87.2\%$ of rollouts failing at the first planning decision because the predicted next view is infeasible.
MA-SCVP is less brittle to a single infeasible prediction because its finite predicted set can be filtered before execution, but physical reachability still removes a large fraction of the predicted plan.
On average, a raw prediction of $70.05$ views is reduced to $22.93$ retained feasible views, corresponding to $24.93$ total views after including the initial observation and the native farthest-point bootstrap view.
Overall, the two methods exhibit distinct interface-level effects under physical reachability: hard early termination for single-index prediction and substantial plan truncation for finite-set prediction.
We do not include a separate reinforcement-learning baseline in the current benchmark, but RL policies with the same single-view discrete action interface may exhibit similar interface-level sensitivity under reachability constraints.

\begin{table}[t]
\centering
\caption{
Learned-planner executability under physical reachability over 25 object--placement configurations and 1,134 reachable-view starts.
Rates and view counts are computed within each configuration and then macro-averaged over configurations.
Total views include the initial observation; for MA-SCVP, they also include the farthest-point bootstrap view and the retained predicted views.
For MA-SCVP, the removal fraction is computed for each rollout before configuration-level and cross-configuration averaging.
}
\label{tab:robot_action_interface_diagnostic}
\begin{tabular}{lcccc}
\toprule
Method & Raw output & Executable output & Total views & Constraint effect \\
\midrule
NBV-Net
& 1 index/step
& 0.18 additional views
& 1.18
& 87.2\% first-decision failure \\
MA-SCVP
& 70.05 views
& 22.93 retained views
& 24.93
& 69.5\% prediction removed \\
\bottomrule
\end{tabular}
\end{table}

\subsection{Shared-Reachable Replay on Real RGB-D}
\label{app:robot:replay}

Having established the effect of physical reachability on the executable view space and learned planning interfaces, we next evaluate planner behavior using real RGB-D observations acquired at physically reachable candidate views.
For each of the $25$ object--placement configurations, the robot captures observations from the reachable Tammes-128 poses, forming a placement-specific pool of posed real RGB-D frames.
Compatible planners are then evaluated by replaying their selected views against this shared captured pool under the same physical reachability constraints.
This protocol preserves the planners' sequential decision process while ensuring that different methods receive the same underlying sensor evidence for a given view.
It is intended as a controlled real-sensing replay study rather than a repeated closed-loop robot reconstruction benchmark.

\paragraph{Real-capture protocol.}
For each object--placement configuration, we instantiate the physical Tammes-128 candidate sphere using the home-view RGB-D observation: the object center and robust scale are estimated from the observed point cloud, and the physical viewing radius is selected from the estimated object size and the minimum reliable camera working distance.
The robot motion-planning stack then evaluates all candidate poses under the same kinematic, collision, and camera-pose constraints used in the reachability study.
For every candidate pose deemed reachable, the robot executes the corresponding motion and acquires a posed RGB-D frame.
The resulting placement-specific observation pool therefore contains real sensor measurements collected only at physically executable viewpoints, rather than rendered observations or synthetically masked views.

\paragraph{Shared replay protocol.}
Before replay, each captured RGB-D frame is converted to a posed point cloud and restricted using a fixed placement-specific workspace crop.
The crop is defined once for each object--placement configuration based on the measured table height and a fixed local workspace, manually verified before evaluation, and then applied identically to all methods and all captured views from that configuration.
This preprocessing is performed after physical reachability has been determined and therefore does not affect the reachable-view mask or introduce any unobserved surface geometry.
During replay, all methods receive the same initial observation, the same physically reachable candidate-view set, and the same pre-captured real RGB-D observation associated with each selected view.
Each method nevertheless retains its native internal state representation and update procedure; the shared protocol standardizes the available sensor evidence rather than forcing a common planner-internal fusion mechanism.

Replay follows each method's native interaction pattern.
Iterative planners select one candidate view at a time, receive the corresponding real RGB-D observation, update their internal state, and then make the next decision.
Set-based or budgeted planners first produce an ordered finite plan under the current reachable-view constraint, after which the corresponding captured observations are consumed sequentially in that order.
Pre-capturing the physically reachable views avoids repeating identical robot motions for every planner and ensures that methods selecting the same candidate view receive the same underlying sensor measurement.
The protocol therefore preserves the benchmark's sequential planning semantics while factoring out repeated low-level robot execution variability.

We evaluate the eight baselines whose native action spaces can be replayed faithfully from the fixed Tammes-128 capture pool.
BENBV-Net is excluded from this real-capture replay because it generates continuous online viewpoints rather than selecting from the fixed Tammes-128 candidate set, so its native actions cannot be reproduced faithfully from the pre-captured observation pool.

\paragraph{Evaluation regimes.}
To mirror the three execution regimes used in the synthetic benchmark, we evaluate the real-capture replay under a small fixed budget, a large fixed budget, and native finite-plan termination.
Specifically, we use $K=5$ and $K=30$ as the small- and large-budget settings, where $K$ denotes the maximum number of additional acquisitions beyond the initial observation.
Iterative planners may terminate before reaching the specified budget according to their native stopping or feasibility behavior.
For sequential methods whose trajectories are generated under the $K=30$ cap, the $K=5$ result is obtained from the corresponding trajectory prefix consisting of the initial observation and the first five acquired views.

For the automatic regime, we report native finite-plan termination for MA-SCVP and PoinTr-C+SCP, consistent with their execution semantics in the synthetic benchmark.
We restrict automatic evaluation in this real-capture study to native finite-plan termination.
Map-stabilization stopping could also be computed from the accumulated observations without ground-truth geometry, but its behavior under sensing noise and registration error requires separate validation.
For $K=30$, three of the $25$ configurations contain fewer than $31$ physically reachable views, so full-budget completion is possible for $88\%$ of the configurations.

\paragraph{Captured-reference coverage.}
The synthetic benchmark evaluates reconstruction quality against known object geometry, but an equivalent geometric reference is unavailable for the physical objects.
In particular, strict SC cannot be computed without the complete object surface, and the clean externally observable reference used by NSC cannot be reproduced from real captures alone.
We therefore construct a placement-specific empirical captured reference, $\mathcal{R}_{\mathrm{cap}}$, by fusing the cropped RGB-D observations from all physically reachable candidate views using a common $5$\,mm voxelization.
For a replay trajectory, we similarly fuse the observations acquired from its executed views into $\mathcal{R}_{\mathrm{traj}}$ and define Captured-Reference Coverage as
\begin{equation}
\mathrm{CRC}@\tau
=
\frac{
\left|
\left\{
p \in \mathcal{R}_{\mathrm{cap}}
\;\middle|\;
\min_{q \in \mathcal{R}_{\mathrm{traj}}}
\|p-q\|_2 \le \tau
\right\}
\right|
}{
|\mathcal{R}_{\mathrm{cap}}|
}.
\end{equation}
We report CRC at $\tau\in\{1\,\mathrm{cm},\,2\,\mathrm{cm}\}$.
CRC measures how much of the surface evidence captured across all physically reachable real observations is recovered by the executed-view reconstruction.
Because $\mathcal{R}_{\mathrm{cap}}$ itself contains depth noise, pose and calibration error, and finite-view sampling effects, CRC is used only as a controlled relative proxy under the shared real-capture protocol and is not interpreted as ground-truth object-surface coverage.

\paragraph{Path cost and aggregation.}
We report the same normalized collision-aware view-space path cost used in the synthetic benchmark.
For each transition, path length is computed between benchmark-normalized camera centers using the benchmark collision-aware distance around the unit-sphere obstacle, rather than direct Euclidean distance.
This quantity should therefore be interpreted as a standardized camera-motion proxy, not as measured robot joint-space trajectory length.
The evaluator additionally computes the corresponding path in world coordinates, and the two representations are related by the placement-specific physical viewing radius.

For all reported metrics, we first average over physically reachable start views within each object--placement configuration and then macro-average the resulting configuration-level means over the $25$ configurations.
This gives each physical configuration equal weight regardless of how many candidate views are reachable in that placement.

\begin{table}[t]
\centering
\caption{
Shared-reachable replay on real RGB-D over $25$ physical object--placement configurations.
Views include the initial observation.
CRC measures coverage of the placement-specific captured reference, and Path reports the normalized collision-aware view-space path cost.
Statistics are first averaged over physically reachable starts within each configuration and then macro-averaged over configurations.
\(^{\dagger}\)The low NBV-Net path cost results from early invalid-action termination and should not be interpreted as motion efficiency.
}
\label{tab:real_rgbd_replay}
\small
\setlength{\tabcolsep}{4pt}
\begin{tabular}{lcccc}
\toprule
Method & Views & CRC@1cm $\uparrow$ & CRC@2cm $\uparrow$ & Path $\downarrow$ \\
\midrule

\multicolumn{5}{l}{\textit{Fixed-budget cap, $K=5$}} \\
Random+TSP     & 6.00  & 0.862 & 0.953 & 10.28 \\
NBV-Net        & 1.18  & 0.517 & 0.709 & 0.89$^\dagger$  \\
RSE            & 6.00  & 0.891 & 0.968 & 20.74 \\
RSE-Mov        & 6.00  & 0.872 & 0.961 & 13.13 \\
PoinTr-C+NBV   & 5.98  & 0.838 & 0.940 & 12.63 \\
PoinTr-C+MCP   & 6.00  & 0.895 & 0.969 & 15.45 \\
\midrule

\multicolumn{5}{l}{\textit{Fixed-budget cap, $K=30$}} \\
Random+TSP     & 30.52 & 0.983 & 0.996 & 30.95 \\
NBV-Net        & 1.18  & 0.517 & 0.709 & 0.89$^\dagger$  \\
RSE            & 30.52 & 0.981 & 0.996 & 87.04 \\
RSE-Mov        & 30.52 & 0.981 & 0.996 & 61.51 \\
PoinTr-C+NBV   & 30.43 & 0.980 & 0.993 & 68.67 \\
PoinTr-C+MCP   & 30.52 & 0.985 & 0.997 & 35.96 \\
\midrule

\multicolumn{5}{l}{\textit{Native finite-plan termination}} \\
MA-SCVP        & 24.93 & 0.939 & 0.981 & 29.46 \\
PoinTr-C+SCP   & 13.82 & 0.937 & 0.984 & 22.49 \\
\bottomrule
\end{tabular}
\end{table}

Table~\ref{tab:real_rgbd_replay} reports the shared-reachable real-RGB-D replay results under the three evaluation regimes described above.
The real-capture replay reproduces several of the principal budget-dependent behaviors observed in the synthetic benchmark.
At the small budget, geometry-aware planners retain a measurable advantage in captured-reference coverage: PoinTr-C+MCP reaches $\mathrm{CRC}@2\mathrm{cm}=0.969$, compared with $0.953$ for Random+TSP, while RSE attains a similar level at $0.968$.
As the budget increases to $K=30$, these coverage differences largely collapse near saturation: all non-terminating methods reach $\mathrm{CRC}@2\mathrm{cm}\geq0.993$.
The larger budget nevertheless preserves substantial differences in path efficiency.
In particular, RSE-Mov matches RSE in $\mathrm{CRC}@2\mathrm{cm}$ ($0.996$ for both) while reducing normalized path cost from $87.04$ to $61.51$.

The real-capture results also preserve the reachability-induced distinction between the two learned planning interfaces.
NBV-Net remains dominated by infeasible-action termination, averaging only $1.18$ total views and correspondingly low captured-reference coverage under both fixed-budget caps.
In contrast, finite-set planning remains executable after feasibility filtering: MA-SCVP reaches $\mathrm{CRC}@2\mathrm{cm}=0.981$ with $24.93$ total views under native plan termination.
PoinTr-C+SCP achieves a similar coverage level, $\mathrm{CRC}@2\mathrm{cm}=0.984$, with fewer views ($13.82$) and a lower normalized path cost ($22.49$ versus $29.46$).
These results are consistent with the physical executability analysis: single-index prediction is brittle to feasibility mismatch, whereas finite-plan methods can retain an executable subset under the same real reachability constraints.

Taken together, the shared-reachable replay provides complementary physical evidence for the benchmark conclusions under both robot-derived reachability and real RGB-D sensing.
Several method-level behaviors observed in simulation, including budget-dependent coverage saturation, the quality--motion trade-off of RSE-Mov, and the contrasting failure modes of single-index and finite-plan interfaces, remain visible under the real-capture protocol.
At the same time, this experiment should not be interpreted as a full closed-loop robotic reconstruction benchmark: candidate poses are pre-captured once and subsequently replayed, low-level execution variability is not reintroduced for each planner, and CRC is defined relative to an empirical captured reference rather than complete object geometry.
The study therefore serves as a controlled bridge between the synthetic benchmark and physical deployment, while leaving repeated closed-loop execution with geometric ground truth or external metrology to future work.

\newpage
\section*{NeurIPS Paper Checklist}

\begin{enumerate}

\item {\bf Claims}
    \item[] Question: Do the main claims made in the abstract and introduction accurately reflect the paper's contributions and scope?
    \item[] Answer: \answerYes{} 
    \item[] Justification: The main claims made in the abstract and introduction accurately reflect the
main paper’s contributions and scope.

\item {\bf Limitations}
    \item[] Question: Does the paper discuss the limitations of the work performed by the authors?
    \item[] Answer: \answerYes{} 
    \item[] Justification: The paper discusses limitations and future extensions in Section~\ref{sec:conclusion}, including the current focus on single-object object-centric reconstruction, idealized rendered RGB-D observations, and future extensions to appearance-centric objectives, sensing noise, multi-object scenes, and broader deployment settings.

\item {\bf Theory assumptions and proofs}
    \item[] Question: For each theoretical result, does the paper provide the full set of assumptions and a complete (and correct) proof?
    \item[] Answer: \answerNA{} 
    \item[] Justification: The paper does not present formal theoretical results requiring proof. The mathematical formulations in the paper define benchmark quantities, evaluation metrics, and optimization-based difficulty annotations; their implementation details are provided in the appendix.

    \item {\bf Experimental result reproducibility}
    \item[] Question: Does the paper fully disclose all the information needed to reproduce the main experimental results of the paper to the extent that it affects the main claims and/or conclusions of the paper (regardless of whether the code and data are provided or not)?
    \item[] Answer: \answerYes{} 
    \item[] Justification: The paper fully discloses all the information required to reproduce the main experimental results. We provide a detailed description of the dataset construction process, the benchmark, and evaluation metrics to support reproducibility. \textbf{Appendices} contain all necessary details for reproducing our results.

\item {\bf Open access to data and code}
    \item[] Question: Does the paper provide open access to the data and code, with sufficient instructions to faithfully reproduce the main experimental results, as described in supplemental material?
    \item[] Answer: \answerYes{} 
    \item[] Justification: The benchmark code, dataset, and reproduction instructions are publicly accessible through our project website.

\item {\bf Experimental setting/details}
    \item[] Question: Does the paper specify all the training and test details (e.g., data splits, hyperparameters, how they were chosen, type of optimizer) necessary to understand the results?
    \item[] Answer: \answerYes{} 
    \item[] Justification: Experiments section and the Appendices contain all experimental details.

\item {\bf Experiment statistical significance}
    \item[] Question: Does the paper report error bars suitably and correctly defined or other appropriate information about the statistical significance of the experiments?
    \item[] Answer: \answerNo{} 
    \item[] Justification: This paper is primarily a benchmarking study, where the main objective is to compare and rank different methods rather than to claim statistically significant marginal improvements. As such, error bars or statistical significance tests are not reported for the experimental results.

\item {\bf Experiments compute resources}
    \item[] Question: For each experiment, does the paper provide sufficient information on the computer resources (type of compute workers, memory, time of execution) needed to reproduce the experiments?
    \item[] Answer: \answerYes{} 
    \item[] Justification: We report computational and human annotation resources in the \textbf{appendix}, along with details of the compute resources and hyper-parameters used for the experiments.
    
\item {\bf Code of ethics}
    \item[] Question: Does the research conducted in the paper conform, in every respect, with the NeurIPS Code of Ethics \url{https://neurips.cc/public/EthicsGuidelines}?
    \item[] Answer: \answerYes{} 
    \item[] Justification: We confirm the research conducted in the paper conform, in every respect, with the NeurIPS Code of Ethics.

\item {\bf Broader impacts}
    \item[] Question: Does the paper discuss both potential positive societal impacts and negative societal impacts of the work performed?
    \item[] Answer: \answerYes{} 
    \item[] Justification: ObjView-Bench supports reproducible evaluation of robotic active perception and geometric reconstruction. While the benchmark itself focuses on object-centric evaluation, downstream reconstruction technologies could raise privacy concerns when applied to sensitive environments without appropriate consent.
    
\item {\bf Safeguards}
    \item[] Question: Does the paper describe safeguards that have been put in place for responsible release of data or models that have a high risk for misuse (e.g., pre-trained language models, image generators, or scraped datasets)?
    \item[] Answer: \answerNA{} 
    \item[] Justification: Our work does not involve high-risk models or sensitive datasets that would require specific safeguards. The released benchmark dataset is constructed from publicly available or appropriately processed data, and does not pose significant risks of misuse.

\item {\bf Licenses for existing assets}
    \item[] Question: Are the creators or original owners of assets (e.g., code, data, models), used in the paper, properly credited and are the license and terms of use explicitly mentioned and properly respected?
    \item[] Answer: \answerYes{} 
    \item[] Justification: All assets are well cited and properly credited.

\item {\bf New assets}
    \item[] Question: Are new assets introduced in the paper well documented and is the documentation provided alongside the assets?
    \item[] Answer: \answerYes{} 
    \item[] Justification: All assets are documented and the documentation is released with our codes.

\item {\bf Crowdsourcing and research with human subjects}
    \item[] Question: For crowdsourcing experiments and research with human subjects, does the paper include the full text of instructions given to participants and screenshots, if applicable, as well as details about compensation (if any)? 
    \item[] Answer: \answerNA{} 
    \item[] Justification: The paper does not involve crowdsourcing nor research with human subjects.

\item {\bf Institutional review board (IRB) approvals or equivalent for research with human subjects}
    \item[] Question: Does the paper describe potential risks incurred by study participants, whether such risks were disclosed to the subjects, and whether Institutional Review Board (IRB) approvals (or an equivalent approval/review based on the requirements of your country or institution) were obtained?
    \item[] Answer: \answerNA{} 
    \item[] Justification: The paper does not involve crowdsourcing nor research with human subjects.

\item {\bf Declaration of LLM usage}
    \item[] Question: Does the paper describe the usage of LLMs if it is an important, original, or non-standard component of the core methods in this research? Note that if the LLM is used only for writing, editing, or formatting purposes and does \emph{not} impact the core methodology, scientific rigor, or originality of the research, declaration is not required.
    \item[] Answer: \answerNA{} 
    \item[] Justification: LLMs were used only for writing, editing, formatting, and auxiliary implementation assistance. They were not used as part of the proposed benchmark definitions, experimental protocol, evaluation criteria, data generation decisions, or any original scientific component of the research.

\end{enumerate}

\end{document}